\documentclass[conference]{IEEEtran}

\usepackage[numbers]{natbib}
\usepackage{multicol}
\usepackage{multirow} %
\usepackage{graphicx} %
\usepackage{booktabs} %
\usepackage{siunitx} 
\DeclareSIUnit\megapixel{MP}
\DeclareSIUnit{\decibel}{dB}
\DeclareSIUnit{\fps}{fps}
\DeclareSIUnit{\USD}{\text{USD}}
\DeclareSIUnit{\ggrav}{\ensuremath{g_{\text{grav}}}}
\DeclareSIUnit{\Mbps}{Mbps}
\DeclareSIUnit{\Gbps}{Gbps}
\DeclareSIUnit{\ppm}{ppm}
\DeclareSIUnit\points{points}
\usepackage{subcaption}
\usepackage{comment}
\usepackage{makecell}
\usepackage{xcolor} %

\usepackage{amssymb}
\usepackage{amsmath}
\usepackage{todonotes}
\usepackage{gensymb} %
\usepackage{layouts}
\usepackage[flushleft]{threeparttable}

\usepackage{lipsum}
\usepackage{times}
\usepackage[bottom]{footmisc}
\usepackage[utf8]{inputenc} %
\usepackage[T1]{fontenc}    %
\usepackage{url}            %
\usepackage{nicefrac}       %
\usepackage{microtype}      %
\usepackage{adjustbox}      %
\usepackage{graphics,color}

\usepackage{algorithm}
\usepackage[noend]{algpseudocode}
\algrenewcommand\algorithmicindent{1em}
\algrenewcommand{\algorithmiccomment}[1]{%
\bgroup\hskip2em\textcolor{ourdarkgreen}{//~\textsl{#1}}\egroup}

\usepackage{amsfonts}       %
\usepackage{xspace}

\let\labelindent\relax   %
\usepackage{enumitem}
\usepackage{pifont}
\usepackage{xr-hyper}

\usepackage{tcolorbox}
\usepackage{ifthen}

\newboolean{anonymous}
\setboolean{anonymous}{false}  %

\definecolor{boxi_gray}{RGB}{220,220,220}
\definecolor{boxi_yellow}{RGB}{245,245,245} %
\definecolor{boxi_yellow}{RGB}{245,245,245}
\definecolor{boxi_green}{rgb}{0.0, 0.5, 0.0}
\definecolor{boxi_lightgreen}{rgb}{0.3, 0.65, 0.3}
\newcommand{\tickbox}{\ensuremath{%
  \ooalign{%
    \hfil$\square$\hfil\cr%
    \hidewidth\kern0.3ex\raisebox{0.2ex}{$\checkmark$}\hidewidth\cr%
  }%
}}

\definecolor{scirobred}{rgb}{0.694, 0.0078, 0.0235}
\definecolor{orange}{rgb}{0.88, 0.611, 0.23529} %

\NewDocumentEnvironment{highlightbox}{O{Title not defined}}
{
    \begin{tcolorbox}[
    title=\textcolor{black}{#1}
    arc=0mm, 
    bottomtitle=0.5mm,
    boxrule=0mm,
    colbacktitle=black!10!white, 
    coltitle=black,
    colframe=orange, 
    fonttitle=\bfseries, 
    fonttitle=\small,
    fontupper=\small,
    left=2.5mm,
    leftrule=1mm,
    right=3.5mm,
    title={#1},
    toptitle=0.75mm
    ]
}
{
    \end{tcolorbox}
}

\NewDocumentEnvironment{magic}{O{Title not defined}}
{
    \begin{tcolorbox}[colback=boxi_yellow, colframe=boxi_gray, colbacktitle=boxi_gray, 
    fonttitle=\small,
    fontupper=\small,
    boxsep=1pt,
    left=4pt, right=2pt, top=6pt, bottom=2pt,
    title=\textcolor{black}{#1}]
    \small %
    \begin{checklist}
}
{
    \end{checklist}
    \end{tcolorbox}
}

\newlist{checklist}{itemize}{1}
\setlist[checklist]{label=\tickbox, leftmargin=*, itemindent=1pt}

\newlist{todolist}{itemize*}{2}
\setlist[todolist]{label=$\square$}

\usepackage[colorlinks=false, linkcolor=gray, citecolor=gray, urlcolor=gray]{hyperref}

\definecolor{MK_Two_One}{RGB}{178,24,43} %
\definecolor{MK_Two_Two}{RGB}{239,138,98}
\definecolor{MK_Two_Three}{RGB}{253,219,199}
\definecolor{MK_Two_Four}{RGB}{209,229,240}
\definecolor{MK_Two_Five}{RGB}{103,169,207}
\definecolor{MK_Two_Six}{RGB}{33,102,172} %

\newenvironment{leftenumerate}
  {%
    \begin{list}{\arabic{enumi}.}{%
      \usecounter{enumi}%
      \setlength{\leftmargin}{0pt}%
      \setlength{\itemindent}{15pt}%
      \setlength{\labelwidth}{0pt}%
      \setlength{\labelsep}{5pt}%
      \setlength{\itemsep}{0.0pt}%
      \setlength{\topsep}{0.0pt}%
    }%
  }
  {%
    \end{list}%
  }

\hypersetup{
colorlinks=true
,linkcolor=MK_Two_Six
,citecolor=MK_Two_Six
,filecolor=MK_Two_Six
,urlcolor= MK_Two_Six
,menucolor=MK_Two_Five
,runcolor=MK_Two_Four
,linkbordercolor=MK_Two_One
,citebordercolor=MK_Two_Two
,filebordercolor=MK_Two_Three
,urlbordercolor=MK_Two_Six
,menubordercolor=MK_Two_Five
,runbordercolor=MK_Two_Four
}

\usepackage[noabbrev,capitalise]{cleveref}

\usepackage[abbreviations]{glossaries-extra}

\glssetcategoryattribute{abbreviation}{indexonlyfirst}{true}

\glssetcategoryattribute{abbreviation}{nohyperfirst}{true}

\newabbreviation{ate}{ATE}{Absolute Trajectory Error}
\newabbreviation{ape}{APE}{Absolute Position Error}

\newabbreviation{auroc}{AUROC}{Area Under the Receiver Operating Characteristic Curve}
\newabbreviation{accuracy}{Acc}{Accuracy}

\newabbreviation{cnn}{CNN}{Convolutional Neural Network}

\newabbreviation{dof}{DoF}{Degrees of Freedom}

\newabbreviation{fov}{FoV}{Field of View}
\newabbreviation{fpr}{FPR}{False Positive Ratio}
\newabbreviation{fem}{FEM}{Finite Element Method}

\newabbreviation{gmsf}{GMSF}{Graph-based Multi-Sensor Fusion}
\newabbreviation{gnn}{GNN}{Graph Neural Network}
\newabbreviation{gcn}{GCN}{Graph Convolutional Network}
\newabbreviation{gnss}{GNSS}{Global Navigation Satellite System}
\newabbreviation{imu}{IMU}{Inertial Measurement Unit}
\newabbreviation{irl}{IRL}{Inverse Reinforcement Learning}

\newabbreviation{knn}{KNN}{K-Nearest Neighbors}

\newabbreviation{lagr}{LAGR}{Learning Applied to Ground Vehicles}
\newabbreviation{lidar}{LiDAR}{Light Detection and Ranging}
\newabbreviation{lio}{LIO}{LiDAR Interial Odometry}

\newabbreviation{mlp}{MLP}{Multi-Layer Perceptron}
\newabbreviation{mpc}{MPC}{Model Predictive Controller}
\newabbreviation{mse}{MSE}{Mean Squared Error}

\newabbreviation{ood}{OOD}{out-of-distribution}

\newabbreviation{ptp}{PTP}{Precision Time Protocol}
\newabbreviation{ppp}{PPP}{Precise Point Positioning}

\newabbreviation{rbf}{RBF}{Radial Basis Function}
\newabbreviation{rmp}{RMP}{Riemannian Motion Policies}
\newabbreviation{ros}{ROS}{Robot Operating System}
\newabbreviation{ros1}{ROS~1}{Robot Operating System}
\newabbreviation{roc}{ROC}{Receiver Operating Characteristic}
\newabbreviation{rf}{RF}{Random Forest}
\newabbreviation{rts}{RTS}{Robotic Total Station}
\newabbreviation{rte}{RTE}{Relative Trajectory Error}
\newabbreviation{rpe}{RPE}{Relative Position Error}
\newabbreviation[shortplural=TPS]{tps}{TPS}{total station}

\newabbreviation{sdf}{SDF}{Signed Distance Field}
\newabbreviation{slam}{SLAM}{Simultaneous Localization and Mapping}
\newabbreviation{svm}{SVM}{Support Vector Machine}
\newabbreviation{svc}{SVC}{Support Vector Classifier}
\newabbreviation{wvn}{WVN}{Wild Visual Navigation}

\newabbreviation{vit}{ViT}{Vision Transformer}
\newabbreviation{vio}{VIO}{Visual Interial Odometry}

\setkeys{glslink}{hyper=false}

\newcommand{\na}{\rule[3.0pt]{0.5cm}{0.5pt}}

\makeatletter
\DeclareRobustCommand\onedot{\futurelet\@let@token\@onedot}
\def\@onedot{\ifx\@let@token.\else.\null\fi\xspace}
\makeatother
\newcommand{\eg}{e.g\onedot}

\definecolor{customgreen}{RGB}{13, 109, 10}
\definecolor{customred}{RGB}{165, 0, 6}

\newcommand{\cmark}{{\color{customgreen}\ding{52}}}%
\newcommand{\xmark}{{\color{customred}\ding{56}}}%

\newcommand{\alias}[1]{\textcolor{gray}{(\texttt{#1})}}
\newcommand{\dataset}[1]{\textcolor{gray}{\emph{#1}}}
\newcommand{\cam}[2]{\textcolor{gray}{\texttt{#1-#2}}}

\newcommand{\ra}[1]{\renewcommand{\arraystretch}{#1}}

\begin{document}
\title{\textit{Boxi: }Design Decisions in the Context of Algorithmic Performance for Robotics}

\ifthenelse{\boolean{anonymous}}{
    \author{Author Names Omitted for Anonymous Review. Paper-ID [93]}
}{
    \author{Jonas Frey$^{\star,1,3}$ \hspace{4pt} Turcan Tuna$^{\star,1}$ \hspace{4pt} Lanke Frank Tarimo Fu$^{\star,2}$ \hspace{4pt} Cedric Weibel$^1$ \hspace{4pt} \hspace{4pt}  Katharine Patterson$^1$ \\ Benjamin Krummenacher$^{1}$ \hspace{2pt}  Matthias Müller$^{1}$ \hspace{2pt}  Julian Nubert$^{1,3}$ \hspace{2pt}  Maurice Fallon$^2$ \hspace{2pt} Cesar Cadena$^1$ \hspace{2pt} Marco Hutter$^1$\\[4pt]
$^1$\,ETH Zurich \hspace{12pt} $^2$\,University of Oxford \hspace{12pt} $^3$\,Max Planck Institute for Intelligent Systems\\[2pt]
$^\star$\,Equal contribution. Contact: \texttt{jonfrey@ethz.ch}
}
}

\twocolumn[{%
\renewcommand\twocolumn[1][]{#1}%
\maketitle
\begin{center}
    \vspace{-5pt}
    \centering
    \includegraphics[width=1.0\linewidth]{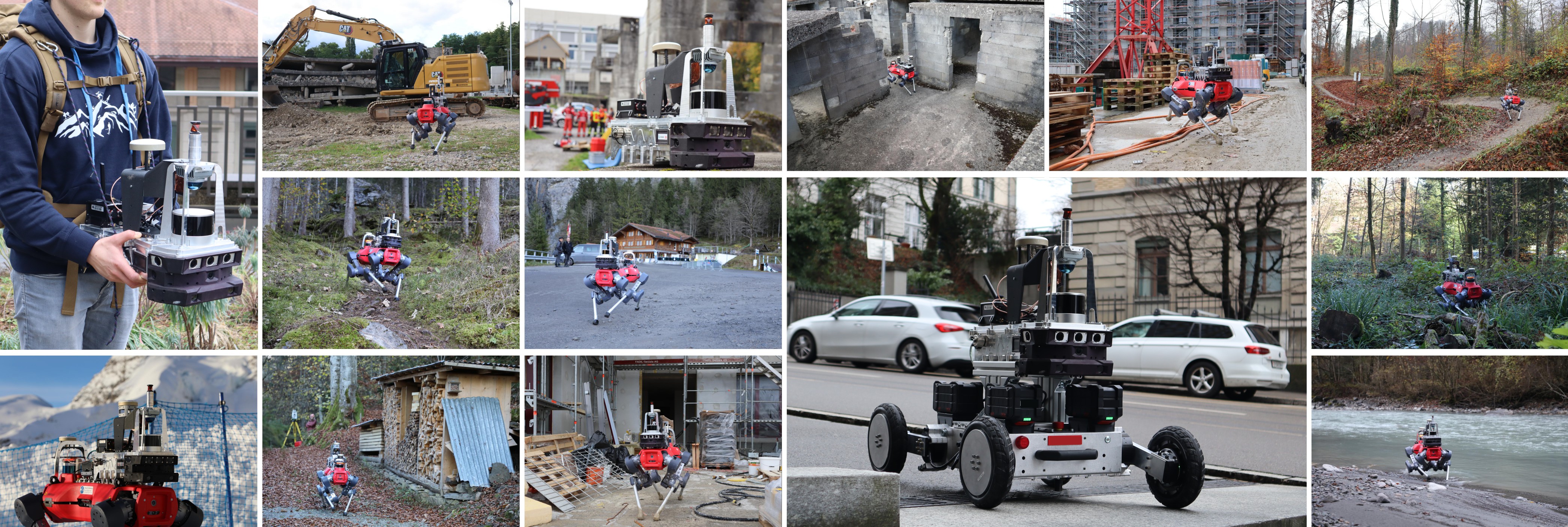}
    \captionof{figure}{\textit{Boxi} can be deployed handheld, on a legged robot, or a wheeled robot across a variety of environments.}
    \label{fig:header}
\end{center}%
}]

\begin{abstract}
Achieving robust autonomy in mobile robots operating in complex and unstructured environments requires a multimodal sensor suite capable of capturing diverse and complementary information. However, designing such a sensor suite involves multiple critical design decisions, such as sensor selection, component placement, thermal and power limitations, compute requirements, networking, synchronization, and calibration. While the importance of these key aspects is widely recognized, they are often overlooked in academia or retained as proprietary knowledge within large corporations. To improve this situation, we present \textit{Boxi}, a tightly integrated sensor payload that enables robust autonomy of robots in the wild. This paper discusses the impact of payload design decisions made to optimize algorithmic performance for downstream tasks, specifically focusing on state estimation and mapping. \textit{Boxi} is equipped with a variety of sensors: two LiDARs, 10 RGB cameras including high-dynamic range, global shutter, and rolling shutter models, an RGB-D camera, 7 inertial measurement units (IMUs) of varying precision, and a dual antenna RTK GNSS system. Our analysis shows that time synchronization, calibration, and sensor modality have a crucial impact on the state estimation performance. We frame this analysis in the context of cost considerations and environment-specific challenges. We also present a mobile sensor suite `cookbook` to serve as a comprehensive guideline, highlighting generalizable key design considerations and lessons learned during the development of \textit{Boxi}. Finally, we demonstrate the versatility of \textit{Boxi} being used in a variety of applications in real-world scenarios, contributing to robust autonomy. More details and code:
\url{https://github.com/leggedrobotics/grand_tour_box}
\end{abstract}

\IEEEpeerreviewmaketitle
\section{Introduction}
\label{sec:introduction}

\begin{table*}[t!]
\footnotesize %
\centering %
\resizebox{\textwidth}{!}{%
\begin{threeparttable}
\begin{tabular}{l c c c c c c c c c c c c c c c } 
\toprule
\textbf{System Name} & \multicolumn{3}{c}{\textbf{Cameras}} & \multicolumn{2}{c}{\textbf{Depth}} & \multicolumn{3}{c}{\textbf{IMUs}} & \textbf{ Other Sensors} & \textbf{GNSS Solution} & \textbf{Reference}  & \textbf{Sync} & \textbf{Robot} \\
\cmidrule(r){2-4} \cmidrule(r){5-6} \cmidrule(r){7-9}
 & GS & RS & HDR & Stereo & LiDAR & Nav. & Tact. & Co. &  &  &  &  &  &  \\

\midrule
SubT-MRS\cite{zhao2024subt}  & 
4x \cmark  & %
\xmark  & %
\xmark  & %
\xmark  &  %
2x \cmark  &  %
\xmark  &  %
\xmark  &  %
2x \cmark  &  %
FLIR Boson &  %
RTK  &  %
Scanner  &  %
HW, SW& %
Versatile %
\\

CatPack\cite{hudson2021heterogeneous, ramezani2022wildcat}  & 
\xmark & %
4x \cmark & %
\xmark & %
\xmark & %
1x \cmark & %
\xmark & %
\xmark & %
1x \cmark & %
FLIR Lepton 3.5 thermal & %
\xmark & %
- & %
HW, SW & %
Legged %
\\

VersaVIS~\cite{tschopp2020versavis} & 
2x \cmark & %
2x \cmark & %
\xmark & %
1x\cmark & %
1x \cmark & %
\xmark & %
\xmark & %
3x \cmark & %
Thermal& %
\xmark & %
- & %
HW &%
Versatile %
\\

Shepard`s Stick\cite{jelavic2021towards} & 
1x \cmark & %
2x \cmark & %
\xmark & %
1x \cmark & %
1x \cmark & %
\xmark& %
\xmark & %
2x \cmark & %
- & %
\xmark & %
- & %
SW & %
Handheld %
\\

FusionPortable\cite{jiao2022fusionportable} & 
2x \cmark & %
2x \cmark & %
\xmark & %
\xmark & %
1x \cmark & %
\xmark & %
1x \cmark & %
3x \cmark & %
2x Event Camera & %
RTK & %
RTK/MOCAP/Scanner & %
SW & %
Versitale %
\\

Phasma\cite{helmberger2022hilti} & 
5x \cmark & %
1x \cmark & %
\xmark & %
\xmark & %
1x \cmark & %
\xmark & %
\xmark & %
1x \cmark & %
- & %
\xmark & %
Scanner & %
HW, FPGA&%
Hilti %
\\

VBR\cite{brizi2024vbr} & 
2x \cmark & %
\xmark & %
\xmark & %
\xmark & %
1x \cmark & %
\xmark  & %
1x \cmark & %
1x \cmark & %
- & %
RTK & %
Scanner / RTK & %
HW, SW&%
Rome %
\\

Botanic\cite{liu2024botanicgarden} & 
4x \cmark & %
\xmark & %
\xmark & %
\xmark & %
2x \cmark & %
\xmark & %
1x \cmark & %
2x \cmark & %
 & %
\xmark & %
Scanner & %
HW (STM32 MCU)&%
Scout V1.0 %
\\

PIRVS \cite{zhang2018pirvs} & 
2x \cmark & %
\xmark & %
\xmark & %
\xmark & %
\xmark & %
\xmark  & %
\xmark & %
1x \cmark & %
- & %
\xmark & %
- & %
HW&%
Versatile %
\\

(Industrial) BLK2Go \cite{leica_blk2go}& 
 3x \cmark &
 1x \cmark &
 \xmark &
 \xmark &
 1x \cmark&
 \xmark &
\xmark &
 1x \cmark &
 - &
 \xmark &
 no &
 HW  &
 Handheld 
\\

 WHU-Helmet\cite{li2023whu} & 
 \xmark & %
 \xmark & %
 \xmark & %
 \xmark & %
 1x \cmark & %
 1x \cmark &
 \xmark &
 \xmark &
 - &
 RTK &
 GNSS &
 SW &
 Helmet
\\

 EverySync \cite{wu2024everysync} &
 1x \cmark & %
 \xmark & %
 \xmark & %
 \xmark & %
 1x \cmark & %
 \xmark & %
 \xmark & %
 1x \cmark & %
 - & %
 RTK & %
 RTK & %
 HW, SW&%
 Wheeled Robot
\\

 S3E \cite{feng2024s3e} & 
 \xmark & %
 2x \cmark  & %
 \xmark  & %
 \xmark  & %
  1x \cmark & %
 \xmark & %
 \xmark & %
 1x \cmark & %
 UWB & %
 RTK & %
 INS / MOCAP & %
 HW&%
 Wheeled Robot
\\

TartanDrive \cite{triest2022tartandrive, sivaprakasam2024tartandrive} &
2x \cmark & %
\xmark & %
\xmark & %
1x \cmark & %
3x \cmark & %
\xmark  & %
1x \cmark & %
1x \cmark & %
 - & %
RTK & %
RTK, INS & %
SW&%
Offroad Car
\\

MCD \cite{mcdviral2024} &
2x \cmark & %
\xmark & %
\xmark & %
1x \cmark & %
2x \cmark & %
\xmark & %
\xmark & %
3x \cmark & %
UWB & %
\xmark & %
Scanner & %
SW&%
Handheld / Wheeled
\\

\textit{Boxi} (Ours) & 
3x \cmark & %
5x \cmark & %
3x \cmark & %
1x \cmark & %
2x \cmark & %
1x \cmark & %
2x \cmark & %
4x \cmark & %
1x Barometer & %
RTK & %
INS, RTS, GNSS & %
HW, SW&%
Legged Robot %
\\
\bottomrule
\end{tabular}

\begin{tablenotes}\footnotesize
\item[*] IMUs are categorized into Navigation (Nav.), Tactical (Tac.), and Industrial \& Consumer (Co.) grades. Cameras are categorized into global shutter (GS), rolling shutter (RS), and high-dynamic range (HDR).
\end{tablenotes}
\end{threeparttable}
}
\caption{Overview of Existing Sensor Payloads. Symbols: \cmark = present, \xmark = absent. }
\label{table:simplified}
\vspace{-0.5cm}
\end{table*}

Designing a capable perception payload is fundamental to the success of mobile robotics systems. 
The performance of perception algorithms is inherently upper bounded by the quality of the input sensor data, which might be corrupted by aleatoric sensor noise and epistemic design issues, regardless of the applied algorithm. 
Therefore, it is key to select the appropriate sensors while considering time synchronization, sensor calibration, latency, communication, thermal compensation, and hardware design decisions to enable the next generation of mobile robots to operate reliably in complex environments.
Previous studies have emphasized the critical role of time synchronization~\cite{wu2024everysync, tschopp2020versavis, bry2015spatial}, rolling shutter correction~\cite{smid2019rolling}, SLAM performance under various conditions~\cite{maddern2016real, zhang2018pirvs}, and accurate intrinsic and extrinsic calibration~\cite{rehder2016extending, furgale2013kalibr, cramariuc2020learning, hagemann2022inferring, liu2024botanicgarden} in determining downstream task performance.

To this end, we introduce \textit{Boxi}, a compact, standalone sensing payload designed to enable the development of autonomy solutions for mobile robots, with a particular focus on quadrupedal legged robots. \textit{Boxi} integrates best practices in mechanical, electrical, and software engineering, as well as system design, and was co-developed in collaboration with Leica Geosystems. We leverage \textit{Boxi} to establish a direct link between design decisions and algorithmic performance on downstream tasks such as robot localization. \\

The design of \textit{Boxi} is driven by four primary objectives: (1)~developing a state-of-the-art mobile robot sensor suite, (2)~fostering a better understanding between payload design decisions and algorithmic performance, (3)~allowing the collection of large scale multi-sensor datasets with high-quality calibration and time synchronization and (4)~enabling autonomous mobile navigation.
Based on these objectives, we equipped \textit{Boxi} with accessible state-of-the-art sensors featuring two spinning LiDARs (repetitive and non-repetitive), 10 cameras (high-dynamic range, global shutter, and rolling shutter models), a stereo-based depth camera, 7 inertial measurement units (IMUs) of varying precision and cost (\SI{5}{\USD} to \SI{12000}{\USD}), as well as a dual antenna RTK-GNSS system \cite{novatel_cpt7} and tight integration to Leica Geosystems MS60 Total Station (Reflector Prism \& Receiver Link).
Compared to autonomous driving payloads, \textit{Boxi} is more compact and lightweight, but shares similar design requirements provided in~\cite{geiger2013vision}. It is based on the principle that payload design must consider algorithmic performance~\cite{zhang2018pirvs}. To support and extend this notion to general robotics, we present a sensory payload \textit{'cookbook'} that summarizes design decisions and lessons learned, enabling the development of future perception systems beyond \textit{Boxi}.

Further, we showcase how \textit{Boxi}, being \textit{SLAM-ready}~\cite{zhang2018pirvs}, in addition, can be classified as \textit{autonomy-ready} by performing onboard state estimation, mapping, and navigation.
At the core of this work, we use \textit{Boxi} to conduct a rigorous study to investigate the state estimation performance across multiple dimensions (\eg, accuracy, robustness to time, and extrinsic calibration) with respect to hardware and software design decisions in 7 different environments. 

We begin by analyzing the performance of different sensor modalities using popular and state-of-the-art approaches for each modality (LiDAR, Camera, Proprioception, and GNSS). 
We then compare different camera types (high dynamic range, global shutter, and rolling shutter) in terms of image quality and \gls{vio} performance.
In addition, we compare the performance impact of IMUs (\SIrange{5}{12000}{USD}) on dead-reckoning and \gls{lio} performance. 
Lastly, we simulate time synchronization and extrinsic calibration errors to investigate the effect on state estimation.
Through the steps described above, we will address the following research question: \textit{Which sensors should be purchased given the mapping accuracy requirements, budget constraints, and deployment environment, and how important are extrinsic and intrinsic calibration, as well as time synchronization, in achieving the desired accuracy?}
To foster further research, the full hardware and software design is made available open-source.

In summary, the main contributions of this work are:  
\begin{itemize}
    \setlength{\itemsep}{2pt}
    \item Design, development, and evaluation of \textit{Boxi}, a versatile multi-modal sensor payload optimized for data collection on mobile robotic platforms.
    \item Ablation studies on the effect of modalities, time synchronization accuracy, and extrinsic calibration accuracy across multiple real-world environments.
    \item Compiling a comprehensive \textit{cookbook}, featuring insights, lessons learned, and best practices for designing payloads for robotic platforms.\looseness-1
\end{itemize}

\section{Related Works}
\label{sec:related}
\begin{figure*}[!t] 
    \centering 
    \includegraphics[width=1.0\linewidth]{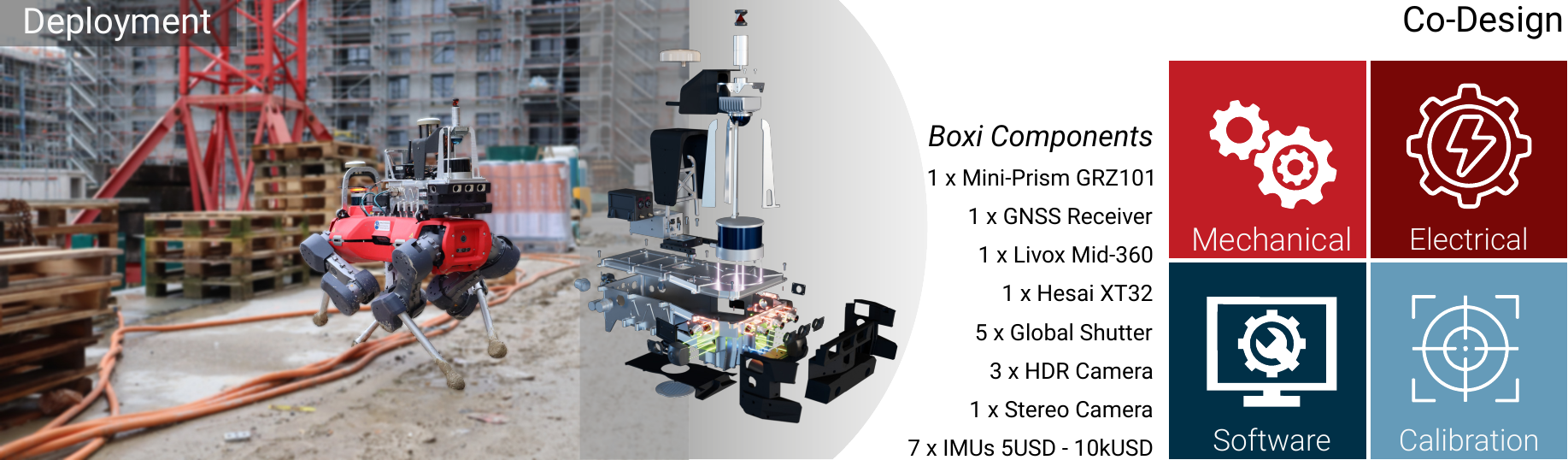} 
    \caption{Co-design of hardware, electrical systems, software, and calibration is essential for developing a robust sensor payload.}
    \label{fig:components} 
\end{figure*}
\subsection{Payloads for Autonomous Driving}
The development of mobile mapping systems initially focused on integration into wheeled platforms due to their capacity to accommodate heavy sensors and their ability to drive long distances~\cite{geiger2013vision}. 
These systems often incorporated a combination of cameras, LiDAR sensors, IMU, and GNSS---a sensor suite that continues to form the foundation of our work. 
One of the milestones in the development of autonomous driving was marked by the introduction of the KITTI dataset~\cite{geiger2013vision}, which featured comprehensive data from cameras, LiDAR, IMU, and RTK GNSS, and which continues to be relevant in the field today. 
Subsequent data collection platforms focused on semantic understanding~\cite{cordts2015cityscapes} and temporal changes~\cite{maddern2017}. 
While the previous works were primarily driven by academia, multiple industry-driven sensory payloads were used to release high-quality datasets, such as NuScenes~\cite{caesar2020nuscenes}, Waymo~\cite{sun2020scalability}, and Argoverse~\cite{chang2019argoverse}. These datasets often feature similar sensor configurations; however, detailed information on calibration, time synchronization, and sensor payload design is not publicly available.
More recent efforts have introduced modern sensors such as \SI{360}{\degree} \gls{fov} radar~\cite{burnett2023boreas} or event cameras~\cite{gehrig2021dsec}.\looseness-1

\subsection{Compact Vision Payloads}
Within the last decade, a variety of more mobile and lightweight payloads based on multi-camera setups were introduced. The authors in~\cite{schmid2013stereo} introduced the DLR~7, a stereo camera system with FPGA-based time synchronization, enabling handheld operation at just \SI{800}{\gram}. Similarly,~\citet{nikolic2014synchronized} developed an ultra-lightweight unit integrating stereo cameras and an IMU with FPGA-based synchronization and exposure time compensation, tailored for online visual feature extraction.
\citet{zhang2018pirvs} presented the PIRVS system, which directly integrated the SLAM algorithms on-board, while considering the latency between the image exposure period and CPU processing. Expanding on these designs,~\citet{tschopp2020versavis} introduced VersaVIS, an open, versatile multi-camera and IMU sensor suite with precise hardware synchronization, which later evolved into the commercial Sevensense CoreResearch~\cite{sevensense_coreresearch}. Although all previous work highlights the importance of calibration and time synchronization and includes FPGAs or ISPs for processing or triggering, no systematic experimental evaluation of their impact on state estimation performance is provided.\looseness-1

\subsubsection{Compact Multi-Modal Sensor Payloads}
The development of more compact multi-modal sensor suites has been of interest in industry and research ~\cite{tao2024oxford, wu2024everysync, ramezani2022wildcat} driven by their promising potential to enhance precision and resilience in challenging environments. An industrial example is the Leica Geosystems BLK2Go~\cite{blk2go}, a tightly integrated handheld scanning device, which is equipped with three global shutter cameras, an IMU, and an \SI{830}{\nano\meter} laser scanner (420k points/s) that provides measurements up to \SI{3}{\milli\meter} accuracy.

EverySync~\cite{wu2024everysync} is a multi-sensor payload focusing on time synchronization accuracy. EverySync relies on the methods developed in VersaVIS~\cite{tschopp2020versavis} to perform triggering and time synchronization. Moreover, the authors of EverySync provided an extensive analysis of accurate time synchronization and achieved a time synchronization accuracy of \SI{1}{\milli\second} between GNSS, visual, and LiDAR sensors. Similarly, LiDAR-Visual and Inertial payloads~\cite{helmberger2022hilti, ramezani2022wildcat, jelavic2021towards} showed the importance of time synchronization and extrinsic calibration between modalities for SLAM and state estimation purposes.~\citet{liu2024botanicgarden} integrated and demonstrated a multi-LiDAR, multi-camera, and multi-IMU setup with precise time synchronization up to \SI{1}{\micro\second} using \gls{ptp} and a self-designed STM32 MCU triggering board. 
Recently,~\citet{mcdviral2024} integrated a sensor payload and proposed using a continuous time B-spline trajectory representation for reference pose estimation. They showed that this formulation mitigates the impact of time synchronization errors for SLAM and state estimation evaluation.

\section{Overview}
\label{sec:overview}
\begin{table*}[h]
\centering
\resizebox{\textwidth}{!}{%
\begin{tabular}{llllr}
\toprule
\textbf{ } & \textbf{Name} & \textbf{Description} & \textbf{Power} & \textbf{est. Price} \\
\midrule
\multicolumn{5}{l}{\textbf{LiDARs}} \\
\alias{Livox} & Livox Mid-360 & vFoV: \SI{59}{\degree},\quad Range: \SI{0.1}{\meter},\quad \: \SI{40}{\meter},\quad \:  \SI{10}{\hertz},\quad Accuracy: $\pm$\SI{0.02}{\meter} & \SI{6.5}{\watt} (avg) & \SI{749}{\USD} \\
\alias{Hesai} & Hesai XT-32 & vFoV: \SI{31}{\degree},\quad  Range: \SI{0.05}{\meter},\quad \SI{120}{\meter},\quad  \SI{10}{\hertz},\quad Accuracy: $\pm$\SI{0.01}{\meter}  & \SI{10}{\watt} (avg) & \SI{3000}{\USD} \\
\midrule
\multicolumn{5}{l}{\textbf{Cameras}} \\

\alias{CoreResearch} & Sevensense CoreResearch & 1440$\times$1080 @ \SI{10}{\fps},\quad FoV: \SI{126}{\degree}$\times$\SI{92.4}{\degree},\quad \SI{71.6}{\decibel},\quad GS,\quad \SI{1.6}{\megapixel} & \SI{12}{\watt} (max) & \SI{2000}{\USD} \\
\alias{HDR} & TierIV C1 & 1920$\times$1280 @ \SI{30}{\fps},\quad FoV: \SI{120}{\degree}$\times$\SI{80.0}{\degree},\quad \SI{120}{\decibel},\quad RS,\quad \SI{2.5}{\megapixel} & \SI{1.7}{\watt} (max) & \SI{900}{\USD} \\
\alias{ZED2i} & Stereolabs ZED2i & 1920$\times$1080 @ \SI{30}{\fps},\quad FoV: \SI{126}{\degree}$\times$\SI{92.4}{\degree},\quad \SI{64.6}{\decibel},\quad RS,\quad \SI{4}{\megapixel} & \SI{2}{\watt} (max) & \SI{500}{\USD} \\
\midrule
\multicolumn{5}{l}{\textbf{Reference}} \\
\alias{AP20}$^*$ & Modified Leica AP20 & MS60 Total Station Bluetooth Communication \& Synchronization & \na & \SI{5000}{\USD} \\
\alias{Prism} & Mini-Prism GRZ101  & Static Accuracy $\pm$\SI{2}{\milli\meter} (3$\sigma$),\quad hFoV \SI{360}{\degree},\quad vFoV $\pm$\SI{30}{\degree} & \rule[3.0pt]{0.5cm}{0.5pt} & \SI{1000}{\USD} \\

\alias{CPT7} & Novatel Span CPT7 & Dual RTK GPS, Intertial Explorer, TerraStar & \SI{18}{\watt}(max) & \SI{40000}{\USD} \\
 & 3GNSSA-XT-1 & Multiband GPS Antenna [BeiDou, L-Band, GLONASS, GPS, Galileo, Navic] & \SI{1}{\watt} (each) &  \\
\midrule
\multicolumn{5}{l}{\textbf{IMUs}} \\
\alias{HG4930} & Honeywell HG4930 & \SI{100}{\hertz},\quad \SI[parse-numbers = false]{\pm400}{\degree\per\second},\quad \: \SI[parse-numbers = false]{\pm20}{\ggrav}, GBI \SI{0.25}{\degree\per\hour},\quad ABI \SI{0.025}{\milli\gram} & \SI{2}{\watt} (max) & \SI{12000}{\USD} \\
\alias{STIM320} & Safran STIM320 & \SI{500}{\hertz},\quad \SI[parse-numbers = false]{\pm400}{\degree\per\second},\quad \: \SI[parse-numbers = false]{\pm5}{\ggrav},\quad GBI \SI{0.3}{\degree\per\hour},\quad & \SI{2}{\watt} (max) & \SI{8000}{\USD} \\
\alias{AP20-IMU} & Proprietary IMU (AP20) & \SI{200}{\hertz}  & \na & Mid-Range \\
\alias{ADIS} & Analog Devices ADIS16475-2 & \SI{200}{\hertz},\quad \SI[parse-numbers = false]{\pm500}{\degree\per\second},\quad \: \SI[parse-numbers = false]{\pm8}{\ggrav},\quad GBI \SI{2.5}{\degree\per\hour},\quad ABI \SI{0.0036}{\milli\gram} & \na & \SI{500}{\USD} \\
\alias{Bosch} & Bosch BMI085 (CoreResearch) & \SI{200}{\hertz},\quad \SI[parse-numbers = false]{\pm2000}{\degree\per\second},\quad \SI[parse-numbers = false]{\pm16}{\ggrav} & \na & \SI{5}{\USD} \\
\alias{TDK} & TDK ICM40609 (Livox) & \SI{200}{\hertz},\quad \SI[parse-numbers = false]{\pm2000}{\degree\per\second},\quad \SI[parse-numbers = false]{\pm32}{\ggrav} & \na & \SI{5}{\USD} \\
\alias{ZED2i-IMU} & Unknown (ZED2i) & \SI{400}{\hertz},\quad \SI[parse-numbers = false]{\pm1000}{\degree\per\second},\quad \SI[parse-numbers = false]{\pm8}{\ggrav}  & \na & \na \\

\bottomrule
\end{tabular}
}
\caption{\textit{Boxi} Components: GS=GlobalShutter, RS=RollingShutter. Used camera settings are provided under their respective sections. ABI=Accelerometer Bias Instability, GBI=Gyro Bias Instability. $^*$ The \alias{AP20} requires a Leica MS60 Total Station at a cost of \SI{50000}{\USD}.}
\label{tab:components}
\end{table*}

\subsection{How to read the paper}
This paper is structured as follows. \Cref{subsec:overview} details \textit{Boxi}'s key specifications, providing the foundation for the algorithmic performance analysis provided in \Cref{sec:algo}, which is self-contained and highlights the most relevant experiments and findings. Finally, \Cref{sec:cookbook} serves as a comprehensive \emph{cookbook}, describing the design of \textit{Boxi} and presenting key lessons learned as well as generalizable findings for the design of future sensor payloads.

\subsection{Boxi Overview}
\label{subsec:overview}
\textit{Boxi} is designed to enable autonomy and provide a comprehensive understanding of the payload design decisions on downstream task performance. This requires the integration of different sensing modalities and sensor types.  
Therefore, \textit{Boxi} includes exteroceptive sensors such as two LiDARs and a stereo camera setup for geometric perception. In addition, the payload is equipped with high-dynamic range, global shutter, and rolling shutter cameras for visual perception, and various tiers of IMUs for inertial sensing. 
\textit{Boxi} provides time synchronization accuracy ranging between \SI{5}{\milli\second} to \SI{10}{\micro\second} (sensor-dependent \cref{subsec:communication}), and accurate extrinsic calibrations (cam-to-cam \SI{0.3}{\milli\meter}@1$\sigma$, cam-to-LiDAR \SI{1.0}{\milli\meter} corner detection accuracy) between all cameras and LiDARs.
Precise extrinsic calibration is achieved by CNC machining \textit{Boxi} out of an aluminum mono-block with a manufacturing precision of \SI{0.05}{\milli\meter} in combination with state-of-the-art calibration algorithms (customized version of Kalibr~\cite{rehder2016extending, furgale2013kalibr}, IMU intrinsic~\cite{AllanVarianceRos}, LiDAR extrinsic~\cite{diffcal}).\looseness-1

The cameras and LiDARs on \textit{Boxi} are carefully positioned to capture the near-field surroundings of the robot, which is crucial for locomotion and navigation as opposed to a reality capture setup. %
The complete sensor payload weighs \SI{7.1}{\kilo\gram}, making it suitable for deployment with a harness or with adaptable mounting options that can be deployed on different robotic platforms.
The sensor arrangement is optimized for near-field perception (stereo camera baseline of \SI{12}{\centi\meter}) and long-range using a high-accuracy repetitive LiDAR.
Throughout this work, each sensor will be referred to by its abbreviation outlined in \cref{tab:components}.\looseness-1
\section{Algorithmic Performance}
\label{sec:algo}

\subsection{Real-World Datasets}
\begin{figure*}[!t] 
    \centering 
    \includegraphics[width=1.0\linewidth]{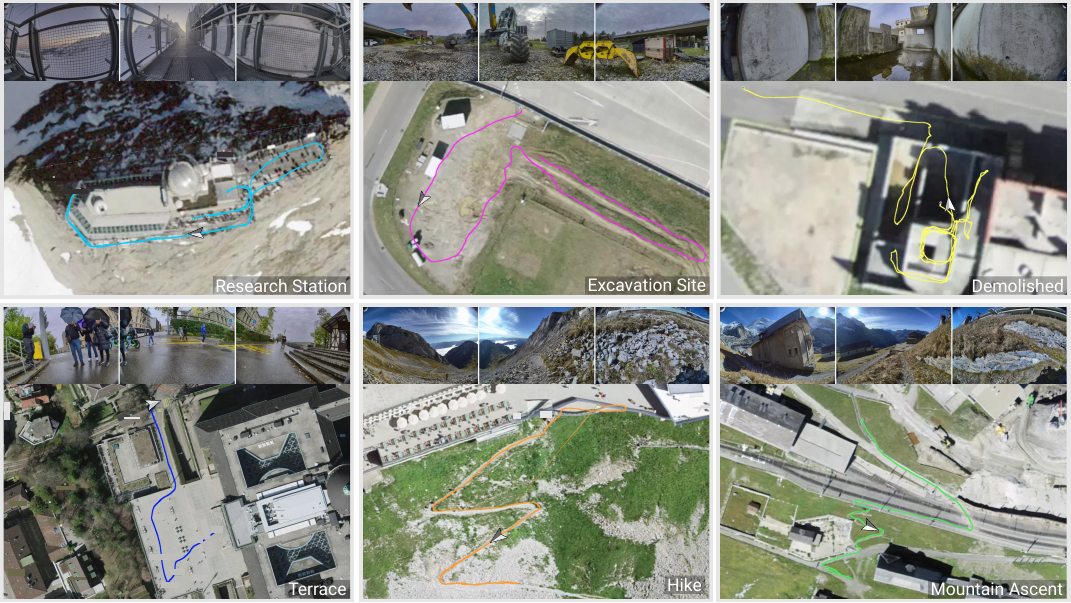} 
    \caption{Subset of the GrandTour Datasets Overview: (top-row) left, front, and right HDR camera images; (bottom-row) satellite image with projected GNSS path and camera image viewpoint; The warehouse dataset is not displayed.}
    \label{fig:deployment_envs} 
\end{figure*}
We evaluated our hardware and software decisions on data from seven real-world environments collected in the wild (\cref{fig:deployment_envs}), which are part of the GrandTour dataset. All data were gathered by the legged robot ANYmal~\cite{anymal} equipped with \textit{Boxi} and teleoperated by an expert human. The selected scenarios are diverse in lighting and include indoor and outdoor settings, confined spaces, wide open spaces, and both structured and unstructured scenes. The deployments take between \SI{3}{\min}-\SI{7}{\min} and cover distances around \SI{300}{\meter} and more details are available in \cref{app:dataset_desc}.\looseness-1

\subsection{Algorithms}
As a representative of multi-camera visual-inertial state estimation, we use OKVIS2~\cite{leutenegger2022okvis2}, which is configured to perform loop closures. Moreover, a customized version of the state-of-the-art LiDAR-inertial geometric-observer based odometry pipeline Direct LiDAR Inertial Odometry (DLIO)~\cite{dlio} is used to represent tightly coupled LiDAR-inertial methods. In addition, we use the kinematic-inertial estimator TSIF~\cite{bloesch2017two} to represent the performance of proprioceptive state-estimation methods. Alongside these modules, the post-processed GNSS-IMU solution of the Inertial Explorer~\cite{novatel_ie} is provided as an industrial positioning solution for outdoor datasets. The summary of the algorithms is provided in \Cref{tab:state_estimation_summary}.
\begin{table}[h]
    \centering
    \ra{1.2}
    \resizebox{1\columnwidth}{!}{
    \begin{tabular}{lcc}
        \toprule
        \textbf{Algorithm} & \textbf{Modality} & \textbf{Description} \\
        \midrule 
        OKVIS2~\cite{leutenegger2022okvis2} & Visual-Inertial & Multi-Camera VIO with Loop Closure \\
        DLIO~\cite{dlio} & LiDAR-Inertial & Tightly coupled Filter-based LIO \\
        TSIF~\cite{bloesch2017two} & Kinematic-Inertial & Proprioceptive Filter-based \\
        Inertial Explorer~\cite{novatel_ie}  & GNSS-Inertial & Industrial offline INS \\
        \bottomrule
    \end{tabular}
    }
    \caption{Summary of the State Estimation Algorithms}
    \label{tab:state_estimation_summary}
\end{table}

\subsection{State Estimation - Effect of Modality}
\label{subsec:app_state}
\begin{table}[b]
\centering
\resizebox{1.0\columnwidth}{!}{
\begin{tabular}{lcccc}
\toprule
\textbf{Mission Name} & \textbf{Kinematic}~\cite{bloesch2017two} & \textbf{LiDAR}~\cite{dlio} & \textbf{GNSS}~\cite{novatel_ie} & \textbf{Camera}~\cite{leutenegger2022okvis2} \\ 
\midrule
Terrace & 0.009$\;\pm\;$0.016 & 0.02$\;\pm\;$0.016 & 0.003$\;\pm\;$0.003 & 0.033$\;\pm\;$0.031 \\ 
Research Station  & 0.007$\;\pm\;$0.012 & 0.021$\;\pm\;$0.017 & 0.008$\;\pm\;$0.010 & 0.027$\;\pm\;$0.046 \\ 
Mountain Ascent & 0.016$\;\pm\;$0.040 & 0.015$\;\pm\;$0.013 & 0.004$\;\pm\;$0.003 & 0.025$\;\pm\;$0.021 \\ 
Hike & 0.024$\;\pm\;$0.031 & 0.023$\;\pm\;$0.019 & 0.004$\;\pm\;$0.003 & 0.017$\;\pm\;$0.014 \\ 
Excavation Site & 0.014$\;\pm\;$0.019 & 0.020$\;\pm\;$0.014 & 0.003$\;\pm\;$0.002 & 0.038$\;\pm\;$0.044 \\ 
Demolished Building & 0.017$\;\pm\;$0.019 & 0.015$\;\pm\;$0.009 & 0.025$\;\pm\;$0.040 & 0.031$\;\pm\;$0.032 \\ 
Warehouse & 0.016$\;\pm\;$0.047 & 0.024$\;\pm\;$0.023 & nan & 0.045$\;\pm\;$0.042 \\ 
\midrule
\textbf{Average} & \textbf{0.015} & 0.02 & \textbf{0.008} & 0.031 \\ 
\bottomrule
\end{tabular}
}
\caption{Modality Comparison in RTE per dataset.}
\label{tab:modality_rte}
\end{table}
\begin{table}[ht]
\centering
\resizebox{1.0\columnwidth}{!}{
\begin{tabular}{lcccc}
\toprule
\textbf{Mission Name} & \textbf{Kinematic}~\cite{bloesch2017two} & \textbf{LiDAR}~\cite{dlio} & \textbf{GNSS}~\cite{novatel_ie} & \textbf{Camera}~\cite{leutenegger2022okvis2} \\ 
\midrule
Terrace & 0.309$\;\pm\;$0.113 & 0.023$\;\pm\;$0.015 & 0.01$\;\pm\;$0.004 & 0.524$\;\pm\;$0.208 \\ 
Research Station& 0.154$\;\pm\;$0.091 & 0.025$\;\pm\;$0.012 & 0.138$\;\pm\;$0.08 & 0.198$\;\pm\;$0.104 \\ 
Mountain Ascent & 0.589$\;\pm\;$0.368 & 0.046$\;\pm\;$0.019 & 0.016$\;\pm\;$0.006 & 0.563$\;\pm\;$0.184 \\ 
Hike & 1.353$\;\pm\;$0.559 & 0.038$\;\pm\;$0.019 & 0.059$\;\pm\;$0.021 & 0.243$\;\pm\;$0.105 \\ 
Excavation Site & 0.360$\;\pm\;$0.284 & 0.021$\;\pm\;$0.012 & 0.016$\;\pm\;$0.007 & 0.372$\;\pm\;$0.212 \\ 
Demolished Building & 0.266$\;\pm\;$0.110 & 0.019$\;\pm\;$0.013 & 0.113$\;\pm\;$0.082 & 0.118$\;\pm\;$0.064 \\ 
Warehouse & 0.429$\;\pm\;$0.242 & 0.063$\;\pm\;$0.047 & nan & 0.419$\;\pm\;$0.180 \\ 
\midrule
\textbf{Average} & 0.49 $\;\pm\;$ 0.40 & \textbf{0.03 $\;\pm\;$ 0.02} & 0.06 $\;\pm\;$ 0.06 & 0.35 $\;\pm\;$ 0.17 \\ 
\bottomrule
\end{tabular}
}
\caption{Modality Comparison in ATE per dataset.}
\label{tab:modality_ate}
\end{table}
Different sensors provide complementary information that allows a robot to operate in various environments that are challenging for one or more sensor modalities. 
The goal of this analysis is to present a rough estimate of how different modalities perform across diverse environments, helping to select the most suitable sensor for specific environments and state estimation requirements.
Before interpreting the results, it is crucial to understand that the performance of each modality is highly dependent on the motion pattern (\eg, motion blur, saturation effects) and the environment (\eg, LiDAR degradation, smoke, symmetry, lighting, visual features, GNSS denial). It is not possible to provide fully generalized claims that apply to all possible scenarios, but we can deduce guiding principles.

We compare DLIO~\cite{dlio} (fuses data from Hesai and HG4930), OKVIS2~\cite{leutenegger2022okvis2} (fuses data from CoreResearch-Stereo pair and HG4930), GNSS~\cite{novatel_ie} (IE-TC more details in \cref{app:ground_truth}) and TSIF~\cite{bloesch2017two}.  
\Cref{tab:modality_rte} and \Cref{tab:modality_ate} presents \gls{rte} and \gls{ate} results respectively across all seven tested environments compared against the \gls{tps} measurements.
DLIO, consistently outperforms OKVIS2 and kinematic-based state estimation in terms of \gls{ate}.
With sufficient satellite coverage, the offline-optimized GNSS solution provides highly accurate performance, both locally (\gls{rte}) and globally (\gls{ate}).
In terms of \gls{rte}, kinematic-based state estimation performs well, verifying its common use for control.\looseness-1
\subsection{State Estimation - LiDAR Comparison}
\label{subsec:comparision_lidar}
\begin{figure}[t] 
    \centering 
    \includegraphics[width=1.0\columnwidth]{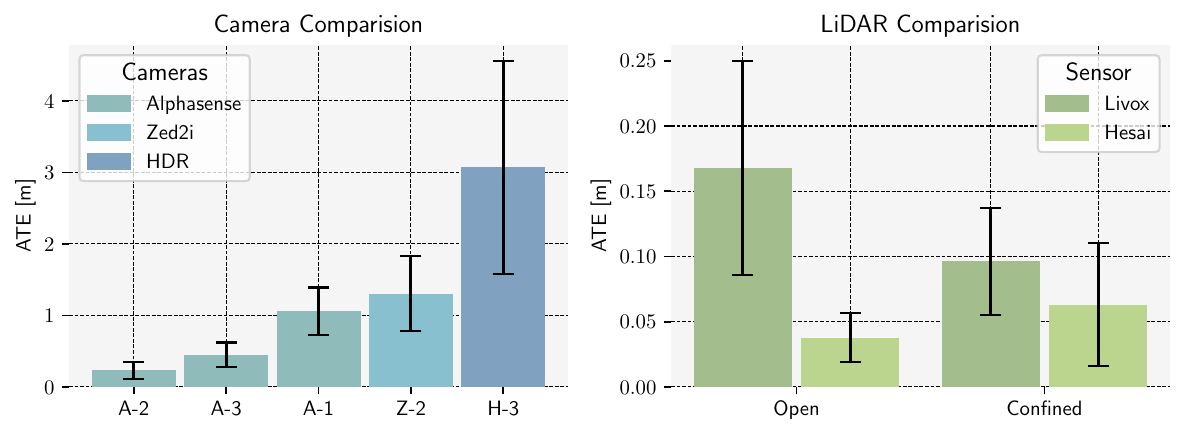} 
    \caption{Camera and LiDAR Comparison using OKVIS2 (HG4930) and DLIO (HG4930). (A-2, A-3, A-1, Z-2, H-3) indicates respective camera configuration (see \cref{subsec:comparision_camera}). The Camera is evaluated on the \emph{Mountain Ascend} dataset while the LiDAR comparison is provided on the \emph{Hike} (Open) and \emph{Warehouse} (Confided) datasets.}
    \label{fig:combined} 
\end{figure}
In the following, all \gls{ape} and \gls{rpe} error metric computations are done with respect to our 6-\gls{dof} ground truth pose estimate, which is obtained by fusing \gls{tps}, IMU, and Inertial Explorer post-processed poses (see \cref{app:ground_truth}).
We test how the Hesai, an accurate long-range LiDAR, compares against the lower-cost and lightweight Livox. 
For this comparison, we selected an open (\emph{Hike}) and confined environment (\emph{Warehouse}) and reported the results in~\cref{fig:combined}. 
The \textit{Hike} dataset contains limited geometric features compared to the \emph{Warehouse} dataset. 
Across both datasets, the Hesai outperforms the Livox. However, in the dataset with a large featureless space, the gap is more significant.\looseness-1

\subsection{State Estimation - Camera Comparison}
\label{subsec:comparision_camera}
In this study, we compare the accuracy of different cameras for VIO using OKVIS2. We evaluated multiple different camera configurations:\looseness-1
\begin{itemize}
    \item CoreResearch: (A-1) Single, (A-2) Stereo, (A-3) Triple
    \item HDR: (H-1) Single, (H-3) Triple
    \item ZED2i: (Z-1) Single, (Z-2) Stereo,
\end{itemize}
with the number representing the camera configuration.
Detailed mounting and FoV specifications are available in \cref{subsub:component_placement}.
We evaluated the performance in the \emph{Mountain Ascent} dataset (see \cref{fig:combined}). 
Both the H-1 and Z-1 configurations diverged, highlighting challenges with rolling shutter cameras. 
The CoreResearch unit, utilizing global shutter cameras, consistently delivers the best accuracy across all environments.
For the HDR image, we provide OKVIS2 with the timestamp of the center line in the middle of the exposure period; however, the first line is captured approximately \SI{15}{\milli\second} earlier, while the last line is captured \SI{15}{\milli\second} later. Our results indicated that not accounting for the rolling shutter effect significantly degrades performance. 
The ZED2i rolling shutter camera is connected via USB (unlike the HDR camera, which uses GMSL2). Therefore, it does not allow for accurate timestamping and is limited to the accuracy of the USB serial buffer readout. 
The three HDR cameras have limited overlapping FoV compared to the ZED2i stereo-pair. This explains the worse performance of the three HDR cameras in comparison to the two ZED2i cameras.
Our results highlight the need to account for the rolling shutter effect, and our paired multi-camera dataset enables rigorous future studies and algorithmic development.

 \begin{figure}
    \centering
    \includegraphics[width=0.95\linewidth]{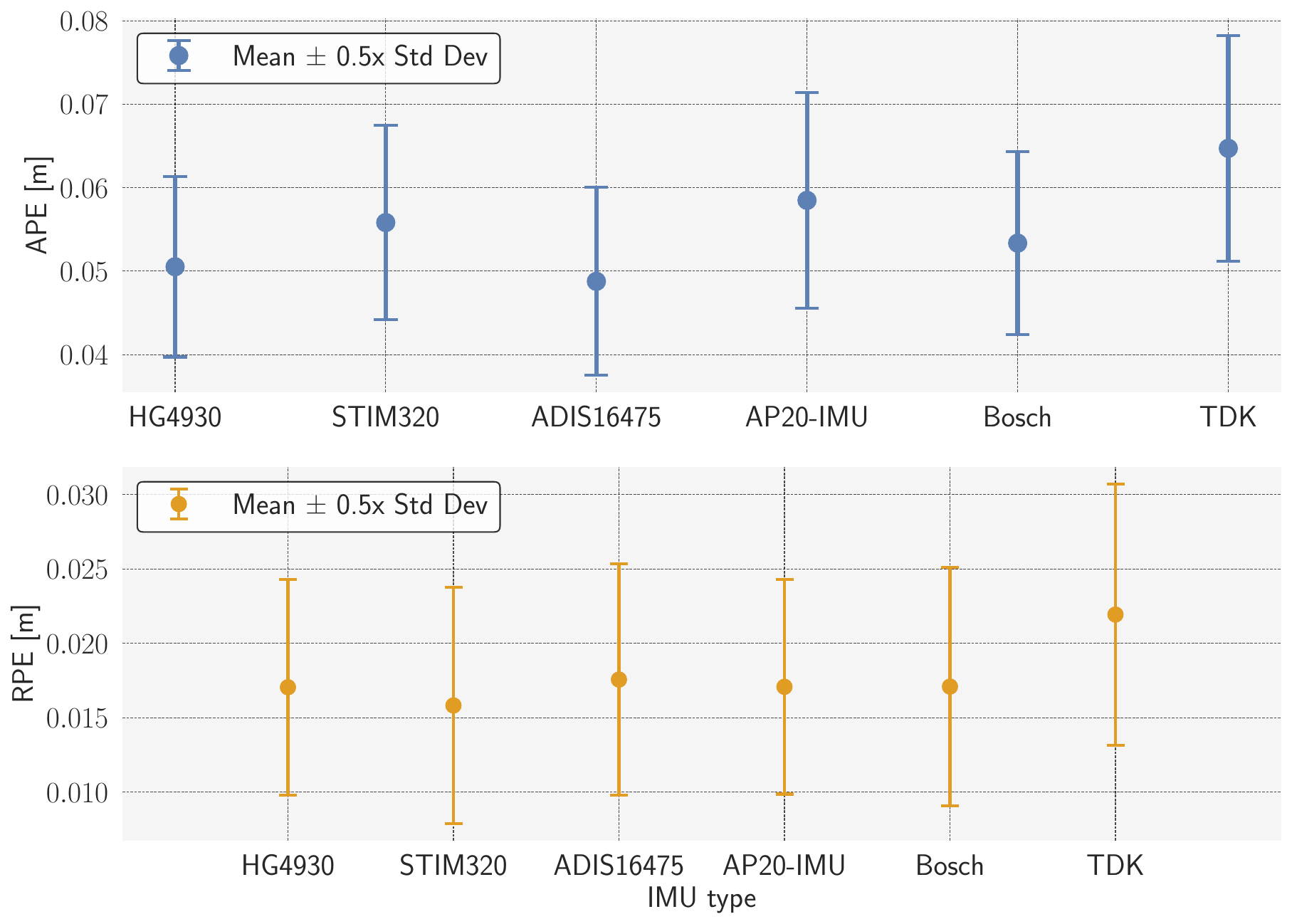}
    \caption{Effect of different IMUs on the performance of DLIO module on the \emph{Mountain Ascend} dataset. The generated \gls{ape} and \gls{rpe} errors are visualized.}
    \label{fig:imu_comp}
\end{figure}

\subsection{State Estimation - Effect of IMUs}
\label{subsec:effect_of_imus}
IMUs are fundamental to state estimation due to their high-rate measurements and their ability to perceive angular velocity alongside linear acceleration. 
Numerous IMUs with different noise characteristics and costs are available to researchers. While the general relation between noise characteristics and algorithmic performance is well-understood, how this relation can manifest itself in real-world scenarios has received limited attention~\cite{de2010performances, nobili2017heterogeneous}. 
To bridge this gap, we tested six different IMUs using DLIO~\cite{dlio} with Hesai LiDAR again on the \emph{Mountain Ascend} dataset. 
\Cref{fig:imu_comp} shows only a marginal difference in \gls{ape} for different IMUs. 
We hypothesize that this is the case due to the tightly coupled LiDAR point cloud registration. 
The point cloud registration corrects the inaccuracies in the IMU state propagation and ensures the correct bias estimation.

\begin{figure}
    \centering
    \includegraphics[width=1.0\linewidth]{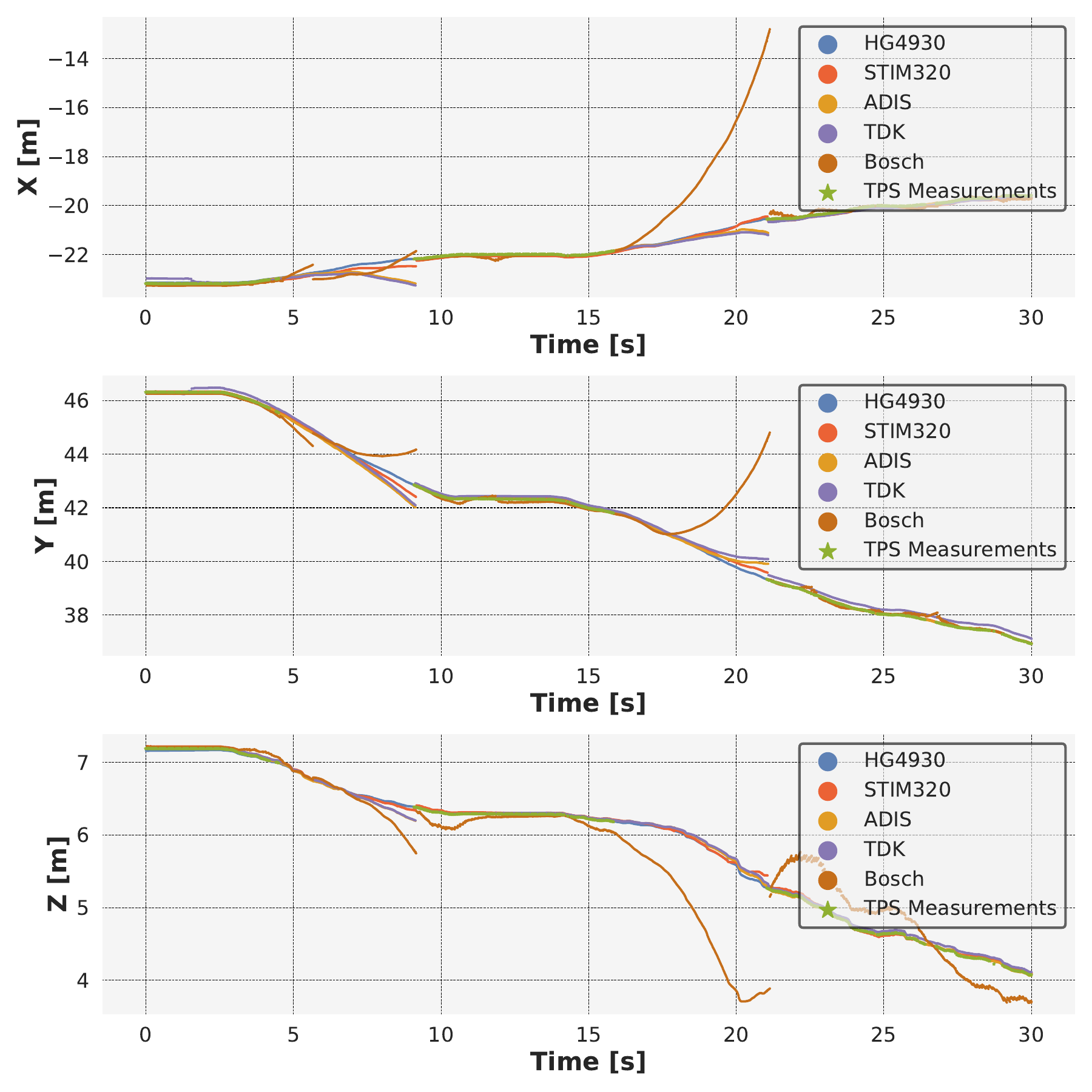}
    \caption{IMU Dead-Reckoning during \gls{tps} measurement dropout (\SI{4.5}{\second}-\SI{9}{\second} and \SI{17}{\second}-\SI{22}{\second}).}
    \label{fig:imu_dead_reckon}
    \vspace{-0.5cm}
\end{figure}

To analyze the impact of IMU noise characteristics independent of the tightly-coupled LiDAR processing, we conducted a dead-reckoning experiment.
The \gls{tps} measurements are fused with IMU measurements using Holistic Fusion (HF) \cite{nubert2025holistic}, similar to our ground truth pose estimate explained in \cref{app:ground_truth}. During a \gls{tps} measurement dropout, IMU dead-reckoning provides insight into the performance difference based on the noise characteristics. 
The results are shown in \Cref{fig:imu_dead_reckon}. TPS measurements are unavailable between \SIrange{4.5}{9}{\second} as well as from \SIrange{17}{22}{\second}. 
As seen in \Cref{fig:imu_dead_reckon}, the tactile-grade HG4930 performs the best, resulting in a smooth alignment with the TPS measurement after the dropout period. On the other hand, consumer-grade IMUs diverge after a short period (around \SI{5}{\second}) of dead-reckoning time. 
For a fair comparison, Holistic Fusion was provided with the noise characteristics specified in the IMU datasheet.

\subsection{State Estimation - Ablations}

The importance of accurate time synchronization and extrinsic sensor calibration has been highlighted by the community as discussed in \Cref{sec:introduction} and \Cref{sec:related}. Despite this common knowledge, the required time synchronization precision and calibration accuracy are often vaguely stated. Therefore, we provide an ablation study on how different time and extrinsic calibration offsets affect performance.

\subsubsection{Perturbation on Time Synchronization}
\label{sec:ablations:time}
\begin{figure}[t] 
    \centering 
    \includegraphics[width=1\linewidth]{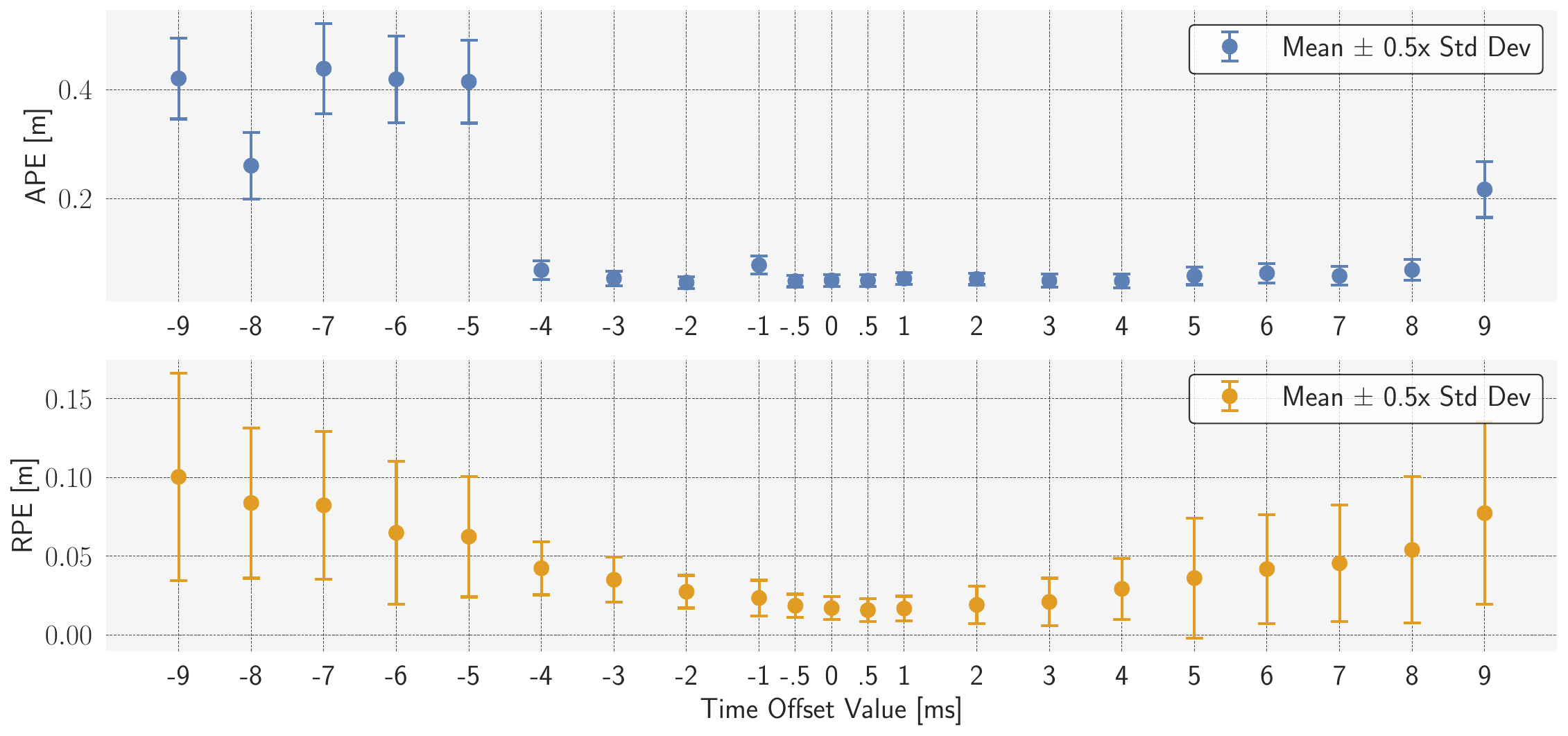} 
    \caption{Time Offset - Performance DLIO (Hesai + HG4930) in \gls{ape} and \gls{rpe} with offset applied to the IMU measurements.}
    \label{fig:dlio_time_offset} 
\end{figure}

Similar to \Cref{subsec:effect_of_imus}, we deployed DLIO fusing Hesai and HG4930 measurements. We added different constant time delays to the IMU measurements and provided \gls{ape} and \gls{rpe} metrics in \Cref{fig:dlio_time_offset}.
A time offset less than \SI{5}{\milli\second} negligibly impacts performance. We attribute this to the fact that the HG4930 IMU operates at \SI{100}{\hertz}, with a respective measurement period of \SI{10}{\milli\second}, which is larger than the time offset. We validated the accuracy of the IMU time synchronization in \cref{subsec:communication} and further found that different IMUs have varying offsets, which can be potentially attributed to internal filtering.

\subsubsection{Perturbation on Extrinsic Calibration}
\begin{figure}[t] 
    \centering 
    \includegraphics[width=1\linewidth]{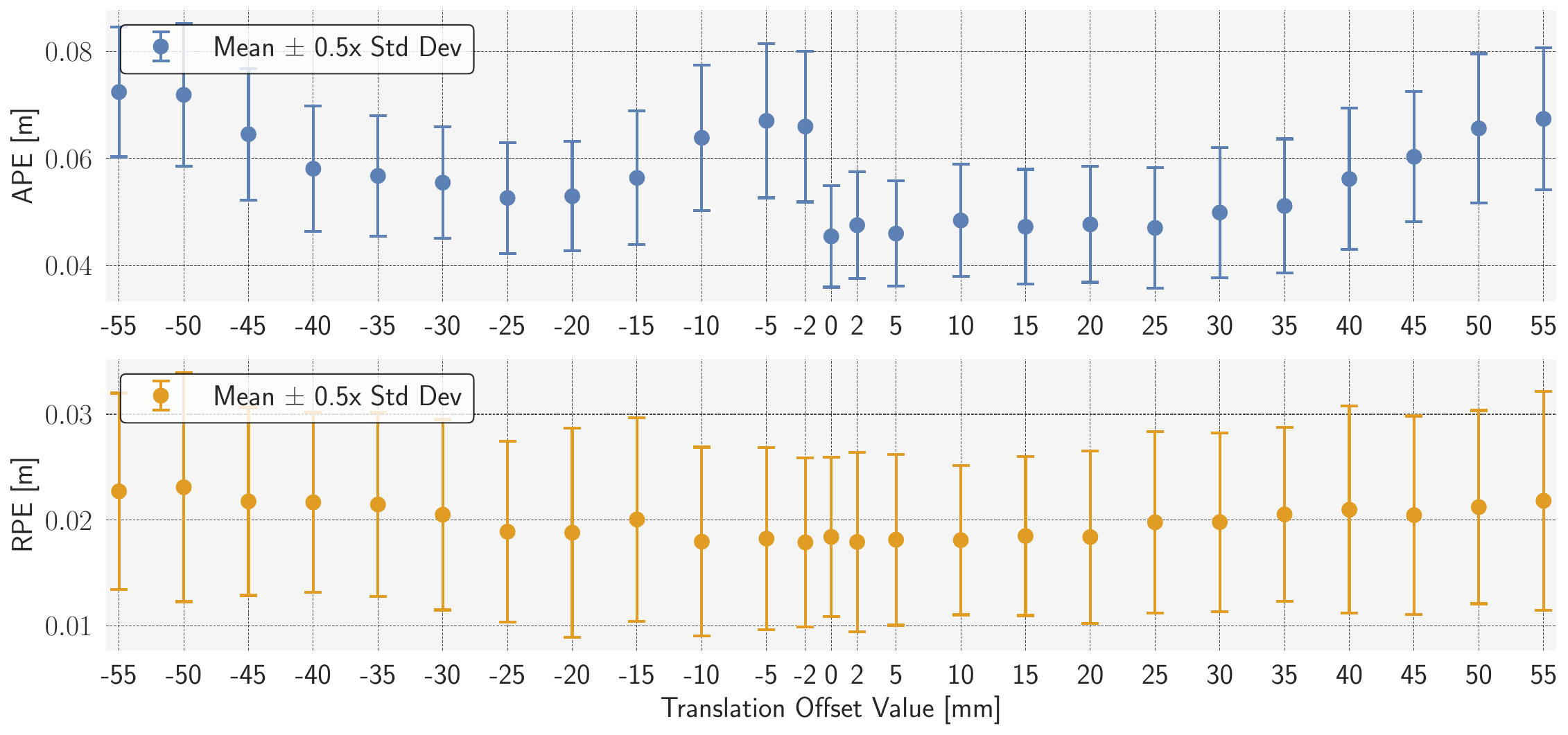} 
    \caption{
    Extrinsic Translation Offset - Performance DLIO (Hesai + HG4930) in \gls{ape} and \gls{rpe} with an offset applied to the positive X-axis of the extrinsic calibration.}
    \label{fig:dlio_translation_offset} 
\end{figure}

\begin{figure}[t] 
    \centering 
    \includegraphics[width=1\linewidth]{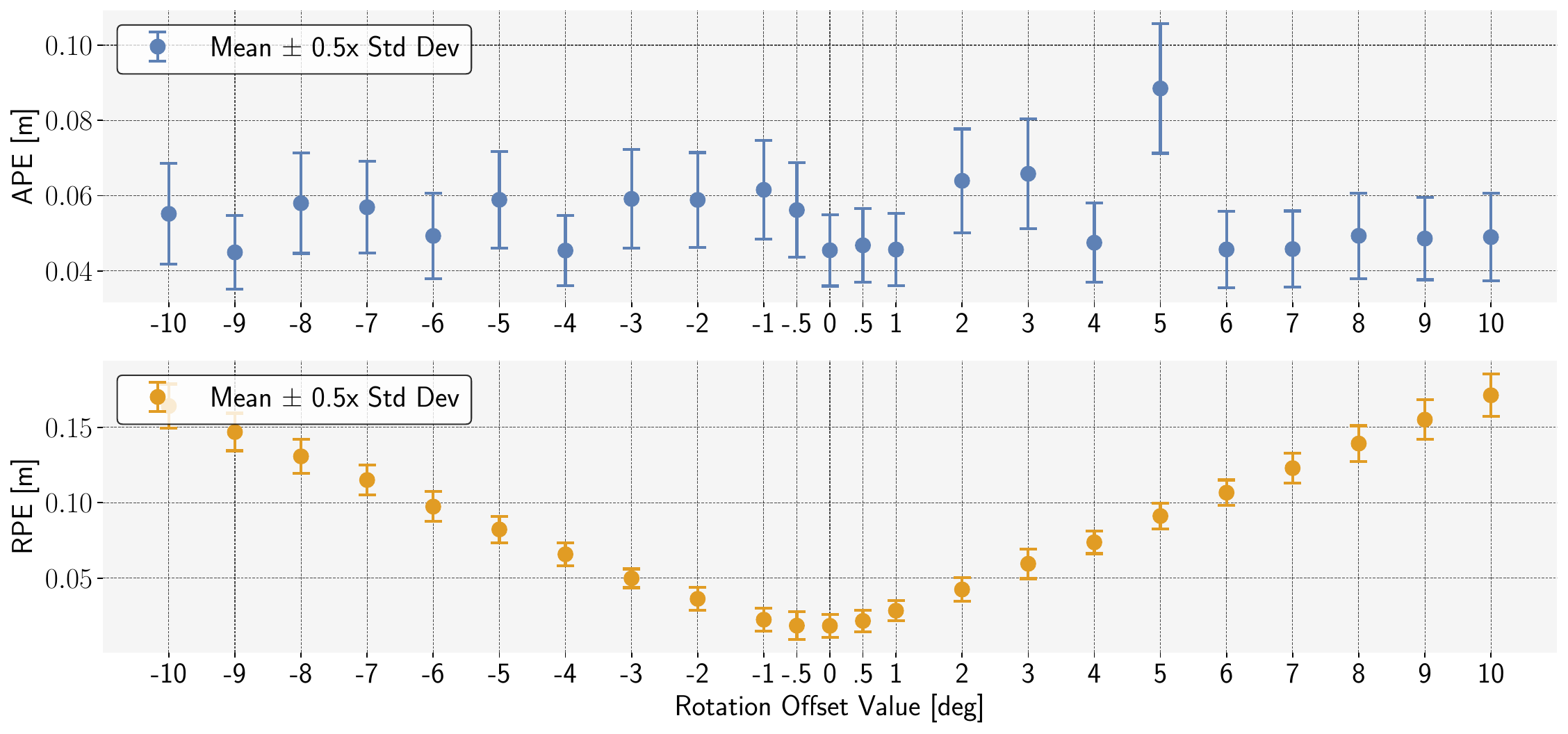} 
    \caption{Extrinsic Rotational Offset - Performance DLIO (Hesai + HG4930) in \gls{ape} and \gls{rpe} with an offset applied to the positive pitch angle of the extrinsic calibration.}
    \label{fig:dlio_rotation_offset} 
\end{figure}

Another fundamental property of a well-designed sensor payload is the availability of reliable and accurate extrinsic calibration between sensors. Similar to \Cref{sec:ablations:time}, we use DLIO and perturb the extrinsic calibration in translation and rotation and provide the results in \cref{fig:dlio_translation_offset} and \cref{fig:dlio_rotation_offset}, respectively.

Below \SI{5}{\milli\meter} translation offset, DLIO performance is minimally impacted (\Cref{fig:dlio_translation_offset}). Moreover, as the offset increases, the tightly coupled LiDAR-Inertial method suffers from this systematic error, and the LiDAR registration cannot converge. DLIO does not perform online extrinsic calibration. The translation offset might be negated by the point cloud registration process employed after the IMU state propagation.\looseness-1%

The rotational component of the extrinsic calibration is more sensitive to perturbations, and errors as small as \SI{2}{\degree} cause the estimates to diverge (\Cref{fig:dlio_rotation_offset}). This observation aligns with the fact that the Hesai is a long-range LiDAR, and the far-away measurements are affected more by the rotational offset. Importantly, DLIO runs a geometric observer to estimate and propagate the state; this geometric observer has convergence characteristics in rotation and subsequently in translation, which explains the robustness against the perturbations.\looseness-1

\subsubsection{Summary}
Using \textit{Boxi}, we demonstrated that performance across modalities varies among the tested environments. 
Naively replacing images recorded from a global shutter camera with rolling shutter images results in a significant performance drop. 
Similarly, switching from a long-range high-accuracy LiDAR to a lower-accuracy short-range LiDAR leads to a significant decrease in performance. 
While using a tactile-grade IMU does not directly lead to superior state estimation for DLIO, in the dead reckoning scenarios, the lower noise tactile-grade IMU can temporarily mitigate drift effectively. 
Our findings show that DLIO is robust to small time-delay errors (up to \SI{2}{\milli\second}) and large translation offsets. However, accurate rotational extrinsic calibration remains critical to ensure precise state estimation.
\subsection{Perception - Mapping}
\label{subsec:app_perception}
\begin{figure}[t] 
    \centering
    \includegraphics[width=1.0\linewidth]{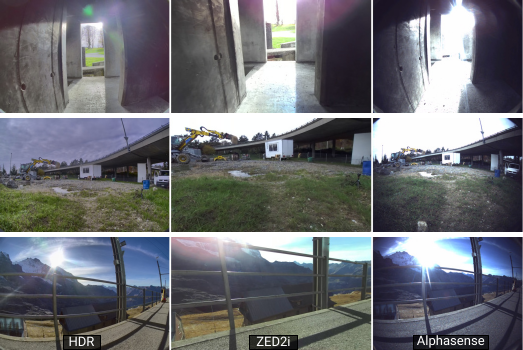}    
    \caption{Image Quality Comparison. The HDR camera excels in dark and low-light conditions. The ZED2i misses details in challenging scenarios (Row 1: Tree, Row 2: Sky). The CoreResearch struggles with overexposure and misses crucial details that are essential for localization.\looseness-1}
    \label{fig:image_quality}
    \vspace{-0.5cm}
\end{figure}
\begin{figure*}[!ht] 
    \centering 
    \includegraphics[width=1.0\linewidth]{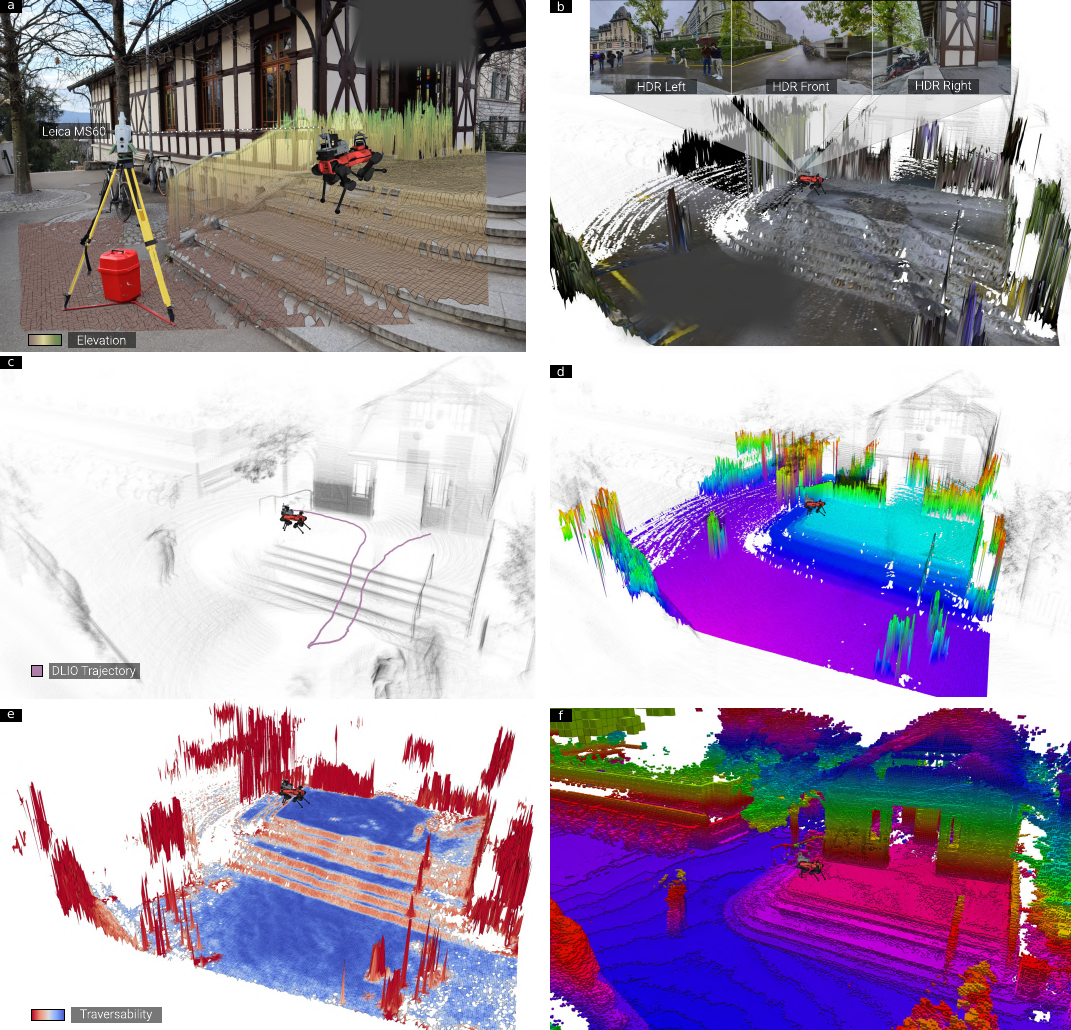} 
    \caption{Example Deployment of Boxi:
(a) \gls{tps} setup used for the ground truth pose generation.
(b) Accurate intrinsic and extrinsic camera calibration enables precise projection of RGB data onto the colorized elevation map.
(c) Accumulated point cloud and trajectory visualization based on the registration by the DLIO module.
(d) The 2.5D elevation mapping shows the smoothness of the local geometry perception.
(e) Traversability analysis derived from the elevation map.
(f) 3D Volumetric map generated using the Wavemap~\cite{reijgwart2023wavemap} module.}
    \label{fig:autonomy}
    \vspace{-0.5cm}
\end{figure*}
To showcase that \textit{Boxi} is not only a SLAM-ready payload but also an autonomy-ready payload, we integrated on-board mapping and traversability analysis. 
Namely, we integrated a previously established elevation mapping~\cite{erni2023mem} framework that combines visual and geometric information (see~\cref{fig:autonomy}-d). 
This framework relies on accurate timestamps and precise extrinsic calibration to associate camera images with the map. 
We choose HDR camera images to be projected onto the map for their superior image quality. While various works have previously shown that different cameras yield varying performance for state estimation, in \cref{fig:image_quality}, we show the influence of different lighting conditions on image quality. It is clear that the HDR camera consistently provides the best visual image quality across all environments, which is crucial for effective mapping and scene understanding. Additionally, we demonstrate how the elevation map can be utilized to assess traversability risks for a legged robot (\cref{fig:autonomy}-e).

We also integrated the 3D volumetric mapping framework, Wavemap~\cite{reijgwart2023wavemap}, which generates a detailed volumetric representation of the scene and is capable of handling overhanging obstacles (\cref{fig:autonomy}-f). The 3D representations and elevation map can be used by the robot for downstream path planning.

\subsection{Limitations - Algorithmic Performance}
\label{sec:limitations_algo}
Even when evaluating methods over a large number of datasets, it is challenging to make generalizable claims that hold across environments, motion characteristics, and different systems. However, the above represents our best attempt to quantitatively assess the sensitivity of different hardware design decisions on state estimation performance.

We have tuned the parameters of DLIO to the best of our knowledge to achieve competitive performance across all datasets. However, for OKVIS2, we did not fine-tune the parameters for each camera setting. In addition, using a framework that directly incorporates rolling shutter compensation could offer deeper insights, as performance is significantly affected when rolling shutter effects are not accounted for.
Additionally, for many experiments, we do not measure effects in isolation — \eg, changing the shutter type alone is not feasible, as the underlying imaging sensor differs between cameras. While we ensured that resolution and \gls{fov} were comparable (\eg, between HDR and CoreResearch), other factors, such as dynamic range and sensor characteristics, still vary.
While we can provide accurate extrinsic and intrinsic calibration for the camera and LiDARs, we found preliminary evidence for a time synchronization offset between IMUs and other sensors within the IMU periods. However, since our evaluation across multiple datasets did not reveal a strong dependence of DLIO on time synchronization, we remain confident that the presented results are valid. Further time synchronization validations are available in \cref{subsec:communication}.
\section{Cookbook to a Sensor Payload}
\label{sec:cookbook}
In this section, we provide a generalizable cookbook to design a sensor payload, covering the most important requirements. 
We explain at first general requirements (\cref{subsec:requ}), followed by the component selection (\cref{subsec:component}), mechanical design (\cref{subsec:mechanical}), electronics and communication (\cref{subsec:electronics}), calibration (\cref{subsec:calibration}), software (\cref{subsec:software}), and our lessons learned (\cref{sec:lessons_learned}).

\subsection{Boxi Payload Requirements}
\label{subsec:requ}
We required \textit{Boxi} to rely only on an external power source and provide a communication line for simple and seamless integration into different platforms without requiring substantial adaptations or modifications. In addition to being compact, the system should be flexible and expandable, allowing for the retrofitting of additional sensing modalities.
A rigid and stiff chassis is essential for protecting all components as well as reliable extrinsic calibration, which must remain consistent over time.
For \textit{Boxi}, the observability of the kinematic chain between the robot and the sensor payload is required, therefore rendering hardware low-pass filtering, such as vibration dampeners, not suitable.

The payload should incorporate mature, proven, and production-ready sensors, which have already been tested in real-world scenarios. To enable on-board autonomy, the \textit{Boxi} has to offer sufficient computational power. This includes adequate CPU and GPU capabilities as well as enough bandwidth, RAM, and storage to handle the collection of multi-hour datasets.
In accordance with the sensor and compute requirements, the total power consumption of the payload should not exceed \textbf{\SI{300}{\watt}} to guarantee long-term deployment with limited battery capacity. Similarly, the weight of the payload is not allowed to exceed \textbf{\SI{10}{\kilo\gram}} to allow for hand-held and general-purpose legged robot operation. 

All components must be synchronized and intrinsically calibrated up to industry standard accuracy. Each sensor must have an appropriate \gls{fov} to enable autonomous navigation, as this is the primary task of the payload, and the \gls{fov} between the sensors should overlap adequately. Finally, the system should be designed to be rugged. It must handle occasional falls and impacts while operating in all-weather conditions, and must be dust-proof and waterproof to the IP65 standard.

\subsection{Component Selection}
\label{subsec:component}

\begin{figure*}
    \centering 
    \includegraphics[width=1.0\linewidth]{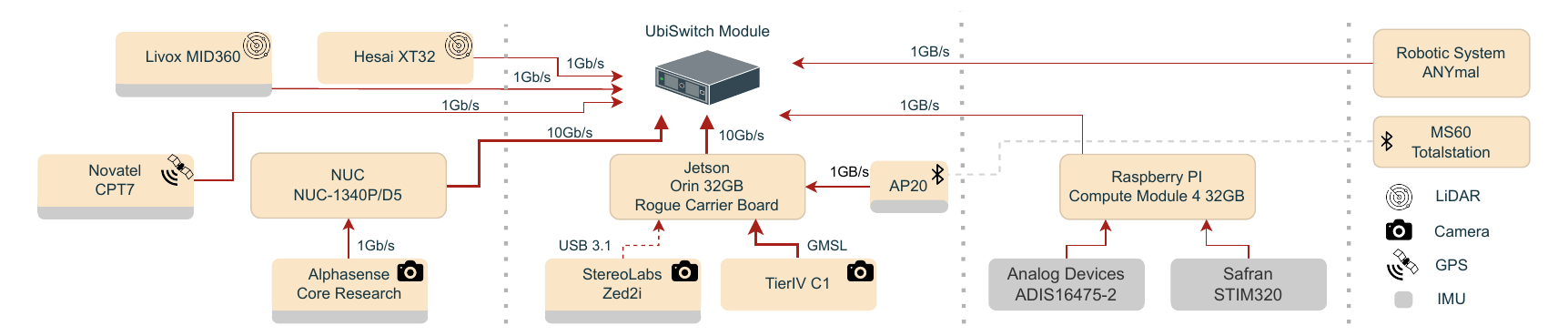} 
    \caption{Communication Overview: The camera sensors are directly connected to the processing compute units to reduce network load.}%
    \label{fig:schematic}
\end{figure*}

\subsubsection{Cameras}
To select the best vision suite, we considered \textit{Sensor \& Lens}, \textit{Interfacing \& Control}, and \textit{Software}. 
Despite many important attributes such as resolution, frame rate, and \gls{fov} being typically taken into consideration, two equally important aspects are often neglected: lens distortion characteristics and depth of field. The former leads to a distorted version of an image compared to a pin-hole camera model, and the latter determines the range at which an object is in focus. The desired focus range depends on the application and can range from centimeter-scale for inspection tasks to \SI{10}{\meter}-\SI{100}{\meter} for autonomous driving.\looseness-1

The camera sensor and shutter mechanism (global or rolling) determine the image quality, as well as the sensor's sensitivity, sensed spectrum, and dynamic range. Although a rolling shutter is more cost-effective, it can introduce additional complexity on the software processing side for certain types of environments and motion characteristics.

The camera interface determines how the captured images are transmitted to the host PC, with common options being USB, Gigabit Multimedia Serial Link version 2 (GMSL2), and Ethernet.
Each connection type has unique benefits or downsides, \eg, additional latency in image transmission time or increased compute requirements on the host PC. 
Among the available interfaces, GMSL2 is currently the emerging industry standard for automotive driving cameras. However, integrating GMSL2 may require additional effort compared to simpler options such as Ethernet and USB, as GMSL2 requires additional deserialization hardware.

Additionally, it is crucial to understand the camera's time synchronization capabilities and whether it supports hardware triggering. 
For advanced applications, synchronizing the camera exposure with the timestamp of other modalities is desirable. In multi-camera setups, particularly for stereo depth estimation, synchronized exposure and triggering may be required.
Lastly, the availability of software drivers (\eg, in ROS1, ROS2, Python, or C++) influences the ease of integration, and the availability of in-built white balancing and color correction features can save significant development effort.

We selected industry-standard cameras across three types: global shutter, off-the-shelf stereo, and high-dynamic range. %
The three selected camera systems are detailed in \cref{app:camera1,app:camera2,app:camera3}.\looseness-1

\begin{magic}[Camera - Recipe]
    \item Interface (\textit{GMSL, Ethernet, USB, ...})
    \item Camera Sensor (\textit{Size, FPS, Shutter, Dynamic Range, ...}) 
    \item Lens (\textit{Distortion, FoV, ...})
    \item Timing (\textit{Synchronization, Triggering, Latency})
    \item Driver Support \& Host Device Requirements
\end{magic}

\subsubsection{LiDARs} 
There exists a variety of LiDAR types, each offering trade-offs in range, accuracy, intensity sensing, multi-return features, wavelength, scan pattern, \gls{fov}, point density, laser disparity, beam divergence, noise characteristics, and cost. 
Most of these trade-offs are straightforward; however, we would like to highlight laser disparity as a key factor, where higher disparity may be desirable for navigation applications, \eg, to detect twigs and branches within a forest, while low disparity is favorable for localization. 
Further, time synchronization is especially crucial for LiDAR. 
Unlike cameras, which have exposure times of multiple milliseconds, LiDARs effectively produce point estimates in time.\looseness-1

For \textit{Boxi}, we chose spinning LiDARs over solid-state LiDARs as the latter typically have a smaller \gls{fov}, are less mature, and are designed mainly for autonomous driving. \textit{Boxi} primarily targets legged robots operating at lower velocities but higher accelerations and with omnidirectional movement, therefore benefiting from the extended \gls{fov} of spinning LiDARs.
More details about the two selected LiDARs are available in~\cref{app:lidar}.\looseness-1

\begin{magic}[LiDAR - Recipe]
    \item Type (\textit{Spinning, Solid State...})
    \item Characteristics (FoV, Accuracy, Rate, Distance, Points/s )
    \item Timing (\textit{Synchronization, Triggering, Latency})
    \item Driver Support \& Host Device Requirements
\end{magic}

\subsubsection{IMUs}
IMUs can be classified into different categories: consumer/industrial, tactical, and navigation grade based on their pricing and stability. Most IMUs used in mobile robots are consumer-grade IMUs and are based on MEMS technology.

\begin{figure*}[t] 
    \centering 
    \includegraphics[width=1.0\linewidth]{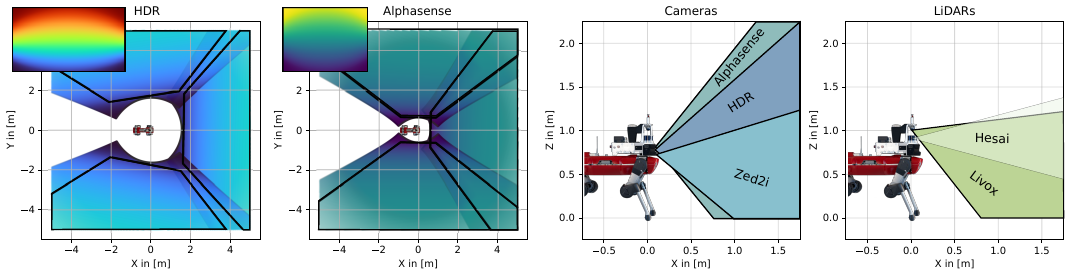} 
    \caption{Visualization of LiDAR and Camera \gls{fov}:
    \textit{(Far Left)} Top-down view of HDR \gls{fov} projected on flat ground plane. The black outline represents the rectified \gls{fov}; The distorted \gls{fov} overlaps while the undistorted one does not overlap. 
    \textit{(Left Center)} Top-down view of the CoreResearch camera \gls{fov} projected on the ground plane. Overlap in both distorted and undistorted \gls{fov}. 
    \textit{(Right Center)} Side view of \gls{fov} for CoreResearch, HDR, and ZED2i cameras, with ZED2i tilted downward at a \SI{15}{\degree} angle and rotated \SI{180}{\degree} in its image axis for better space management within \textit{Boxi}. 
    \textit{(Far Right)} \gls{fov} of Hesai and Livox LiDARs. }
    \label{fig:fov} 
\end{figure*}

Key performance metrics include gyroscope and accelerometer bias instability, hysteresis, temperature drift, aging effects, as well as the random walk characteristics of angular and linear velocity noise. However, datasheet specifications often do not reflect the actual conditions of the target use case. As a result, other error sources, such as temperature dependence or specific motions, can become the dominant contributors to overall measurement errors in real-world applications. 
When choosing an IMU, it is important to match the robot motion profile with the input range, specifically the maximum acceleration and angular rate, to avoid saturation.
In addition, the sampling rate provided for IMU is typically crucial for the motion characteristics of the target application. 
Generally, for all IMUs, it is crucial to ensure that the measurement aggregation configuration and point-of-percussion~\cite{rehder2016extending} correction is available to the user.
Different IMUs require different communication protocols (Serial, SPI, I2C, CAN, Ethernet), which, in combination with time synchronization and setting up on-device filtering chains, can become a multi-week engineering effort. 
When selecting an IMU, we recommend considering software support, bias instability, aging effects, temperature sensitivity, and the availability of low-level drivers provided by the manufacturers, which can significantly reduce integration efforts.
More details of all selected IMUs are provided in \cref{app:imu}.

\looseness-1
\begin{magic}[IMU - Recipe]
    \item Time synchronization requirements
    \item Gyroscope and Accelerometer instability
    \item Maximum Acceleration and Angular Velocity Rating
    \item Software Support - Reducing Integration Effort
\end{magic}

\subsubsection{GNSS}
\gls{gnss} provides drift-free position estimation, making it indispensable for global navigation. Typical position accuracy ranging from \SI{1}{\meter} to \SI{10}{\meter}. Many receivers support external atmospheric correction data to compute a Real-Time Kinematic (RTK) solution, which can significantly improve the overall accuracy to approximately \SI{1}{\centi\meter} in the best case. \gls{gnss} receivers typically support single-antenna or dual-antenna setup; the latter not only yields improved positional accuracy, but also provides heading estimation. The selection of antennas and receiver capabilities ultimately dictates which satellite constellations can be used. Moreover, the considerations of electromagnetic interference (EMI) turned out to be critical when selecting an antenna placement, as noted in \cref{subsubsec:power}.
The atmospheric correction data can be received from one or more nearby base stations via the online GNSS correction providers or a radio link from a local base station.
In environments where real-time communication to the base station is not possible, \gls{ppp} solutions can be employed. With \gls{ppp}, correction data is received through single-path communication via satellite over a large geographic area, resulting in an accuracy up to \SI{2.5}{\centi\meter}. 
More details about our GNSS receiver and postprocessing software are available in \cref{app:gnss}.
\begin{magic}[GNSS - Recipe]
    \item Dual or Single Receiver (Baseline)
    \item Single or multi-band support
    \item Online Corrections (None, RTK, PPP)
    \item EMI (Antenna Placement, Receiver Placement)
    \item Postprocessing Support (Correction data) 
    \item Synchronization against GNSS Time
    \item Connectivity \& Driver Support \& Host Device Requirements
\end{magic}

\subsubsection{Network Switch}
\label{subsub:switch}
Reliable communication between Ethernet-enabled sensors and computers requires a switch or a managed/unmanaged router. 
Key considerations include the number of ports, port speeds, form factor, cost, and switching capacity. One desirable feature is the hardware support for \gls{ptp} time synchronization (IEEE 1588v2), as using a non-compliant switch can increase time synchronization errors. 
More details about the selected switch are provided in \cref{app:switch}.

\begin{magic}[Switch - Recipe]
    \item Management (active or passive)
    \item Switching Capacity, Speed, Size
    \item PTP-enabled
\end{magic}

\subsubsection{Compute}
To establish a versatile research platform capable of deploying various algorithms, we require both a CUDA-accelerated NVIDIA GPU and an x86 CPU architecture. 
Due to the large amount of raw sensory data, two computers are required to perform the required raw data processing and allow for online image compression (see \cref{subsec:software}). This, in addition to the demand for accurate time-stamping and other constraints, led us to integrate three compute modules, namely a NUC, Jetson, and Raspberry Pi (see \cref{fig:schematic}). More details for each selected module are provided in \cref{app:compute}. 
\begin{magic}[Compute - Recipe]
    \item Workload Analysis (CPU, GPU)
    \item Reduce Compute Units to Minimum
    \item Hardware Acceleration (\textit{NVEC H.265, GPU}) 
    \item Connectivity (\textit{Ethernet, USB-X, SPI, GMSL, PCI-E ..})
    \item Time synchronization capabilities (PPS, PTP)
\end{magic}
\subsection{Mechanical}
\label{subsec:mechanical}
The entire payload has been designed with a strong focus on positioning accuracy and stiffness while minimizing weight.

\subsubsection{Component Placement}
\label{subsub:component_placement}

In general, the component placement requires balancing multiple, often conflicting, objectives.
The most important criterion is the safety of each component, specifically the LiDAR sensors and the GNSS antennas, which require a rollover cage. 
Apart from thermal, dust-proofing, and water-proofing considerations, the \gls{fov} for each sensor is essential for the placement of the exteroceptive components. During autonomous operation, each sensor should capture a comprehensive but overlapping view of the environment, which is also key for calibration and multi-modal algorithms. For example, sufficient overlap between LiDARs and cameras allows calibration of the extrinsic parameters using inexpensive targets rather than requiring a dedicated calibration room. 
\Cref{fig:fov} illustrates the overlapping \gls{fov} between all cameras and LiDARs. When mounted on an ANYmal robot with a standing height of approximately \SI{0.8}{\meter}, the cameras capture the ground in front of the feet of the robot starting at a distance of \SI{0.75}{\meter}. This setup effectively balances peripheral coverage and ground visibility for local navigation. 
The front-facing CoreResearch stereo pair has a baseline of \SI{11}{\centi\meter}, which results in accurate depth estimates up to a distance of \SI{20}{\meter} and enables a direct comparison to the ZED2i with rolling shutter.

The Livox is mounted upside down, allowing for accurate estimation of the terrain directly around the robot, which is crucial for 3D near-field mapping. The Hesai is mounted horizontally on top of the camera bundle for optimal \gls{fov} for localization. %

In addition, the primary GNSS antenna is rigidly connected to the lid through standoffs and a copper-shielded element. Our empirical findings showed that the EMI resulting from the LiDARs interferes with the GNSS signal (see \Cref{subsubsec:emi}). 

 \begin{magic}[Component Placement - Recipe]
    \item Interference Analysis
    \item Field of View Analysis
    \item Protection \& Ease of Calibration
\end{magic}

\subsubsection{Tolerances and Assembly}

To meet the \SI{10}{\kilo\gram} weight requirement while maintaining structural integrity, \textit{Boxi} (base and lid) is machined from a single AL-6061 T4 aluminum mono-block with a minimum wall thickness of \SI{2}{\milli\meter}. This material offers a favorable strength-to-weight ratio and enables in-house manufacturing.

All parts follow ISO 2768-1 with a fine tolerance class (\SI{0.05}{\milli\meter}). A full 3D CAD model was created to account for connector spacing and cabling constraints. Positioning pins are used at all interfaces, including between the lid and body and for all sensor mounts, to minimize offsets due to screw backlash and machining tolerances.
A protective cage is mounted on the lid to house the Livox unit and prism while shielding the LiDARs from impact.

\begin{magic}[Tolerances and Assembly  - Recipe]
    \item Material (\textit{Weight, Stiffness, Number of Interfaces})
    \item Thermal behavior of the Material
    \item Manufacturing (\textit{Cost, Precision, Quantity})
    \item Positioning Pins, Thread-locking Adhesives, Torque
    \item Consider Assembly (Plugging in Connectors)
\end{magic}

\subsubsection{Thermal Considerations}
The primary thermal load within \textit{Boxi} originates from the two main compute modules and the DC-DC converter.
To balance cooling performance, simplicity, and cost-effectiveness, we selected a convection-based cooling strategy. All heat-generating components were strategically placed at the rear, while cameras and LiDAR sensors were placed at the front of \textit{Boxi} to limit the adverse effects of thermal expansion. To improve thermal management, cooling fins were integrated beneath the two compute modules, complemented by a single radial fan to actively circulate air. 

\begin{figure}[t] 
    \centering 
    \includegraphics[width=1.0\linewidth]{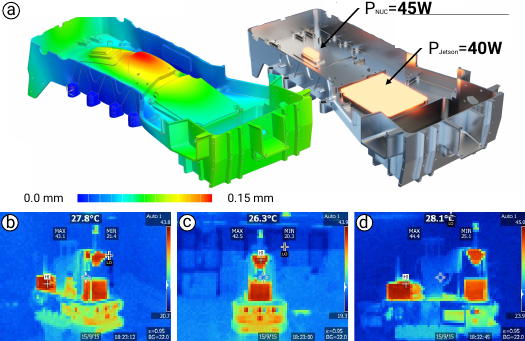} 
    \caption{(a) Thermal deformation from heat generated by the compute modules.
    The deformation is amplified and visualized for clarity, with the mounting points assumed to remain at a constant temperature. (b-d) Thermal imaging of the box during recording.}
    \label{fig:thermal} 
    \vspace{-0.5cm}
\end{figure}

To verify the efficacy of the cooling system and better understand the impact of temperature changes on the deformation of \textit{Boxi} and, in turn, the effect on extrinsic calibration, we conducted a thermal \gls{fem} simulation under maximum thermal load. 
The maximum heat generation from all significant sources adds up to \SI{92}{\watt} dissipation.
The FEM simulation results shown in, \Cref{fig:thermal}, indicate a low overall deformation in the sensor area.
Under maximum load, the relative rotation between the sensors remains within \SI{0.004}{\degree}, which is less than the machine tolerance and the accuracy of the calibration. The simulation validates our thermal design and shows that aluminum AL6061-T4 is sufficient, even though it has double the thermal expansion coefficient of regular stainless steel~\cite{thermal_expansion2005}.\looseness-1

We also verified the thermal design of \textit{Boxi} under normal load (recording, see. \cref{subsec:software}) with an ambient temperature of \SI{19}{\celsius}. We monitored the maximum temperature of the housing using a Fluke Thermal Camera and measured the internal CPU temperature of the NUC at \SI{55}{\celsius} and the Jetson at \SI{51}{\celsius}. 
The entire housing remained well below \SI{30}{\celsius}, while the fan operated at \SI{30}{\%} speed. The CPT7 with \SI{43}{\celsius} and all cameras and LiDARs with a temperature below \SI{40}{\celsius} remained well within the operation specifications.

\begin{magic}[Thermal Considerations - Recipe]
    \item Heat Sources (\textit{Compute Units, Sensors})
    \item Ambient Conditions (\textit{Temp. Range, Humidity Levels})
    \item Thermal FEM Simulation (\textit{Angular and  Linear Deformation})
\end{magic}

\subsubsection{Advanced Topics}
As \textit{Boxi} is intended to be mounted on a legged robotic platform, we also considered topics such as vibrations, weatherproofing, and fall protection. The details of these topics are provided in \cref{app:advanced_hardware}.

\subsubsection{Recommendations and Reflections} 

The dust- and waterproof design proved to be highly effective, particularly with the fins strategically placed on the exterior.
Integrating an off-the-shelf modular cable pass-through was both simple to integrate and offered flexibility. 
When designing rollover cages, it is crucial to always assume the worst-case scenario and identify potential failure points under various conditions.
While the compact form-factor fan was desirable, it generates substantial noise at high RPMs.

We initially used the manufacturer's camera \gls{fov} to design the housing. However, this specification was applied to rectified images, which led to the chassis obstructing a significant portion of the image because of the fisheye effect. 
Manufacturing the base body from a single block of aluminum significantly improved stiffness, precision, and thermal properties while simplifying the design. However, this approach may be impractical and prohibitively expensive for commercial applications.

\subsection{Electronics and Communication}
\label{subsec:electronics}
\subsubsection{Power}
\label{subsubsec:power}
\textit{Boxi} supports a wide input voltage range from \SI{18}{\volt} to \SI{60}{\volt} for compatibility across platforms. Power is distributed via a custom PCB integrating two step-down converters: a \SI{12}{\volt}, \SI{300}{\watt} (Traco TEQ 300-4812WIR) and a \SI{5}{\volt}, \SI{15}{\watt}.
To mitigate EMI and ground loop issues, we used ground-isolated converters, added input filter coils, and implemented a star ground layout. Distinct connector types were chosen for different power levels to prevent misconnection and ensure electrical safety. We recommend analyzing required voltage levels and power consumption, considering the inrush currents of all components, and hot-plug battery exchange if desired.
\begin{magic}[Power - Recipe]
    \item Power Analysis (\textit{Voltage Levels, Power, Inrush Current})
    \item Distribution (\textit{Cabling, Polarized Connectors})
    \item Best practices to prevent EMI
\end{magic}

\subsubsection{Communication and Synchronization}
\label{subsec:communication}

Extending on the basic knowledge on communication and time synchronization provided in~\cref{app:sync_basics}, components that operate on independent clocks require active synchronization to mitigate drift. Typical consumer-grade crystal oscillators drift at a rate of \SI{50}{\ppm} (parts per million), leading to \SI{29.9}{\milli\second} drift per \SI{10}{\min}. 
Two common synchronization protocols are \gls{ptp} (\eg, \texttt{ptp4l}, \texttt{phc2sys}) and NTP (\eg, \texttt{chrony}). \gls{ptp} is generally preferred when reliable network connections are available, as it can achieve nanosecond-level precision.
As mentioned before, the GNSS module CPT7 is used as a clock reference and \gls{ptp} Grandmaster clock.
Since the switch in \textit{Boxi} does not support hardware \gls{ptp} synchronization (IEEE 1588v2; see \cref{subsub:switch}), we conducted a \SI{43}{\hour} continuous, time synchronization drift analysis, showcasing high-quality time-synchronization over the entire period. More details and results on this experiment can be found in \cref{app:ptp}.

\paragraph{Synchronization - Camera and LiDAR}
The time synchronization of both LiDARs has been established with the IEEE 1588v2 (2008) standard \gls{ptp} protocol, which indicates \gls{ptp} clock accuracy $\le \SI{1}{\micro\second}$ and \gls{ptp} clock drift $\le \SI{1}{\micro\second\per\second}$. Both LiDARs provide a per-point timestamp for accurate integration to downstream tasks. Different from the LiDARs, cameras require an exposure time of around \SI{10}{\milli\second} for normal lighting conditions. 
Ideally, the timestamp should correspond to the midpoint of the exposure for each pixel. In the case of our rolling shutter HDR cameras, one needs to calculate the exposure time for every $i$-th horizontal line of the image separately:\looseness-1
\begin{equation}
    i \times \left(\frac{1}{30 \times 1400}\right) = i \times 23.8 \, \mu \text{s}
\end{equation}
Thus, the last line is exposed exactly \SI{30.464}{\milli\second} after the first. 
Similarly to the LiDAR case, compensating for this different trigger and readout time improves state estimation and requires precise synchronization~\cite{sun2020scalability}. Manufacturer data is not available for the ZED2i, regarding exposure time or global shutter compensation.\looseness-1

\paragraph{Synchronization - IMUs}
All 7 IMUs available within \textit{Boxi} are time synchronized to a varying level of precision. The Raspberry Pi Compute Module 4 handles the time synchronization ADIS and STIM320 using hardware timestamping via custom-written kernel modules. 
The other IMUs are synchronized by the manufacturer-provided drivers. 
More information about the IMU time synchronization details can be found in the \cref{app:time_sync_imus}.
\paragraph{Synchronization - Verification}
When it comes to time synchronization, one of the essential factors is the validation of the claimed accuracy. 
While tools such as \texttt{ptp4l} provide the estimated time tracking offset~\cite{watt2015understanding} (see \cref{app:ptp}), this is harder to achieve for sensors synchronized with other means. 
Since we have IMUs connected to each PC, the time synchronization accuracy can be validated by aligning the angular velocities of the IMUs over time, similar to the approaches in~\cite{twistnsync, helmberger2022hilti}.
We have developed a standalone, open-source tool for verifying time synchronization between two IMU time series measurements at arbitrary rates. We ran the verification tool across all recorded missions to obtain a confidence interval per mission using the HG4930 as a reference IMU. We applied the tool to three \SI{30}{\second} snippets per axis for each mission. The results are shown in \cref{tab:alignment} and indicate that the offset between all IMUs can be estimated up to a standard deviation of around \SI{0.5}{ms} and is constantly below \SI{4.5}{ms}, which can be attributed to the exposure time of the IMU and internal time delays. The details of the time synchronization validation tool and procedure are provided in \cref{app:time_sync_validation}.\looseness-1
\begin{table}[h]
\centering
\resizebox{1.0\columnwidth}{!}{
\begin{tabular}{lcccc}
\toprule
\textbf{IMU} & \makecell{ \textbf{x-Axis} \\ $\mu\pm\sigma$  [\SI{}{\milli\second}]} & \makecell{\textbf{y-Axis} \\ $\mu\pm\sigma$  [\SI{}{\milli\second}]} & \makecell{\textbf{z-Axis}\\ $\mu\pm\sigma$  [\SI{}{\milli\second}]} & \makecell{\textbf{AVG} \\ $\mu\pm\sigma$  [\SI{}{\milli\second}]} \\ 
\midrule
ADIS     &  1.41  $\pm$ 0.16 &   1.33  $\pm$   0.09   &    1.35 $\pm$   0.32       &   1.36 $\pm$  0.21  \\
Bosch    &  3.29  $\pm$ 0.14 &   3.20  $\pm$   0.16   &    3.35 $\pm$   0.36       &   3.28 $\pm$  0.25  \\
AP20-IMU &  1.03  $\pm$ 0.13 &   0.85  $\pm$   0.14   &    1.71 $\pm$   0.40       &   1.20 $\pm$  0.45  \\
HG4930   &  0.00  $\pm$ 0.00 &   0.00  $\pm$   0.00   &    0.00 $\pm$   0.00       &   0.00 $\pm$  0.00  \\
TDK      & -4.43  $\pm$ 0.22 &   -4.34  $\pm$  0.18   &   -3.14 $\pm$   1.74     &    -3.97 $\pm$  1.16  \\
\bottomrule
\end{tabular}
}
\caption{Evaluation of the IMU Software Alignment Tool across datasets, with three alignment runs on different sub-trajectories per dataset. We provide alignment results for each axis, along with the mean and standard deviation.}
\label{tab:alignment}
\end{table}
For verification of the camera time synchronization, there is the option to hardware trigger an external LED; however, this procedure is time-intensive and out of the scope of this work~\cite{tschopp2020versavis}. Lastly, all IMUs, and LiDARs operate in free-running mode. It was sufficient for our use case to only trigger cameras of the same type simultaneously. 
\begin{magic}[Synchronization - Recipe]
    \item Bandwidth and Latency 
    \item Precision Time Protocol (PTP) (Preferred)
    \item Network Time Protocol (NTP) (Unreliable Network)
    \item Hardware Timestamping (Linux Kernel Modules)
    \item Verification Tools
\end{magic}

\subsubsection{Electro Magnetic Interface (EMI)}
\label{subsubsec:emi}

EMI mitigation in \textit{Boxi} focused on minimizing both emissions and internal interference. Analog signal integrity was prioritized by placing sensitive components, such as the GNSS receiver, outside the enclosure and using additional EMI absorbers. Testing showed that maximizing the distance between the GNSS antennas and the main housing, avoiding direct contact with the metal structure, and adding a copper ground plane significantly reduced interference.
Inside the enclosure, all signal lines use shielded cables where possible. An exception is the \SI{1}{\Gbps} Ethernet link, which uses unshielded twisted pairs to simplify routing.
Two openings in the aluminum housing on the side panels are fitted with 3D-printed plastic parts, enabling the mounting of Wi-Fi antennas within \textit{Boxi}.
This design represents a trade-off between compact integration and EMI mitigation. As an alternative, the WiFi antennas can be routed outside the box.
\begin{magic}[EMI - Recipe]
    \item EMI Emission Requirements 
    \item Signal Integrity (GNSS Sensitive)
    \item Shielding, Grounding, Minimize Cable Length
\end{magic}

\subsection{Calibration}
\label{subsec:calibration}
Here, we detail the intrinsic and extrinsic calibrations of all sensors, covering the calibration methods, procedures, and verification techniques. These calibrations comprise the intrinsic and extrinsic calibration of all cameras with respect to each other (the bundle of all cameras), the extrinsic calibration of each LiDAR, jointly, to all CoreResearch cameras, the extrinsic calibration of each IMU to the front-facing global shutter cameras, and the prism position in the reference frame of the camera bundle. Additionally, the IMU drift parameters are calibrated.\looseness-1

\begin{figure}[h]
    \centering
    \includegraphics[width=1.0\linewidth]{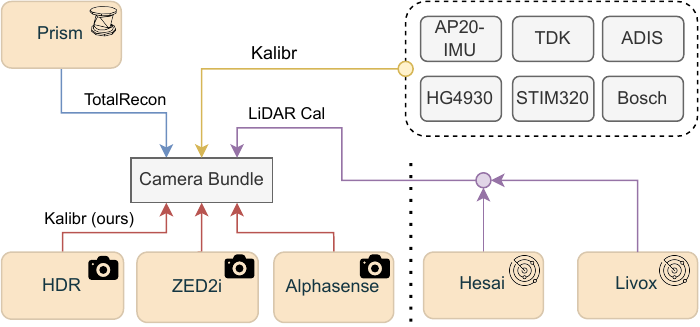}
    \caption{An overview of the calibrations done. The arrows indicate the transforms that are explicitly calibrated.}
    \label{fig:calibration_chains}
\end{figure}
Given the large number of available sensors --- 2 LiDARS, 10 cameras, 7 IMUs, and a reflector prism --- we carefully designed the calibration method, setup, and procedure.\looseness-1
\subsubsection{Calibration Chain}
Central to the calibration of our system are the intrinsic and extrinsic calibrations of the 10 cameras. Since cameras cover a wide \gls{fov}, sufficient overlap is available to robustly calibrate the extrinsic parameters by detecting a calibration target placed in various poses. The camera bundle calibration forms the foundation for calibrating all other sensors in the subsequent steps of the calibration chain~\cref{fig:calibration_chains}.
\subsubsection{Camera Intrinsic and Extrinsic}
For camera intrinsic and extrinsic calibration, we re-implemented Kalibr~\cite{furgale2013kalibr, rehder2016extending} to allow for online calibration and feedback during the recording session. 
The calibration can be decomposed into an intrinsic and extrinsic phase.
During the intrinsic phase, the operator moves an AprilGrid \cite{aprilgrid2011} in front of all cameras until sufficient samples are collected (see \cref{app:calib_int_method}). Then, in the extrinsic phase, it is required to statically place the target at different locations around the \textit{Boxi} or move the payload itself until sufficient static samples are collected to avoid time synchronization errors and motion blur (see \cref{app:calib_int_method}). We periodically estimate the covariance of the intrinsic and extrinsic parameters in real-time to provide feedback on the sufficiency of the data collected.
We used the Kannala-Brandt~\cite{kannalbrandt2006} equidistant distortion model for the wide-angle fisheye CoreResearch~\cite{ramezani2020newer, tao2024oxford} and HDR cameras, and the rad-tan distortion model for the ZED2i. 
More details on why we chose the AprilGrid can be found in \cref{app:target}. Similarly, details on reprojection errors across all cameras are provided in \cref{app:cam_reproj_err}

In \cref{tab:camera_calib_statistics}, we show the numerical evaluation of the estimates of the extrinsic and intrinsic parameters, which are also presented live to the user during the calibration procedure. The standard deviation of the translation and rotation for all cameras is excellent.
The total intrinsic and extrinsic calibration procedure takes around \SI{5}{min} compared to the default Kalibr workflow of multiple hours.\looseness-1

\begin{table*}[ht]
\centering
\ra{1.0}
\begin{tabular}{lcccccccccc}
\toprule
 & \multicolumn{4}{c}{\textbf{Translation} (m)} & \multicolumn{4}{c}{\textbf{Rotation} (\degree)} & \multicolumn{2}{c}{\textbf{Error} (px)} \\ \cmidrule(r){2-5} \cmidrule(r){6-9} \cmidrule(r){10-11} 
Camera & x & y & z & $\sigma_t$ & Roll & Pitch & Yaw & $\sigma_r$ & Mean & Med \\
\midrule
\cam{CoreResearch}{Front-Center} & \_  & \_  & \_ & \_  & \_  & \_  & \_  & \_  & 0.32 & 0.25 \\ 
\cam{CoreResearch}{Front-Right} & 0.0544 & 0.0011 & 0.0002 & 0.0001 & 0.09 & 0.13 & -0.22  & 0.005 & 0.20 & 0.16 \\ 
\cam{CoreResearch}{Front-Left} & -0.0559 & -0.0004 & 0.0009  & 0.0001 &0.04 & -0.01 & 0.96  & 0.005 & 0.28 & 0.20 \\ 
\cam{CoreResearch}{Left} & -0.1019 & 0.0115 & -0.0681 & 0.0002 & 84.10 & -80.23 & -93.59 & 0.008 & 0.24 & 0.20 \\ 
\cam{CoreResearch}{Right} & 0.1041 & 0.0144 & -0.0693 & 0.0002 & 79.62 & 80.68 & 90.57 & 0.010 & 0.39 & 0.31 \\ 
\cam{HDR}{Front} & 0.0002 & 0.0337 & 0.0076  & 0.0001 & 9.89 & 0.23 & 0.29 & 0.006 & 0.52 & 0.44 \\ 
\cam{HDR}{Left} & -0.1028 & 0.0479 & -0.0671 & 0.0002 & 135.55 & -75.47 & -135.43 & 0.010 & 0.39 & 0.34 \\ 
\cam{HDR}{Right} & 0.1021 & 0.0478 & -0.0684 & 0.0003 & 134.93 & 76.10 & 136.21 & 0.012 & 0.39 & 0.34 \\ 
\cam{ZED2i}{Left} & 0.0594 & 0.0759 & 0.0202 & 0.0002 & -0.07 & -0.03 & -179.34 & 0.014 & 0.63 & 0.56 \\ 
\cam{ZED2i}{Right} & -0.0601 & 0.0744 & 0.0203 & 0.0002 & -0.00 & -0.08 & -179.44 & 0.016 & 0.57 & 0.48 \\ 
\bottomrule
\end{tabular}
\caption{Typical results from a full camera-camera intrinsic and extrinsic calibration. The spatial transforms are with respect to CoreResearch-Front-Center, hence is has no entries for these values as it is the origin. Note that we do not impose a stereo constraint on the ZED2i camera during calibration, yet the result is precisely consistent with the stereo baseline of the camera.}
\label{tab:camera_calib_statistics}
\end{table*}

\begin{figure}[t] 
    \centering 
    \includegraphics[width=1.0\linewidth]{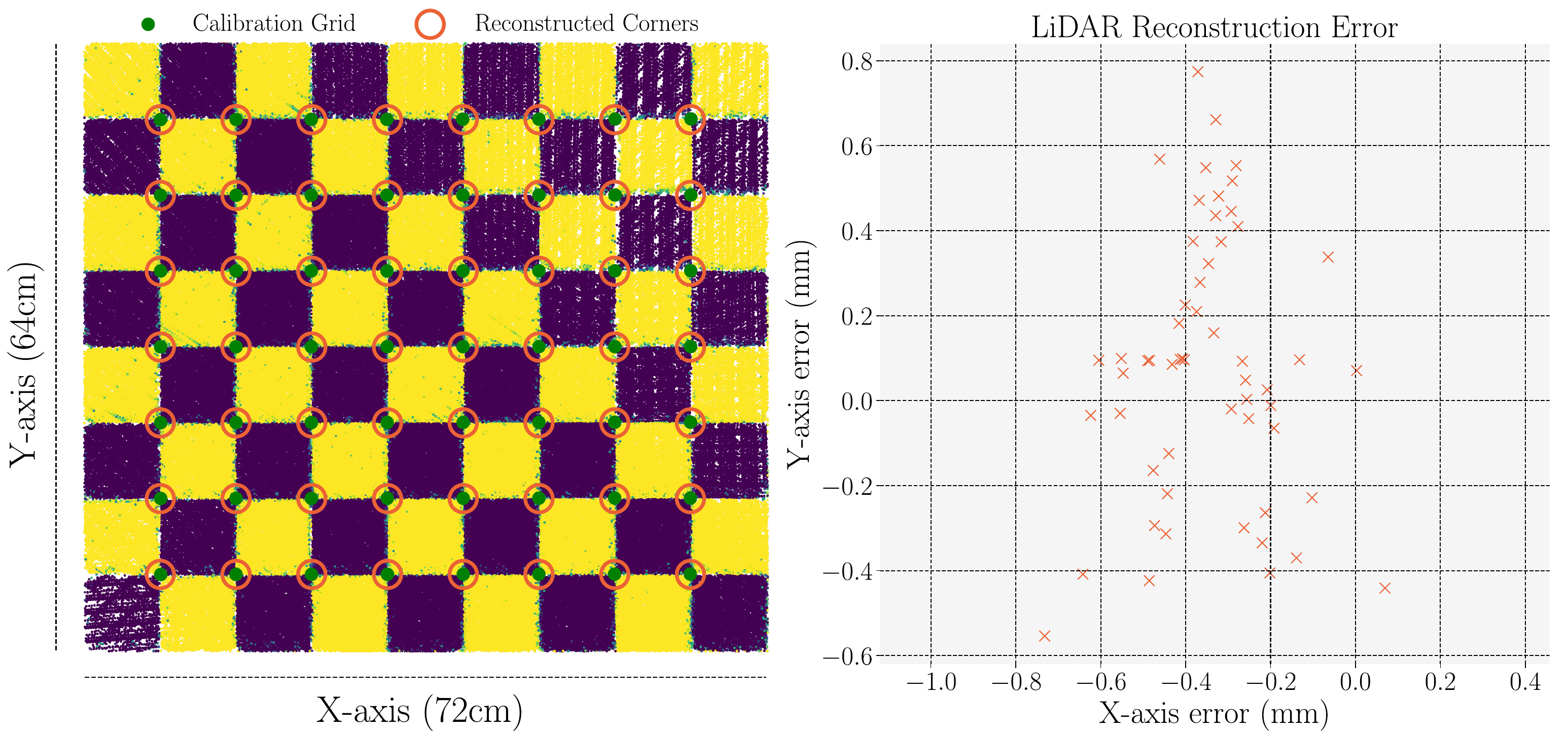} 
    \caption{LiDAR reconstruction of the calibration target using camera-detected poses and the optimized extrinsic calibration. The reconstructed corner locations are sub-millimeter accurate.}
    \label{fig:lidar_reconstruction}
    \vspace{-0.5cm}
\end{figure}

\subsubsection{Camera to IMU}
Before calibrating the camera-IMU extrinsic parameters, we collect static data of the IMUs over \SI{18}{\hour} to model the Allan variances of each of the IMUs using~\cite{AllanVarianceRos}. 
We then perform camera-IMU extrinsic calibration using the front-facing global shutter CoreResearch cameras using Kalibr~\cite{furgale2013kalibr}. To collect camera data capable of capturing the dynamic motion required for the camera-IMU calibration accurately, we set the camera frame rate to \SI{17}{\hertz}, with an exposure time of \SI{1}{\milli\second}. With this exposure time, we also employ uniform artificial lighting of the static calibration target to aid in the detection of the calibration target. More details about the motions and individual IMUs can be found in the \cref{app:camera_to_imu}

\subsubsection{LiDAR to Camera Extrinsic}
To efficiently perform the LiDAR to camera extrinsic calibration, we used a variant of the intensity alignment extrinsic calibration DiffCal~\cite{diffcal}. Unlike plane matching, which requires many samples spanning different orientations, this intensity-matching method can recover the calibration using a single sample. Additionally, intensity matching allows one to incorporate partial observations of the calibration target in the LiDAR point cloud, which is typical for our Hesai LiDAR with a narrow \gls{fov}. Allowing for partial observations enables the collection of calibration data closer to the cameras, where the detected poses are more accurate.

For this calibration, we used a different calibration target formed of a checkerboard pattern. Each LiDAR is individually calibrated against the 5 CoreResearch cameras. Using the calibrated extrinsic parameters between the cameras, the detected calibration target poses from each camera are fused and jointly calibrated with respect to each LiDAR, without refining the intra-camera extrinsic parameters. More details on the calibration target and data collection can be found in \cref{app:lidar_to_camera}. \looseness-1

\begin{figure}[t]
    \centering
    \includegraphics[width=0.45\linewidth]{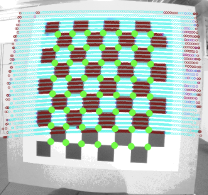}
    \includegraphics[width=0.420\linewidth]{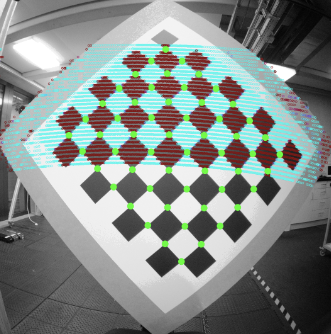}
    \caption{Example calibration data samples used for the LiDAR to camera calibration. Hesai LiDAR points are overlayed on the image, and colored by each point's intensity channel. The calibration target reflects the LiDAR beams with intensity contrast between the black and white squares seen in the camera image.}
    \label{fig:lidar_contrast}
    \vspace{-0.5cm}
\end{figure}

\Cref{fig:lidar_reconstruction} shows a dense, reconstructed point cloud of the calibration target, colored by the intensity channel. This reconstruction has the coordinate system of the calibration target as its origin. From this dense reconstruction, we use a line-based corner detection algorithm (similar to~\cite{cvisic2022}) to detect the locations of the checkerboard corners in the reconstruction. \cref{fig:lidar_contrast} shows the sub-millimeter consistency between the detected corners and their true positions on the canonical calibration target. In~\cref{fig:lidar_contrast}, we show example overlays of the LiDAR points on the calibration target, after optimization. Another example of camera-LiDAR overlay during rapid motion is provided in \cref{app:lidar_to_camera_overlay}.\looseness-1

\subsubsection{Camera to Prism}
Due to the lack of existing methods, we developed a custom calibration procedure, \textit{TotalRecon}, to calibrate the front-facing CoreResearch cameras to the total station prism. Using \gls{tps} measurements and camera detections of a static calibration target, we solve for a 9-\gls{dof} state: the prism position relative to the camera, and the 6-\gls{dof} pose of the target with respect to the total station.
To ensure accurate target detection, the camera is positioned within~\SI{1}{\meter} of the target. The total station tracks the prism while \textit{Boxi} is held static at various poses around the target using a tripod. Sufficient samples are collected to excite all translational and rotational degrees of freedom.
We validate the calibration by checking the consistency of prism estimates across all cameras, transforming them to the front-center camera frame using known extrinsics. The routine yields~\SI{3}{\milli\meter} cross-camera consistency and achieves accuracy below the mechanical mounting tolerance.

\begin{magic}[Calibration - Recipe]
    \item Calibration Requirements and Chains
    \item Available Calibration Methods (CAD, Tools)
    \item Calibration Targets and Procedure Requirements
    \item Calibration Intervals
    \item Change of Calibration (Time, Temperature)
    \item Calibration Verification
\end{magic}

\subsection{Software}
\label{subsec:software}

The hardware and software design of \textit{Boxi} were co-developed. To ensure ease of use for researchers, the software is based on Ubuntu 20.04 LTS with ROS1 Noetic and ROS2 Humble, managed via Docker. A custom kernel was used on the Raspberry Pi Compute Module 4 to enable IIO support and accurate kernel timestamping.
For all ROS1 components, we ran a single \texttt{roscore} instance per PC and enabled inter-PC communication using the fkie multi-master package~\cite{fkie_multimaster}, which utilizes gRPC under the hood. \looseness-1

\subsubsection{Recording} 
In line with best practices established in previous research~\cite{geiger2013vision, ramezani2020newer},  we only run the required sensor drivers and prioritize the recording of raw driver data to minimize artifacts, minimize processing time during recording, and preserve future processing flexibility. As an example, the UDP packets are recorded for Hesai LiDAR instead of point clouds, and the serialized SVO2 files are recorded for the ZED2i RGB-D camera to maintain maximum fidelity.
We implemented a distributed recording system, which stores data on the Jetson, NUC, and CPT7, minimizing network bandwidth and distributing load across machines. 
Specific details about image compression and recording details can be found in \cref{app:image_recording}.
In summary, recording for 10 minutes results in \SI{14.4}{\giga\byte} HDR, \SI{5.77}{\giga\byte} CoreResearch, \SI{1.0}{\giga\byte} ZED2i, \SI{1.14}{\giga\byte} Hesai, \SI{1.76}{\giga\byte} Livox, and >\SI{100}{\mega\byte} for all IMU and other auxiliary status, diagnostic data, totaling at around \SI{25}{\giga\byte} per 10 minutes of deployment.

During this recording, the Jetson exhibited higher CPU usage (72.0\%) compared to the NUC (49.2\%), while only the Jetson showed GPU activity (7.0\%). The NUC utilized the network at \SI{58.2}{\Mbps}, whereas the Jetson showed no network activity. Conversely, the Jetson exhibited higher disk write bandwidth at \SI{18.7}{\Mbps}, compared to \SI{13.5}{\Mbps} on the NUC.
\begin{magic}[Software - Recipe]
    \item Co-development of Hardware and Software
    \item Testing of all Software on Prototype (Desk Setup)
    \item Ensuring Driver and Hardware Compatibility
    \item Adaptable and Modular Software Design
    \item Hardware Acceleration (Video Compression, GPU)
    \item Simplicity \& Maintainability \& Documentation
\end{magic}

\subsection{Limitations - \textit{Boxi} Design}
\label{sec:limitations}
Starting with the mechanical design, \textit{Boxi} lacks a robust rollover cage to protect the system under the full load of a legged robot tipping over. While the lid provides sufficient stiffness for mounting, it cannot absorb the energy generated during falls, especially under the robot's full weight.

To address this issue, the development of a newly designed protective rollover cage is essential. The cage should be fully decoupled from \textit{Boxi} and directly mounted onto the robot, offering enhanced protection and durability.
During our thermal simulation, we mainly considered the heat generated by the compute units. However, the heat generated by individual sensors is also non-neglectable and currently not taken into consideration. \looseness-1

While we can measure highly accurate time synchronization via \gls{ptp} offsets, starting the development with more verification tools in mind would have helped. The developed IMU alignment tool confirmed good time synchronization. 
Further, we did not implement verification, such as active LED targets, to validate the manufacturer's rolling shutter effect compensation. \looseness-1

The decision to use the Jetson Orin with the compact ConnectTech carrier board was motivated by its smaller form factor. However, the absence of real-time interrupt pins resulted in a Raspberry Pi being required to manage the STIM320 and ADIS IMUs. This introduced further complexities, including a custom Raspberry Pi PCB design and \gls{ptp} synchronization. In retrospect, avoiding the integration of IMUs such as the ADIS and STIM320 due to their limited ROS1/ROS2 driver support---would have significantly reduced integration efforts and further improved the maintainability of the software stack.
Comprehensive environmental testing, including shock and vibration resistance assessments, has yet to be performed. Although extrinsic and intrinsic calibrations have demonstrated stability over time, broader environmental validation remains a critical area for future work.

\section{Lessons Learned and Discussions}
\label{sec:lessons_learned}
\begin{leftenumerate}
    \item \textbf{Hardware and software should be jointly developed}: The interdependence of software and hardware turned out to be the most difficult part to get right. 
An exhaustive analysis of software and hardware requirements and the interdependence between components is essential.
\item \textbf{Select components for the application}: While technical specifications in datasets are important, having access to good software support and easy integration can be more important and justify additional cost, weight, size, or worse specifications.\looseness-1
\item \textbf{Time synchronization targeted for a specific application}: Time synchronization requirements might differ based on the application and the software. 
State-estimation and SLAM tasks might require \SI{}{\micro\second} or \SI{}{\milli\second} level time synchronization between LiDAR and IMUs.\looseness-1
\item \textbf{Develop software before hardware}: We recommend implementing the software early in the design process to identify any compatibility issues.\looseness-1
\item \textbf{Develop verification tools for everything}: How can you verify that the software functions as intended? The development of tools to ensure time synchronization and data integration is essential.\looseness-1
\item \textbf{Perform rapid-prototyping during development}: We recommend building a cheap mock-up setup and 3D-printed prototypes before moving to advanced CNC milling.
\item \textbf{Calibration targeted for you application}: Define your calibration requirements, and which tools and setups are required to achieve it. An overlapping FoV, use of cheap targets, or avoiding expensive calibration room setups was desirable for us.\looseness-1
\item \textbf{Keep it simple}: What is better than a ptp-enabled switch? Using no switch. We recommend using as few compute units, sensors, and mechanical components as possible. Only implement custom PCBs if they significantly simplify the design. The same applies to the software: \eg integrating ROS1 and ROS2 inherently increased the complexity. \looseness-1
\end{leftenumerate}

\section{Conclusion}
\label{sec:conclusion}
In conclusion, \textit{Boxi} has proven to be an invaluable tool, enabling the collection of the GrandTour dataset with high-quality ground truth reference data. Through the algorithmic performance analysis (\cref{sec:algo}), we demonstrated that sensor payload design should consider the target application, software availability, and accessible components. 

Out of the selected components, NovAtel SPAN CPT7 and Leica Geosystems components (custom AP20 and MS60) stood out as vital components, due to their excellent software support and performance, despite their cost. The design of the power distribution board also contributed to a clean assembly and minimized EMI problems, with the placement of the GNSS antenna being the only challenge we encountered.

Moreover, we plan an Autonomy \textit{Boxi} version with a reduced set of sensors, which will include the two LiDARs, three HDR cameras, and a single updated next-generation NVIDIA Jetson processor, alongside a \gls{ptp}-enabled switch for improved synchronization. These refinements will further enhance \textit{Boxi}'s capabilities, ensuring it remains a versatile and reliable platform for future research and applications in robotics. Lastly, a drawback of \textit{Boxi} is the lack of auxiliary modalities such as UWB, Radar, event, and thermal cameras, which can improve the usefulness and effectiveness of the sensor payload in various environments.

\bibliographystyle{plainnat}
\bibliography{references}

\begin{thebibliography}{77}
\providecommand{\natexlab}[1]{#1}
\providecommand{\url}[1]{\texttt{#1}}
\expandafter\ifx\csname urlstyle\endcsname\relax
  \providecommand{\doi}[1]{doi: #1}\else
  \providecommand{\doi}{doi: \begingroup \urlstyle{rm}\Url}\fi

\bibitem[nov(2024{\natexlab{a}})]{novatel_cpt7}
Cpt7 product sheet, 2024{\natexlab{a}}.
\newblock URL \url{https://ppmgmbh.com/Download/Produkte/NovAtel/SPAN_CPT7/Datenblatt/Datasheet_NovAtel_CPT7.pdf}.
\newblock Accessed: 2025-01-21.

\bibitem[nov(2024{\natexlab{b}})]{novatel_ie}
Inertial explorer, 2024{\natexlab{b}}.
\newblock URL \url{https://docs.novatel.com/Waypoint/Content/Inertial_Explorer/Overview_of_IE.htm}.
\newblock Accessed: 2025-01-21.

\bibitem[(2005)(2005)]{thermal_expansion2005}
The Engineering~ToolBox (2005).
\newblock Metals - temperature expansion coefficients, 2005.
\newblock URL \url{https://www.engineeringtoolbox.com/thermal-expansion-metals-d_859.html}.

\bibitem[Agarwal et~al.(2023)Agarwal, Mierle, and Team]{Agarwal_Ceres_Solver_2022}
Sameer Agarwal, Keir Mierle, and The Ceres~Solver Team.
\newblock {Ceres Solver}, 10 2023.
\newblock URL \url{https://github.com/ceres-solver/ceres-solver}.

\bibitem[Bloesch et~al.(2017)Bloesch, Burri, Sommer, Siegwart, and Hutter]{bloesch2017two}
Michael Bloesch, Michael Burri, Hannes Sommer, Roland Siegwart, and Marco Hutter.
\newblock The two-state implicit filter recursive estimation for mobile robots.
\newblock \emph{IEEE Robotics and Automation Letters}, 3\penalty0 (1):\penalty0 573--580, 2017.

\bibitem[Brizi et~al.(2024)Brizi, Giacomini, Di~Giammarino, Ferrari, Salem, De~Rebotti, and Grisetti]{brizi2024vbr}
Leonardo Brizi, Emanuele Giacomini, Luca Di~Giammarino, Simone Ferrari, Omar Salem, Lorenzo De~Rebotti, and Giorgio Grisetti.
\newblock Vbr: A vision benchmark in rome.
\newblock \emph{arXiv preprint arXiv:2404.11322}, 2024.

\bibitem[Bry et~al.(2015)Bry, Richter, and Roy]{bry2015spatial}
A.~Bry, C.~Richter, and N.~Roy.
\newblock Spatial and temporal calibration of multi-sensor systems with application to stereo vision.
\newblock In \emph{IEEE International Conference on Robotics and Automation (ICRA)}, pages 3863--3870, 2015.

\bibitem[Buchanan(2021)]{AllanVarianceRos}
Russell Buchanan.
\newblock Allan variance ros, November 2021.
\newblock URL \url{https://github.com/ori-drs/allan_variance_ros}.

\bibitem[Burnett et~al.(2023)Burnett, Yoon, Wu, Li, Zhang, Lu, Qian, Tseng, Lambert, Leung, et~al.]{burnett2023boreas}
Keenan Burnett, David~J Yoon, Yuchen Wu, Andrew~Z Li, Haowei Zhang, Shichen Lu, Jingxing Qian, Wei-Kang Tseng, Andrew Lambert, Keith~YK Leung, et~al.
\newblock Boreas: A multi-season autonomous driving dataset.
\newblock \emph{The International Journal of Robotics Research}, 42\penalty0 (1-2):\penalty0 33--42, 2023.

\bibitem[Burri et~al.(2016)Burri, Nikolic, Gohl, Schneider, Rehder, Omari, Achtelik, and Siegwart]{burri2016euroc}
Michael Burri, Janosch Nikolic, Pascal Gohl, Thomas Schneider, Joern Rehder, Sammy Omari, Markus~W Achtelik, and Roland Siegwart.
\newblock The euroc micro aerial vehicle datasets.
\newblock \emph{The International Journal of Robotics Research}, 35\penalty0 (10):\penalty0 1157--1163, 2016.

\bibitem[Caesar et~al.(2020)Caesar, Bankiti, Lang, Vora, Liong, Xu, Krishnan, Pan, Baldan, and Beijbom]{caesar2020nuscenes}
Holger Caesar, Varun Bankiti, Alex~H Lang, Sourabh Vora, Venice~Erin Liong, Qiang Xu, Anush Krishnan, Yu~Pan, Giancarlo Baldan, and Oscar Beijbom.
\newblock nuscenes: A multimodal dataset for autonomous driving.
\newblock In \emph{Proceedings of the IEEE/CVF conference on computer vision and pattern recognition}, pages 11621--11631, 2020.

\bibitem[Carballo et~al.(2020)Carballo, Lambert, Monrroy, Wong, Narksri, Kitsukawa, Takeuchi, Kato, and Takeda]{carballo2020libre}
Alexander Carballo, Jacob Lambert, Abraham Monrroy, David Wong, Patiphon Narksri, Yuki Kitsukawa, Eijiro Takeuchi, Shinpei Kato, and Kazuya Takeda.
\newblock Libre: The multiple 3d lidar dataset.
\newblock In \emph{2020 IEEE intelligent vehicles symposium (IV)}, pages 1094--1101. IEEE, 2020.

\bibitem[Chang et~al.(2019)Chang, Lambert, Sangkloy, Singh, Bak, Hartnett, Wang, Carr, Lucey, Ramanan, et~al.]{chang2019argoverse}
Ming-Fang Chang, John Lambert, Patsorn Sangkloy, Jagjeet Singh, Slawomir Bak, Andrew Hartnett, De~Wang, Peter Carr, Simon Lucey, Deva Ramanan, et~al.
\newblock Argoverse: 3d tracking and forecasting with rich maps.
\newblock In \emph{Proceedings of the IEEE/CVF conference on computer vision and pattern recognition}, pages 8748--8757, 2019.

\bibitem[Chen et~al.(2023)Chen, Nemiroff, and Lopez]{dlio}
Kenny Chen, Ryan Nemiroff, and Brett~T Lopez.
\newblock Direct lidar-inertial odometry: Lightweight lio with continuous-time motion correction.
\newblock In \emph{2023 IEEE international conference on robotics and automation (ICRA)}, pages 3983--3989. IEEE, 2023.

\bibitem[Cordts et~al.(2015)Cordts, Omran, Ramos, Scharw{\"a}chter, Enzweiler, Benenson, Franke, Roth, and Schiele]{cordts2015cityscapes}
Marius Cordts, Mohamed Omran, Sebastian Ramos, Timo Scharw{\"a}chter, Markus Enzweiler, Rodrigo Benenson, Uwe Franke, Stefan Roth, and Bernt Schiele.
\newblock The cityscapes dataset.
\newblock In \emph{CVPR Workshop on the Future of Datasets in Vision}, volume~2, page~1, 2015.

\bibitem[Cramariuc et~al.(2020)Cramariuc, Petrov, Suri, Mittal, Siegwart, and Cadena]{cramariuc2020learning}
Andrei Cramariuc, Aleksandar Petrov, Rohit Suri, Mayank Mittal, Roland Siegwart, and Cesar Cadena.
\newblock Learning camera miscalibration detection.
\newblock In \emph{2020 IEEE International Conference on Robotics and Automation (ICRA)}, pages 4997--5003. IEEE, 2020.

\bibitem[Cvisic et~al.(2022)Cvisic, Markovic, and Petrovic]{cvisic2022}
Igor Cvisic, Ivan Markovic, and Ivan Petrovic.
\newblock Enhanced calibration of camera setups for high-performance visual odometry.
\newblock \emph{Robotics Auton. Syst.}, 155:\penalty0 104189, 2022.
\newblock URL \url{https://doi.org/10.1016/j.robot.2022.104189}.

\bibitem[Daum et~al.(2023)Daum, Vaidis, and Pomerleau]{daum2023benchmarking}
Effie Daum, Maxime Vaidis, and Fran{\c{c}}ois Pomerleau.
\newblock Benchmarking ground truth trajectories with robotic total stations.
\newblock \emph{arXiv preprint arXiv:2309.05134}, 2023.

\bibitem[Davis and Hu(2011)]{suitesparse}
Timothy~A. Davis and Yifan Hu.
\newblock The university of florida sparse matrix collection.
\newblock \emph{ACM Trans. Math. Softw.}, 38\penalty0 (1), December 2011.
\newblock ISSN 0098-3500.
\newblock \doi{10.1145/2049662.2049663}.
\newblock URL \url{https://doi.org/10.1145/2049662.2049663}.

\bibitem[De~Agostino et~al.(2010)De~Agostino, Manzino, and Piras]{de2010performances}
Mattia De~Agostino, Ambrogio~Maria Manzino, and Marco Piras.
\newblock Performances comparison of different mems-based imus.
\newblock In \emph{IEEE/ION Position, Location and Navigation Symposium}, pages 187--201. IEEE, 2010.

\bibitem[Del~Duca and Machado(2023)]{blk2go}
Graziella Del~Duca and Carol Machado.
\newblock Assessing the quality of the leica blk2go mobile laser scanner versus the focus 3d s120 static terrestrial laser scanner for a preliminary study of garden digital surveying.
\newblock \emph{Heritage}, 6\penalty0 (2):\penalty0 1007--1027, 2023.

\bibitem[Erni et~al.(2023)Erni, Frey, Miki, Mattamala, and Hutter]{erni2023mem}
Gian Erni, Jonas Frey, Takahiro Miki, Matias Mattamala, and Marco Hutter.
\newblock Mem: Multi-modal elevation mapping for robotics and learning.
\newblock In \emph{2023 IEEE/RSJ International Conference on Intelligent Robots and Systems (IROS)}, pages 11011--11018. IEEE, 2023.

\bibitem[Faizullin et~al.(2021)Faizullin, Kornilova, Akhmetyanov, and Ferrer]{twistnsync}
Marsel Faizullin, Anastasiia Kornilova, Azat Akhmetyanov, and Gonzalo Ferrer.
\newblock Twist-n-sync: Software clock synchronization with microseconds accuracy using mems-gyroscopes.
\newblock \emph{Sensors}, 21\penalty0 (1), 2021.
\newblock ISSN 1424-8220.
\newblock \doi{10.3390/s21010068}.
\newblock URL \url{https://www.mdpi.com/1424-8220/21/1/68}.

\bibitem[Feng et~al.(2024)Feng, Qi, Zhong, Chen, Chen, Chen, Wu, and Ma]{feng2024s3e}
Dapeng Feng, Yuhua Qi, Shipeng Zhong, Zhiqiang Chen, Qiming Chen, Hongbo Chen, Jin Wu, and Jun Ma.
\newblock S3e: A multi-robot multimodal dataset for collaborative slam.
\newblock \emph{IEEE Robotics and Automation Letters}, 2024.

\bibitem[Fu et~al.(2023)Fu, Chebrolu, and Fallon]{diffcal}
Lanke Frank~Tarimo Fu, Nived Chebrolu, and Maurice Fallon.
\newblock Extrinsic calibration of camera to lidar using a differentiable checkerboard model.
\newblock In \emph{2023 IEEE/RSJ International Conference on Intelligent Robots and Systems (IROS)}, pages 1825--1831, 2023.
\newblock \doi{10.1109/IROS55552.2023.10341781}.

\bibitem[Furgale et~al.(2013)Furgale, Rehder, and Siegwart]{furgale2013kalibr}
Paul Furgale, Joern Rehder, and Roland Siegwart.
\newblock Unified temporal and spatial calibration for multi-sensor systems.
\newblock In \emph{2013 IEEE/RSJ International Conference on Intelligent Robots and Systems}, pages 1280--1286. IEEE, 2013.

\bibitem[Gehrig et~al.(2021)Gehrig, Aarents, Gehrig, and Scaramuzza]{gehrig2021dsec}
Mathias Gehrig, Willem Aarents, Daniel Gehrig, and Davide Scaramuzza.
\newblock Dsec: A stereo event camera dataset for driving scenarios.
\newblock \emph{IEEE Robotics and Automation Letters}, 6\penalty0 (3):\penalty0 4947--4954, 2021.

\bibitem[Geiger et~al.(2013)Geiger, Lenz, Stiller, and Urtasun]{geiger2013vision}
Andreas Geiger, Philip Lenz, Christoph Stiller, and Raquel Urtasun.
\newblock Vision meets robotics: The kitti dataset.
\newblock \emph{The International Journal of Robotics Research}, 32\penalty0 (11):\penalty0 1231--1237, 2013.

\bibitem[Grupp(2017)]{grupp2017evo}
Michael Grupp.
\newblock evo: Python package for the evaluation of odometry and slam.
\newblock \url{https://github.com/MichaelGrupp/evo}, 2017.

\bibitem[Hagemann et~al.(2022)Hagemann, Knorr, Janssen, and Stiller]{hagemann2022inferring}
Annika Hagemann, Moritz Knorr, Holger Janssen, and Christoph Stiller.
\newblock Inferring bias and uncertainty in camera calibration.
\newblock \emph{International Journal of Computer Vision}, pages 1--16, 2022.

\bibitem[Helmberger et~al.(2022)Helmberger, Morin, Berner, Kumar, Cioffi, and Scaramuzza]{helmberger2022hilti}
Michael Helmberger, Kristian Morin, Beda Berner, Nitish Kumar, Giovanni Cioffi, and Davide Scaramuzza.
\newblock The hilti slam challenge dataset.
\newblock \emph{IEEE Robotics and Automation Letters}, 7\penalty0 (3):\penalty0 7518--7525, 2022.

\bibitem[Hu et~al.(2024)Hu, Zheng, Wu, Geng, Yu, Wei, Tang, Wang, Jiao, and Liu]{hu2024paloc}
Xiangcheng Hu, Linwei Zheng, Jin Wu, Ruoyu Geng, Yang Yu, Hexiang Wei, Xiaoyu Tang, Lujia Wang, Jianhao Jiao, and Ming Liu.
\newblock Paloc: Advancing slam benchmarking with prior-assisted 6-dof trajectory generation and uncertainty estimation.
\newblock \emph{IEEE/ASME Transactions on Mechatronics}, 2024.

\bibitem[Hudson et~al.(2021)Hudson, Talbot, Cox, Williams, Hines, Pitt, Wood, Frousheger, Surdo, Molnar, et~al.]{hudson2021heterogeneous}
Nicolas Hudson, Fletcher Talbot, Mark Cox, Jason Williams, Thomas Hines, Alex Pitt, Brett Wood, Dennis Frousheger, Katrina~Lo Surdo, Thomas Molnar, et~al.
\newblock Heterogeneous ground and air platforms, homogeneous sensing: Team csiro data61's approach to the darpa subterranean challenge.
\newblock \emph{arXiv preprint arXiv:2104.09053}, 2021.

\bibitem[Hutter et~al.(2017)Hutter, Gehring, Lauber, Gunther, Bellicoso, Tsounis, Fankhauser, Diethelm, Bachmann, Bl{\"o}sch, et~al.]{anymal}
Marco Hutter, Christian Gehring, Andreas Lauber, Fabian Gunther, Carmine~Dario Bellicoso, Vassilios Tsounis, P{\'e}ter Fankhauser, Remo Diethelm, Samuel Bachmann, Michael Bl{\"o}sch, et~al.
\newblock Anymal-toward legged robots for harsh environments.
\newblock \emph{Advanced Robotics}, 31\penalty0 (17):\penalty0 918--931, 2017.

\bibitem[Jelavic et~al.(2021)Jelavic, Jud, Egli, and Hutter]{jelavic2021towards}
Edo Jelavic, Dominic Jud, Pascal Egli, and Marco Hutter.
\newblock Towards autonomous robotic precision harvesting: Mapping, localization, planning and control for a legged tree harvester.
\newblock \emph{arXiv preprint arXiv:2104.10110}, 2021.

\bibitem[Jiao et~al.(2022)Jiao, Wei, Hu, Hu, Zhu, He, Wu, Yu, Xie, Huang, et~al.]{jiao2022fusionportable}
Jianhao Jiao, Hexiang Wei, Tianshuai Hu, Xiangcheng Hu, Yilong Zhu, Zhijian He, Jin Wu, Jingwen Yu, Xupeng Xie, Huaiyang Huang, et~al.
\newblock Fusionportable: A multi-sensor campus-scene dataset for evaluation of localization and mapping accuracy on diverse platforms.
\newblock In \emph{2022 IEEE/RSJ International Conference on Intelligent Robots and Systems (IROS)}, pages 3851--3856. IEEE, 2022.

\bibitem[Kannala and Brandt(2006)]{kannalbrandt2006}
Juho Kannala and Sami Brandt.
\newblock A generic camera model and calibration method for conventional, wide-angle, and fish-eye lenses.
\newblock \emph{IEEE transactions on pattern analysis and machine intelligence}, 28:\penalty0 1335--40, 09 2006.
\newblock \doi{10.1109/TPAMI.2006.153}.

\bibitem[{Leica Geosystems}()]{prism}
{Leica Geosystems}.
\newblock Leica grz101 360° mini reflector.
\newblock \url{https://shop.leica-geosystems.com/de-DE/survey/accessory/geosystems/leica-grz101-360-degree-mini-reflector/buy}.
\newblock Accessed: 2025-01-20.

\bibitem[{Leica Geosystems}(2020)]{ms60}
{Leica Geosystems}.
\newblock Leica nova ms60 multistation.
\newblock \url{https://leica-geosystems.com/products/total-stations/multistation/leica-nova-ms60}, 2020.
\newblock Accessed: 2025-01-20.

\bibitem[{Leica Geosystems AG}(2024)]{leica_blk2go}
{Leica Geosystems AG}.
\newblock Leica {BLK2GO} - {Handheld} {Laser} {Scanner}, 2024.
\newblock URL \url{https://shop.leica-geosystems.com/global/leica-blk/blk2go/buy}.
\newblock Accessed: 19 Oct 2024.

\bibitem[Leutenegger(2022)]{leutenegger2022okvis2}
Stefan Leutenegger.
\newblock Okvis2: Realtime scalable visual-inertial slam with loop closure.
\newblock \emph{arXiv preprint arXiv:2202.09199}, 2022.

\bibitem[Li et~al.(2023)Li, Wu, Yang, Zou, Yang, Zhao, and Dong]{li2023whu}
Jianping Li, Weitong Wu, Bisheng Yang, Xianghong Zou, Yandi Yang, Xin Zhao, and Zhen Dong.
\newblock Whu-helmet: A helmet-based multi-sensor slam dataset for the evaluation of real-time 3d mapping in large-scale gnss-denied environments.
\newblock \emph{IEEE Transactions on Geoscience and Remote Sensing}, 2023.

\bibitem[Li et~al.(2021)Li, Li, and Hanebeck]{li2021towards}
Kailai Li, Meng Li, and Uwe~D Hanebeck.
\newblock Towards high-performance solid-state-lidar-inertial odometry and mapping.
\newblock \emph{IEEE Robotics and Automation Letters}, 6\penalty0 (3):\penalty0 5167--5174, 2021.

\bibitem[Liu et~al.(2024)Liu, Fu, Qin, Xu, Xu, Chen, Goossens, Sun, Yu, Liu, et~al.]{liu2024botanicgarden}
Yuanzhi Liu, Yujia Fu, Minghui Qin, Yufeng Xu, Baoxin Xu, Fengdong Chen, Bart Goossens, Poly~ZH Sun, Hongwei Yu, Chun Liu, et~al.
\newblock Botanicgarden: A high-quality dataset for robot navigation in unstructured natural environments.
\newblock \emph{IEEE Robotics and Automation Letters}, 2024.

\bibitem[Maddern et~al.(2016)Maddern, Stewart, Linegar, and Newman]{maddern2016real}
W.~Maddern, A.~Stewart, C.~Linegar, and P.~Newman.
\newblock Real-time kinematic ground truth for the oxford robotcar dataset.
\newblock In \emph{IEEE International Conference on Robotics and Automation (ICRA)}, pages 5279--5286, 2016.

\bibitem[Maddern et~al.(2017)Maddern, Pascoe, Linegar, and Newman]{maddern2017}
Will Maddern, Geoffrey Pascoe, Chris Linegar, and Paul Newman.
\newblock 1 year, 1000 km: The oxford robotcar dataset.
\newblock \emph{The International Journal of Robotics Research}, 36\penalty0 (1):\penalty0 3--15, 2017.

\bibitem[Nguyen et~al.(2024)Nguyen, Yuan, Nguyen, Yin, Cao, Xie, Wozniak, Jensfelt, Thiel, Ziegenbein, and Blunder]{mcdviral2024}
Thien-Minh Nguyen, Shenghai Yuan, Thien~Hoang Nguyen, Pengyu Yin, Haozhi Cao, Lihua Xie, Maciej Wozniak, Patric Jensfelt, Marko Thiel, Justin Ziegenbein, and Noel Blunder.
\newblock Mcd: Diverse large-scale multi-campus dataset for robot perception.
\newblock In \emph{Proceedings of the IEEE/CVF Conference on Computer Vision and Pattern Recognition}, pages 22304--22313, 6 2024.
\newblock URL \url{https://mcdviral.github.io/}.

\bibitem[Nikolic et~al.(2014)Nikolic, Rehder, Burri, Gohl, Leutenegger, Furgale, and Siegwart]{nikolic2014synchronized}
Janosch Nikolic, Joern Rehder, Michael Burri, Pascal Gohl, Stefan Leutenegger, Paul~T Furgale, and Roland Siegwart.
\newblock A synchronized visual-inertial sensor system with fpga pre-processing for accurate real-time slam.
\newblock In \emph{2014 IEEE international conference on robotics and automation (ICRA)}, pages 431--437. IEEE, 2014.

\bibitem[Nobili et~al.(2017)Nobili, Camurri, Barasuol, Focchi, Caldwell, Semini, and Fallon]{nobili2017heterogeneous}
Simona Nobili, Marco Camurri, Victor Barasuol, Michele Focchi, Darwin Caldwell, Claudio Semini, and Maurice Fallon.
\newblock Heterogeneous sensor fusion for accurate state estimation of dynamic legged robots.
\newblock In \emph{Robotics: Science and System XIII}, 2017.

\bibitem[Nubert et~al.(2022)Nubert, Walther, Khattak, and Hutter]{nubert2022learning}
Julian Nubert, Etienne Walther, Shehryar Khattak, and Marco Hutter.
\newblock Learning-based localizability estimation for robust lidar localization.
\newblock In \emph{2022 IEEE/RSJ International Conference on Intelligent Robots and Systems (IROS)}, pages 17--24. IEEE, 2022.

\bibitem[Nubert et~al.(2025)Nubert, Tuna, Frey, Cadena, Kuchenbecker, Khattak, and Hutter]{nubert2025holistic}
Julian Nubert, Turcan Tuna, Jonas Frey, Cesar Cadena, Katherine~J Kuchenbecker, Shehryar Khattak, and Marco Hutter.
\newblock Holistic fusion: Task-and setup-agnostic robot localization and state estimation with factor graphs.
\newblock \emph{arXiv preprint arXiv:2504.06479}, 2025.

\bibitem[Olson(2011)]{aprilgrid2011}
Edwin Olson.
\newblock Apriltag: A robust and flexible visual fiducial system.
\newblock In \emph{2011 IEEE International Conference on Robotics and Automation}, pages 3400--3407, 2011.
\newblock \doi{10.1109/ICRA.2011.5979561}.

\bibitem[Ramezani et~al.(2020)Ramezani, Wang, Camurri, Wisth, Mattamala, and Fallon]{ramezani2020newer}
Milad Ramezani, Yiduo Wang, Marco Camurri, David Wisth, Matias Mattamala, and Maurice Fallon.
\newblock The newer college dataset: Handheld lidar, inertial and vision with ground truth.
\newblock In \emph{2020 IEEE/RSJ International Conference on Intelligent Robots and Systems (IROS)}, pages 4353--4360. IEEE, 2020.

\bibitem[Ramezani et~al.(2022)Ramezani, Khosoussi, Catt, Moghadam, Williams, Borges, Pauling, and Kottege]{ramezani2022wildcat}
Milad Ramezani, Kasra Khosoussi, Gavin Catt, Peyman Moghadam, Jason Williams, Paulo Borges, Fred Pauling, and Navinda Kottege.
\newblock Wildcat: Online continuous-time 3d lidar-inertial slam.
\newblock \emph{arXiv preprint arXiv:2205.12595}, 2022.

\bibitem[Rehder et~al.(2016{\natexlab{a}})Rehder, Nikolic, Schneider, Hinzmann, and Siegwart]{kalibr_multi_axes2016}
Joern Rehder, Janosch Nikolic, Thomas Schneider, Timo Hinzmann, and Roland Siegwart.
\newblock Extending kalibr: Calibrating the extrinsics of multiple imus and of individual axes.
\newblock In \emph{2016 IEEE International Conference on Robotics and Automation (ICRA)}, pages 4304--4311, 2016{\natexlab{a}}.
\newblock \doi{10.1109/ICRA.2016.7487628}.

\bibitem[Rehder et~al.(2016{\natexlab{b}})Rehder, Nikolic, Schneider, Hinzmann, and Siegwart]{rehder2016extending}
Joern Rehder, Janosch Nikolic, Thomas Schneider, Timo Hinzmann, and Roland Siegwart.
\newblock Extending kalibr: Calibrating the extrinsics of multiple imus and of individual axes.
\newblock In \emph{2016 IEEE International Conference on Robotics and Automation (ICRA)}, pages 4304--4311. IEEE, 2016{\natexlab{b}}.

\bibitem[Reijgwart et~al.(2023)Reijgwart, Cadena, Siegwart, and Ott]{reijgwart2023wavemap}
Victor Reijgwart, Cesar Cadena, Roland Siegwart, and Lionel Ott.
\newblock Efficient volumetric mapping of multi-scale environments using wavelet-based compression.
\newblock In \emph{Robotics: Science and Systems (RSS)}, 2023.
\newblock \doi{10.15607/RSS.2023.XIX.065}.
\newblock URL \url{https://www.roboticsproceedings.org/rss19/p065.pdf}.
\newblock Code available at \url{https://github.com/ethz-asl/wavemap}.

\bibitem[Schmid and Hirschm{\"u}ller(2013)]{schmid2013stereo}
Korbinian Schmid and Heiko Hirschm{\"u}ller.
\newblock Stereo vision and imu based real-time ego-motion and depth image computation on a handheld device.
\newblock In \emph{2013 IEEE International Conference on Robotics and Automation}, pages 4671--4678. IEEE, 2013.

\bibitem[{Sevensense Robotics AG}(2022)]{sevensense_coreresearch}
{Sevensense Robotics AG}.
\newblock Core research development kit, 2022.
\newblock URL \url{https://github.com/sevensense-robotics/core_research_manual}.
\newblock Accessed: 19 Oct 2024.

\bibitem[Simas et~al.(2024)Simas, Di~Gregorio, Simoni, and Gatti]{simas2024parallel}
Henrique Simas, Raffaele Di~Gregorio, Roberto Simoni, and Marco Gatti.
\newblock Parallel pointing systems suitable for robotic total stations: Selection, dimensional synthesis, and accuracy analysis.
\newblock \emph{Machines}, 12\penalty0 (1):\penalty0 54, 2024.

\bibitem[Sivaprakasam et~al.(2024)Sivaprakasam, Maheshwari, Castro, Triest, Nye, Willits, Saba, Wang, and Scherer]{sivaprakasam2024tartandrive}
Matthew Sivaprakasam, Parv Maheshwari, Mateo~Guaman Castro, Samuel Triest, Micah Nye, Steve Willits, Andrew Saba, Wenshan Wang, and Sebastian Scherer.
\newblock Tartandrive 2.0: More modalities and better infrastructure to further self-supervised learning research in off-road driving tasks.
\newblock \emph{arXiv preprint arXiv:2402.01913}, 2024.

\bibitem[Smid and Matas(2019)]{smid2019rolling}
Matej Smid and Jiri Matas.
\newblock Rolling shutter camera synchronization with sub-millisecond accuracy.
\newblock \emph{arXiv preprint arXiv:1902.11084}, 2019.

\bibitem[Sun et~al.(2020)Sun, Kretzschmar, Dotiwalla, Chouard, Patnaik, Tsui, Guo, Zhou, Chai, Caine, et~al.]{sun2020scalability}
Pei Sun, Henrik Kretzschmar, Xerxes Dotiwalla, Aurelien Chouard, Vijaysai Patnaik, Paul Tsui, James Guo, Yin Zhou, Yuning Chai, Benjamin Caine, et~al.
\newblock Scalability in perception for autonomous driving: Waymo open dataset.
\newblock In \emph{Proceedings of the IEEE/CVF conference on computer vision and pattern recognition}, pages 2446--2454, 2020.

\bibitem[Tao et~al.(2024)Tao, Mu{\~n}oz-Ba{\~n}{\'o}n, Zhang, Wang, Fu, and Fallon]{tao2024oxford}
Yifu Tao, Miguel~{\'A}ngel Mu{\~n}oz-Ba{\~n}{\'o}n, Lintong Zhang, Jiahao Wang, Lanke Frank~Tarimo Fu, and Maurice Fallon.
\newblock The oxford spires dataset: Benchmarking large-scale lidar-visual localisation, reconstruction and radiance field methods.
\newblock \emph{arXiv preprint arXiv:2411.10546}, 2024.

\bibitem[Thalmann and Neuner(2024)]{thalmann2024sensor}
Tomas Thalmann and Hans Neuner.
\newblock Sensor fusion of robotic total station and inertial navigation system for 6dof tracking applications.
\newblock \emph{Applied Geomatics}, 16\penalty0 (4):\penalty0 933--949, 2024.

\bibitem[Tiderko et~al.(2016)Tiderko, Hoeller, and R{\"o}hling]{fkie_multimaster}
Alexander Tiderko, Frank Hoeller, and Timo R{\"o}hling.
\newblock \emph{The ROS Multimaster Extension for Simplified Deployment of Multi-Robot Systems}, pages 629--650.
\newblock Springer International Publishing, Cham, 2016.
\newblock ISBN 978-3-319-26054-9.
\newblock \doi{10.1007/978-3-319-26054-9_24}.
\newblock URL \url{https://doi.org/10.1007/978-3-319-26054-9_24}.

\bibitem[Tranzatto et~al.(2022)Tranzatto, Mascarich, Bernreiter, Godinho, Camurri, Khattak, Dang, Reijgwart, Loeje, Wisth, et~al.]{tranzatto2022cerberus}
Marco Tranzatto, Frank Mascarich, Lukas Bernreiter, Carolina Godinho, Marco Camurri, Shehryar Khattak, Tung Dang, Victor Reijgwart, Johannes Loeje, David Wisth, et~al.
\newblock Cerberus: Autonomous legged and aerial robotic exploration in the tunnel and urban circuits of the darpa subterranean challenge.
\newblock \emph{arXiv preprint arXiv:2201.07067}, page~3, 2022.

\bibitem[Triest et~al.(2022)Triest, Sivaprakasam, Wang, Wang, Johnson, and Scherer]{triest2022tartandrive}
Samuel Triest, Matthew Sivaprakasam, Sean~J Wang, Wenshan Wang, Aaron~M Johnson, and Sebastian Scherer.
\newblock Tartandrive: A large-scale dataset for learning off-road dynamics models.
\newblock In \emph{2022 International Conference on Robotics and Automation (ICRA)}, pages 2546--2552. IEEE, 2022.

\bibitem[Trzeciak et~al.(2023)Trzeciak, Pluta, Fathy, Alcalde, Chee, Bromley, Brilakis, and Alliez]{trzeciak2023conslam}
Maciej Trzeciak, Kacper Pluta, Yasmin Fathy, Lucio Alcalde, Stanley Chee, Antony Bromley, Ioannis Brilakis, and Pierre Alliez.
\newblock Conslam: Construction data set for slam.
\newblock \emph{Journal of Computing in Civil Engineering}, 37\penalty0 (3):\penalty0 04023009, 2023.

\bibitem[Tschopp et~al.(2020)Tschopp, Riner, Fehr, Bernreiter, Furrer, Novkovic, Pfrunder, Cadena, Siegwart, and Nieto]{tschopp2020versavis}
Florian Tschopp, Michael Riner, Marius Fehr, Lukas Bernreiter, Fadri Furrer, Tonci Novkovic, Andreas Pfrunder, Cesar Cadena, Roland Siegwart, and Juan Nieto.
\newblock Versavis—an open versatile multi-camera visual-inertial sensor suite.
\newblock \emph{Sensors}, 20\penalty0 (5):\penalty0 1439, 2020.

\bibitem[Tuna et~al.(2023)Tuna, Nubert, Nava, Khattak, and Hutter]{tuna2023x}
Turcan Tuna, Julian Nubert, Yoshua Nava, Shehryar Khattak, and Marco Hutter.
\newblock X-icp: Localizability-aware lidar registration for robust localization in extreme environments.
\newblock \emph{IEEE Transactions on Robotics}, 2023.

\bibitem[Tuna et~al.(2024)Tuna, Nubert, Pfreundschuh, Cadena, Khattak, and Hutter]{tuna2024informed}
Turcan Tuna, Julian Nubert, Patrick Pfreundschuh, Cesar Cadena, Shehryar Khattak, and Marco Hutter.
\newblock Informed, constrained, aligned: A field analysis on degeneracy-aware point cloud registration in the wild.
\newblock \emph{arXiv preprint arXiv:2408.11809}, 2024.

\bibitem[Vaidis et~al.(2024)Vaidis, Shahraji, Daum, Dubois, Gigu{\`e}re, and Pomerleau]{vaidis2024rts}
Maxime Vaidis, Mohsen~Hassanzadeh Shahraji, Effie Daum, William Dubois, Philippe Gigu{\`e}re, and Fran{\c{c}}ois Pomerleau.
\newblock Rts-gt: Robotic total stations ground truthing dataset.
\newblock In \emph{2024 IEEE International Conference on Robotics and Automation (ICRA)}, pages 17050--17056. IEEE, 2024.

\bibitem[Watt et~al.(2015)Watt, Achanta, Abubakari, Sagen, Korkmaz, and Ahmed]{watt2015understanding}
Steve~T Watt, Shankar Achanta, Hamza Abubakari, Eric Sagen, Zafer Korkmaz, and Husam Ahmed.
\newblock Understanding and applying precision time protocol.
\newblock In \emph{2015 Saudi Arabia Smart Grid (SASG)}, pages 1--7. IEEE, 2015.

\bibitem[Wu et~al.(2024)Wu, Sun, Wu, and Fang]{wu2024everysync}
Xuankang Wu, Haoxiang Sun, Rongguang Wu, and Zheng Fang.
\newblock Everysync: An open hardware time synchronization sensor suite for common sensors in slam.
\newblock In \emph{2024 IEEE/RSJ International Conference on Intelligent Robots and Systems (IROS)}, pages 12587--12593. IEEE, 2024.

\bibitem[Zhang et~al.(2018)Zhang, Liu, Tsai, Hu, Chu, and Zheng]{zhang2018pirvs}
Zhe Zhang, Shaoshan Liu, Grace Tsai, Hongbing Hu, Chen-Chi Chu, and Feng Zheng.
\newblock Pirvs: An advanced visual-inertial slam system with flexible sensor fusion and hardware co-design.
\newblock In \emph{2018 IEEE International Conference on Robotics and Automation (ICRA)}, pages 3826--3832. IEEE, 2018.

\bibitem[Zhao et~al.(2024)Zhao, Gao, Wu, Singh, Jiang, Sun, Sarawata, Qiu, Whittaker, Higgins, et~al.]{zhao2024subt}
Shibo Zhao, Yuanjun Gao, Tianhao Wu, Damanpreet Singh, Rushan Jiang, Haoxiang Sun, Mansi Sarawata, Yuheng Qiu, Warren Whittaker, Ian Higgins, et~al.
\newblock Subt-mrs dataset: Pushing slam towards all-weather environments.
\newblock In \emph{Proceedings of the IEEE/CVF Conference on Computer Vision and Pattern Recognition}, pages 22647--22657, 2024.

\end{thebibliography}

\newpage
\clearpage

\appendix

\section{Appendix}
In the appendix, we provide further details on the design and development of \textit{Boxi}.
\ifthenelse{\boolean{anonymous}}{
}{
\section*{Acknowledgments}
\label{sec:acknowledgments}
Jonas Frey and Julian Nubert are supported by the Max Planck ETH Center for Learning Systems. 

This work was supported and partially funded by Leica Geosystems, which is part of Hexagon. We would like to express our gratitude to Benjamin Müller, Benjamin Sch\"oll, Sprenger Bernhard, Eisenreich Stefan, and Jonas Nussdorfer for their invaluable expertise in networking, calibration, time-synchronization, AP20, and their assistance in the collection of calibration data.
We gratefully acknowledge the support of NovAtel Inc., Hexagon - Autonomy \& Positioning division, with special thanks to Ryan Dixon for his contributions.

In addition, this work was supported by the National Centre of Competence in Research Robotics (NCCR Robotics), the ETH RobotX research grant funded through the ETH Zurich Foundation, the European Union's Horizon 2020 research and innovation program under grant agreement No 101016970, No 101070405, and No 101070596, and an ETH Zurich Research Grant No. 21-1 ETH-27.
This work was supported by the Open Research Data Grant at ETH Zurich.
We acknowledge that the International Foundation High Altitude Research Stations Jungfraujoch and Gornergrat (HFSJG), 3012 Bern, Switzerland, made it possible for us to carry out our experiment(s) at the High Altitude Research Station at Jungfraujoch.
We also thank the custodians, Mrs. Daniela Bissig and Mr. Erich Bissig, guide Ms. Doris Graf Jud (Jungfraubahnen), and Annette Fuhrer (Jungfraubahnen) for the support of our activities during the Jungfraujoch trip. 

Our sincere appreciation goes to William Talbot for his work in integrating the ADIS IMU. We also thank Elena Krasnova for her assistance with Blender and for helping develop the new roll cage.
We are grateful to Flurin Schindele for his efforts in integrating the Box onto various robots, and to Victor Reijgwart for his assistance with integrating Wavemap.
We extend our thanks to Johann Schwab for his support in integrating the TierIV C1 cameras, and to Cyrill P\"{u}ntener for his expertise in network configurations.
We thank Ulrich Huemer for his design of the mechanical subsystems within \textit{Boxi}, and Thomas Mantel for his work on the mechanical integration of the Livox MID360 and Prism.
Lastly, we thank Laurin Schmid and Andreas Binkert for their contributions to the integration of various 3D-printed parts.

}

\subsection{Dataset Descriptions}
In this section, the details of each dataset is provided.
\label{app:dataset_desc}
\begin{itemize}
   \setlength{\itemsep}{2pt}
   
    \item \dataset{Hike:} The robot descends and then ascends on loose gravel in a very large environment where optimization degradation~\cite{tuna2024informed} might occur. The robot traverses \SI{330}{\meter} in \SI{405}{\second}.

    \item \dataset{Warehouse:} This dataset consists of the robot walking within a confined warehouse starting from the outdoor court. The environment features items, shelves, and a roll-up door for feature-based state estimation problems. The robot traverses \SI{202}{\meter} in \SI{370}{\second}.

    \item \dataset{Research Station:} Acting as a simple dataset, the robot walks outside a research station featuring industrial metal stairs on an elevated metal grid platform. In this dataset, the robot traverses \SI{210}{\meter} in \SI{262}{\second}.

    \item \dataset{Excavation Site:} In this outdoor dataset, the robot walks around an operating excavator. This dataset contains large stones, water pits, small dirt roads, and construction site containers, which bring unique features. The robot traverses \SI{262}{\meter} in \SI{200}{\second}.

    \item \dataset{Terrace:} This dataset covers the robot walking from a cog railway station onto a terrace towards a large university building where limited features are available for short-range sensors while features are in abundance for large-scale perception. The robot traverses \SI{200}{\meter} in \SI{283}{\second} seconds.\looseness-1

    \item \dataset{Demolished Building:} Walking from the outside into a search and rescue training facility through a demolished building with an open staircase and over multiple floors. The robot traverses \SI{290}{\meter} in \SI{368}{\second}.

    \item \dataset{Mountain Ascent:} The robot starts by walking the scaffolding stairs, crossing a train station, and then traversing a narrow, very steep hiking path in sunny conditions, which presents challenges based on exposure. The robot traverses \SI{300}{\meter} in \SI{384}{\second}.
\end{itemize}

\subsection{Reference Ground Truth Pose Generation}
\label{app:ground_truth}
The availability of a precise 6DoF ground truth robot poses as a reference is critical for evaluating many robotic applications, such as state estimation and SLAM performance.
Contrary to many previous works, our deployments are indoors and outdoors, spanning hundreds of meters, and require millimeter-level precision to evaluate LiDAR-based state estimation methods. These requirements directly render LiDAR-based and only RTK-GNSS based references unsuitable~\cite{geiger2013vision, burnett2023boreas, zhao2024subt}.
While using Motion Capture (MOCAP) systems can provide high precision ground truth pose estimates at high frequency, these systems are restricted to a small workspace~\cite{burri2016euroc, feng2024s3e}, making them unsuitable for large-scale deployments.
Many approaches~\cite{ramezani2020newer, tao2024oxford, trzeciak2023conslam, mcdviral2024, li2023whu} rely on stationary scanners to create an accurate representation of the environment and register LiDAR measurements against these maps. However, the registration accuracy is bounded by the sensor noise characteristics itself, the accuracy of the map, and the registration algorithm~\cite{mcdviral2024}. Moreover, iterative registration methods can not retrieve accurate poses in LiDAR-degraded environments~\cite{tuna2023x, nubert2022learning}, and the generation of accurate maps requires significant manual effort~\cite{zhao2024subt} or may not even be feasible in challenging dynamic environments, such as a forest on a windy day.
Recent works~\cite{vaidis2024rts, thalmann2024sensor, daum2023benchmarking} used one or more \glspl{tps} to obtain ground truth position estimates with millimeter-to-sub-millimeter accuracy~\cite{simas2024parallel}. A \gls{tps} is a device commonly employed in construction and surveying to measure the 3D position of a reflector target. By mounting a prism target on a robot \gls{tps} measurements of the robot's position can be acquired during direct line-of-sight from the \gls{tps} to the robot. 
\begin{figure}[t] 
    \centering 
    \includegraphics[width=1.0\linewidth]{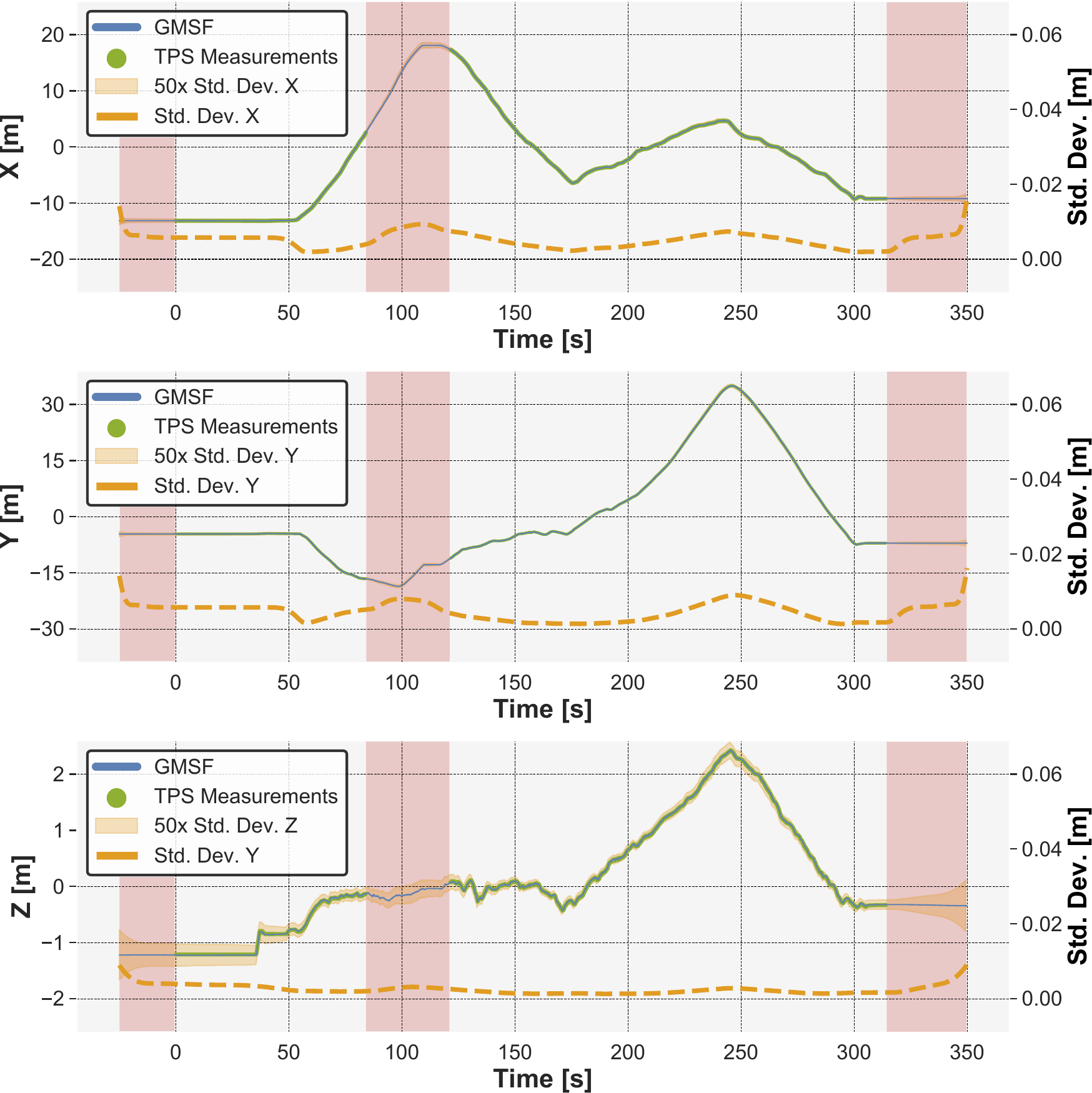} 
    \caption{Ground truth trajectory generated with Holistic Fusion (HF) for the \textit{Excavation Site} dataset is shown with the covariance-based uncertainty region overlaid. Given that the uncertainty is on a millimeter scale, we display 50x the standard deviation as an overlay and provide the numerical value on the right axis in the dotted orange line.
    The red regions indicate \gls{tps} measurement dropouts.
    }
    \label{fig:groundtruth_covariances} 
\end{figure}
To overcome the direct line-of-sight limitation, our solution fuses \gls{tps} measurements with precise IMU and GNSS measurements to obtain 6DoF ground truth poses. Specifically, we use a single Leica Geosystems MS60 Total Station~\cite{ms60}, and the Leica Geosystems \SI{360}{\degree} Mini-Prism GRZ101~\cite{prism}, which has a static accuracy of around $\pm$\SI{1.5}{\milli\meter} (3$\sigma$)  within \SI{360}{\degree} heading rotation and a pitch angle of $\pm$\SI{30}{\degree}.
To stream the data up to \SI{20}{\hertz} from the \gls{tps} to \textit{Boxi}, Leica Geosystems provided the Bluetooth radio handle RH18 together with an AP20 special edition. The AP20 runs a special firmware (not commercially available), which enables streaming of time-synchronized \gls{tps} measurements with time-sync jitter of $\pm$\SI{1}{\milli\second} (3$\sigma$).

The acquired \gls{tps} measurements are then fused with HG4930 IMU measurements, and post-processed GNSS poses. The GNSS solution is provided by the proprietary NovAtel Precise Positioning Product called Inertial Explorer~\cite{novatel_ie} and the NovAtel SPAN CPT7 GNSS receiver mounted on \textit{Boxi}. Using accurate atmospheric corrections, Inertial Explorer can provide a tightly-coupled pose solution with accuracies up to \SI{1}{\centi\meter}~\cite{novatel_ie}, which we will refer to as IE-TC. Inertial Explorer also produces loosely coupled (IE-LC) and a precise point positioning (IE-PPP) solution; NovAtel recommended that in most cases IE-TC is most precise solution. To fuse the \gls{tps} measurements and the IE-TC estimates, we first solve an online and then, an offline factor-graph optimization problem to obtain the fused highly accurate 6DoF robot pose, following a similar approach to PaLoc~\cite{hu2024paloc}. 
We build on the publicly available Holistic Fusion (HF) framework~\cite{nubert2025holistic} and refer to the offline factor-graph optimization output as our ground truth robot pose estimation. Details and code will be made available open-source.\looseness-1

\subsubsection{Ground Truth Verification}
For verification, we compare our ground truth poses against the \gls{tps} measurements. It is important to note that this analysis is conducted to show that the sensor fusion and factor-graph optimization steps do not alter the alignment accuracy against the \gls{tps} measurements but extend it with the HG4930 IMU measurements and the IE-TC odometry. As a result, the Holistic-Fusion-Optimized method is expected to produce the best position estimates against \gls{tps} measurements because it has privileged access to it. Here, extension refers to estimating the robot pose instead of a position even during the dropout of the \gls{tps} measurements.

The analysis is conducted for the \emph{Excavation Site} dataset, which has consistent GNSS coverage and \gls{tps} measurements throughout the deployment except for the beginning, the end, and \SIrange{84}{121}{\second} time period of the mission. 
The error metrics \gls{ate} and \gls{rte} are obtained using evo~\cite{grupp2017evo}. Crucially, as we compare the accuracy against the \gls{tps} measurements, the results provided in \Cref{tab:ground_truth_verification} do not reflect the accuracy of position during the dropout periods of the \gls{tps} measurements.
\begin{table}[t!]\centering
    \resizebox{1\columnwidth}{!}{
    \begin{tabular}{lcc}
    \toprule
    Method & \makecell{RTE \\ $\mu(\sigma)[\SI{}{\meter}]$ } & \makecell{ATE \\ $\mu(\sigma)[\SI{}{\meter}]$ }\\
    \midrule
    HF-Live & 0.0033(0.0028)                        & 0.0117(0.0305) \\[0.3em]
    HF-Optimized w/o IE & \textbf{0.0027}(\textbf{0.0021})& \textbf{0.0021}(\textbf{0.0012}) \\[0.3em]
    HF-Optimized & \textbf{0.0027}(\textbf{0.0021}) & \textbf{0.0021}(\textbf{0.0012}) \\[0.3em]
    IE-TC & \underline{0.0033}(\underline{0.0023})    & 0.0159(\underline{0.0069}) \\[0.3em]
    IE-LC & \underline{0.0033}(0.0023)                & \underline{0.0149}(0.0072) \\[0.3em]
    IE-PPP & 0.0356(0.0648)                           & 0.162(0.0118) \\[0.3em]
    \bottomrule
    \end{tabular}
    }
    \caption{\gls{ate} and \gls{rte} (per \SI{1}{\meter} distance traveled) for the \emph{Excavation Site} dataset are shown (best in \textbf{bold}), and the second best is \underline{underlined}. HS indicates the output of Holistic-Fusion in different operation modes.}
    \label{tab:ground_truth_verification}
\end{table}
\Cref{tab:ground_truth_verification} shows that the HF-Optimized and HF-Optimized w/o IE solutions perform identically and produce the least error against the \gls{tps} measurements. This is expected since we configured the Holistic Fusion (HF) framework to rely on the \gls{tps} measurements when available. As seen, the HS-Live method performs worse than HF-Optimized, which builds a graph or slow batch optimization to minimize the error. Moreover, both the IE-LC and IE-TC methods have higher \gls{ate} errors than Holistic-Fusion, which uses highly accurate \gls{tps} measurements. Lastly, IE-PPP performs the worst as this method relies only on the satellite-based GNSS correction. In terms of \gls{rte}, which is a measure of trajectory smoothness, the methods perform similarly, while HF-Optimized methods perform slightly better. 

During the previously mentioned drop out of \gls{tps} measurements, the HF-Optimized relies on the IE-TC poses instead of \gls{tps}. To highlight this case, the HF-Optimized output is compared against the IE-TC output, and the occlusion period is highlighted in red. \Cref{fig:groundtruth_covariances} shows the generated trajectories. As seen, the \emph{marginal} covariance estimation correlates with the mismatch between the \gls{tps} and IE-TC measurements, and the uncertainty increases in the absence of the \gls{tps} measurements, implying deviation from the absolute true position. The marginal covariance is a measure of how well the measurements align with each other and not the true covariances. %
\subsection{Sevensense CoreResearch}
\label{app:camera1} 
The Sevensense CoreResearch is a global shutter camera unit introduced in 2020~\cite{sevensense_coreresearch}. 
The unit supports up to 8 cameras, with a resolution of \SI{0.4}{\megapixel} or \SI{1.6}{\megapixel}.
It is connected via Gigabit Ethernet (1000BASE-T, IEEE 802.3ab) and \gls{ptp} time-synchronized. The cameras support hardware triggering full exposure time control and synchronization between cameras. 
For \textit{Boxi}, we chose to use $5$ $\times$ \SI{1.6}{\megapixel} cameras ($1440$ $\times$ $1080$) with three color and two monochrome cameras, with the two monochrome cameras positioned in a stereo setup. 
All cameras are equipped with \SI{2.4}{\milli\meter} focal length \SI{165.4}{\degree} horizontal \gls{fov} lenses and run at a maximum rate of around \SI{10}{\hertz} due to bandwidth limitations on the Ethernet interface despite efforts to optimize it.
This unit has been proven reliable and highly valued in academic research, being used for the latest SuBT Challenge~\cite{tranzatto2022cerberus} and popular datasets~\cite{helmberger2022hilti, ramezani2020newer, tao2024oxford}. 
The manufacturer provides well-documented, closed-source ROS1 drivers, but ROS2 drivers are not yet available. Since the provided driver also lacked support for debayering, white balancing, and color correction, we developed an in-house solution to address this.\looseness-1

\subsection{StereoLabs ZED2i}
\label{app:camera2} 
The StereoLabs ZED2i is a \SI{12}{\centi\meter} baseline stereo rolling shutter camera unit introduced in 2023 marketed specifically for outdoor robotic operation.
Through the ROS1/2 drivers provided by StereoLabs, the camera provides classical and learning-based depth estimation, but it requires an NVIDIA GPU (CUDA support) for operation. In addition, the driver provides barometric and thermal measurements at \SI{100}{\hertz} and IMU measurements at \SI{400}{\hertz} during real-time operation.
The camera is connected via USB 3.0 technology, and thus, time-synchronization to the host PC (USB buffer arrival timestamp) or hardware triggering is not supported.
It can provide image resolution up to 2K ($2208$ $\times$ $1242$) and comes with different lens options. We chose the wide \gls{fov} \SI{110}{\degree} horizontal version with a \SI{2.1}{\milli\meter} focal length lens.
Crucially, the driver provided by StereoLabs allows video encoding in H.264, H.265, or lossless formats and supports serialized recording of the measurements with exceptional efficiency. As an exception, the serialized format can record the IMU measurements only up to a rate of \SI{45}{\hertz} as of SDK version 4.2.5.\looseness-1

\subsection{TierIV C1} 
\label{app:camera3} 
The Tier IV C1 is a \SI{120}{\decibel} high dynamic range and rolling shutter camera with \SI{2.5}{\megapixel} chip resulting in ($1920$ $\times$ $1280$) images at \SI{30}{\hertz} with a horizontal \gls{fov} of \SI{120}{\degree}. The camera can be connected through GMSL2. The exact read time per line of the image is provided by the manufacturer, and the camera can be hardware-triggered via an f-sync signal. The camera performs automatic white-balancing and TierIV provides calibration tools as well as a custom lens correction model. 
Integration of the camera requires a sophisticated understanding of the Linux device tree and kernel modules to set up hardware triggering. 

\subsection{LiDARs} 
\label{app:lidar}
Given the above attributes, we integrated the Livox Mid360, a research community favorite for its affordability and unique scan pattern, making it ideal for near-field mapping due to its lower range (\SI{40}{\meter}) and high disparity (multiple \SI{}{\centi\meter} at \SI{20}{\meter}). 
For mapping and localization, we integrated the Hesai XT32 sparse LiDAR, with a range of \SI{0.5}{\meter} to \SI{120}{\meter} and high point accuracy of \SI{1}{\centi\meter}. 
Both LiDARs have well-maintained drivers, \gls{ptp} synchronization, intensity returns, and Ethernet connectivity. Both LiDARs are configured to aggregate measurements over a \SI{100}{\milli\second} period and provide point clouds at \SI{10}{\hertz}. 
A comparison of various spinning LiDARs can be found in~\cite{carballo2020libre} and~\cite{li2021towards} compares a solid-state LiDAR to a spinning LiDAR.\looseness-1

\subsection{IMUs}
\label{app:imu}
For \textit{Boxi}, we choose to integrate three tiers of IMUs.
Two (high-end) tactical-grade MEMS IMUs, namely the HG4930 and the STIM320. 
For the mid-range IMU, we chose the ADIS16475-2 IMU, where the predecessor model has been used for the collection of popular existing datasets~\cite{burri2016euroc}. 
The AP20-IMU is also a mid-range industrial grade IMU, however it only provides measurements, while connected to the Leica Geosystems MS60 Total Station via Bluetooth.
The Livox, CoreResearch, and the ZED2i camera all ship with integrated low-cost, consumer-grade IMUs (see. \cref{tab:components}). 
We operate the IMUs on \textit{Boxi} at a range from $100$-\SI{500}{\hertz}, excluding the ZED2i IMU dropout due to serialization.

\subsection{GNSS}
\label{app:gnss}
We use the NovAtel SPAN CPT7 industrial inertial navigation solution, which, though expensive (see \cref{tab:components}), offers a compact design, features one of the best commercially available MEMS IMU (see HG4930 in \Cref{tab:components}), good ROS1 drivers, and provides Ethernet connectivity.
The tactile grade HG4930 IMU within the CPT7 is pinned down, and the CPT7 internally uses the factory extrinsic calibration provided by Honeywell.
Critically, CPT7 functions as \textit{Boxi}'s Grandmaster clock of the \gls{ptp} network, internally synchronizing with the received GPS time when available. In addition to providing the current GNSS solution and IMU measurements in real-time, the CPT7 allows for onboard data logging.
The log files can be post-processed using Inertial Explorer to correct the GNSS measurements in hindsight through precise multi-constellation atmospheric corrections, obtained from the Leica Geosystems HGxN SmartNet GNSS correction service.
Furthermore, we use NovAtel TerraStar-C Pro Multi-Constellation Correction \gls{ppp} solution, which requires L-Band satellite communication to provide accurate pose estimation without relying on base station measurements during deployment.

\subsection{Network Switch}
\label{app:switch}
According to these criteria, we selected the cost-efficient UbiSwitch Module with UbiSwitch Baseboard that supports three 10GBASE-T and 8 1GBASE-T standard connections. This switch meets most requirements, specifically the excellent form factor and rugged design. However, it lacks IEEE 1588v2 support (see \cref{sec:conclusion}). 

\subsection{Compute}
\label{app:compute}
For handling CPU-intensive tasks, we chose the industrial AsRock Intel i5 NUC-1340P/D5, which offers a strong balance between performance and power efficiency, with excellent single-core performance; the ITX motherboard supports dual \SI{2.5}{\Mbps} Ethernet ports, \SI{64}{\giga\byte} of RAM, and a \SI{4}{\tera\byte} capacity M.2 form factor SSD.\looseness-1

For GPU-based computing, we selected the NVIDIA AGX Jetson Orin compute module paired with the ConnectTech Rogue Carrier Board and the ConnectTech GMSL2 adapter board.  
The carrier board is equipped with dual \SI{10}{\Gbps} Ethernet, two USB 3.2 connections, \SI{64}{\giga\byte} unified RAM, and a \SI{2}{\tera\byte} M.2 SSD. 
However, despite supporting I2C, SPI, and dual UART interfaces, the ConnectTech Rogue Carrier Board exposes an insufficient number of interrupt pins for hardware timestamping. 
To address this gap, we integrated a Raspberry Pi Compute Module 4 to handle timestamping for two IMUs, manage status LEDs and buttons, and control fan speed. 
All computers support \gls{ptp} time synchronization to efficiently synchronize with all sensors that support the protocol.
We emphasize that our compute choices should not be directly applied to commercial systems without careful consideration. In commercial applications, where custom software development and other operating systems are feasible, \eg more power-efficient ARM-based chipsets may be preferable. %
In hindsight, despite the better form factor of the ConnectTech carrier board, it would have been easier to integrate Nvidia's Developer Kit, reducing the number of computers to a minimum. 
\subsection{Cookbook: Time Synchronization Basics}
\label{app:sync_basics}
To understand the relationship between time synchronization and position estimation, we compute the resulting position error for different time synchronization errors for robots moving at different constant velocities, under the assumption of a fully observable state. 
For a robot moving at \SI{1}{\meter\per\second}, achieving millimeter-level position estimation accuracy requires time synchronization errors below \SI{10}{\micro\second}.
The position estimation error \( e \) due to time synchronization error \( \Delta t \) can be approximated as: \( e = v \cdot \Delta t \), where \( v \) is the robot's velocity and \( \Delta t \) is the time synchronization error. 
Similar computations can be performed for the angular velocity.
As a further first principle example of the effects of time-offsets, we consider a LiDAR measurement at a distance of \SI{25}{\meter} while the robot rotates at an angular velocity of \SI{200}{\degree\per\second}. 
A timestamp offset of \SI{1}{\milli\second} will lead to an offset of \SI{0.2}{\degree} or around \SI{8.7}{\centi\meter} error. 
In comparison, the vertical beam divergence of the Hesai is \SI{0.047}{\degree} horizontally.

Therefore, considering communication to sensors, delays in TCP/IP networks, USB controllers, serialization, and kernel scheduling can easily exceed the time precision required for millimeter-precise localization. 
This highlights the clear need for hardware timestamping of all sensor measurements and is particularly crucial for the long-range LiDARs, where each point must be accurately time-stamped. 

\subsection{Cookbook: Time Synchronization PTP}
\label{app:ptp}
\begin{figure}[t]
    \centering
    \includegraphics[width=1.0\linewidth]{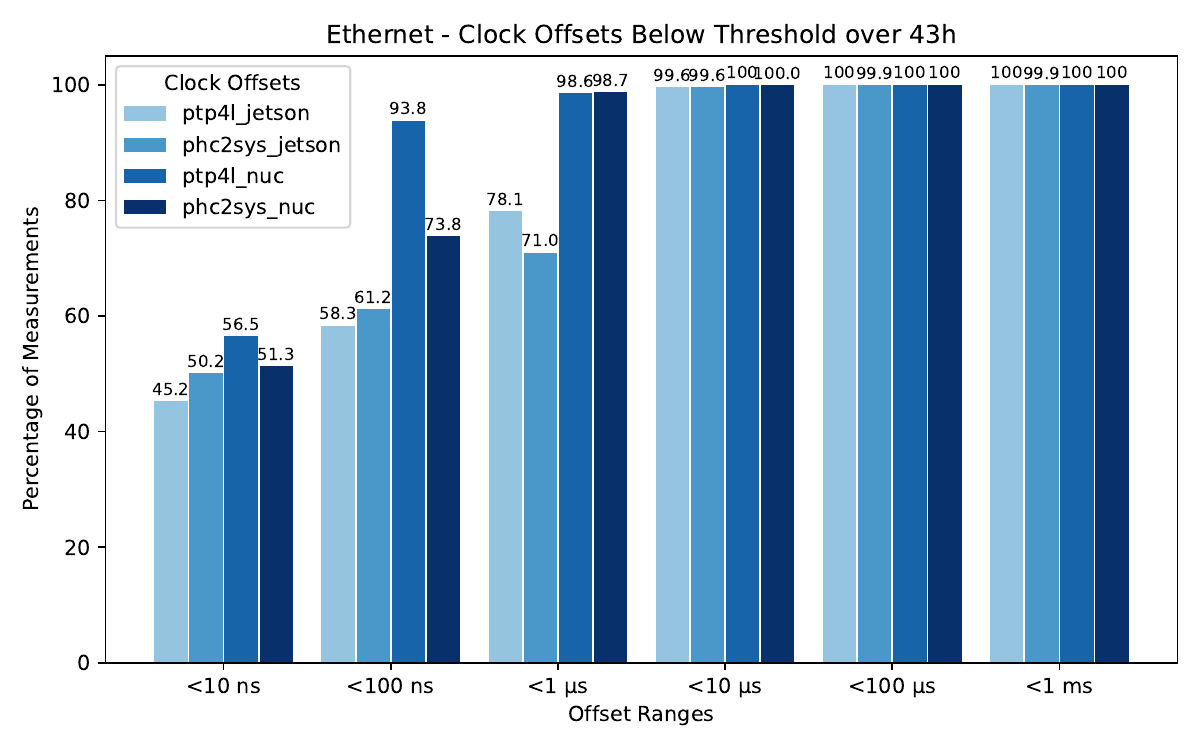}
    \caption{\gls{ptp} synchronization experiment of Jetson and Nuc synchronizing against CPT7 (the \gls{ptp} Grandmaster). We report \texttt{ptp4l} and \texttt{phc2sys} offsets calculated as a percentage of samples within a given time synchronization range.}
    \label{fig:ptp_offset}
    \vspace{-0.5cm}
\end{figure}

To analyze the synchronization between the PCs we logged the offsets of \texttt{ptp4l} and \texttt{phc2sys} every second and summarized the results in \cref{fig:ptp_offset}. 
The \texttt{ptp4l} offset represents the difference between the built-in clock of the network interface card (NIC) of the respective Ethernet port and the CPT7 Ethernet interface Grandmaster clock. The \texttt{phc2sys} offset, on the other hand, measures the discrepancy between the Ethernet interface clock and the system time of the respective PC. This \texttt{phc2sys} offset consistently remains below \SI{10}{\micro\second}.
The absence of IEEE 1588v2 standard synchronization can, in rare cases of packet loss, lead to incorrect estimations of the line delay. This, in turn, may result in improper clock adjustments, manifesting as outlier measurements seen in the \texttt{phc2sys} offset for the Jetson exceeding \SI{1}{\milli\second} in rare events.
Peak offset readings do not inherently indicate poor clock synchronization, as the proportional-derivative (PD) controller compensates for such offsets. Similarly, inaccurate line delay measurements do not directly imply a failure in time synchronization.
Our experiment remained within the normal bandwidth limits of the network switch, so the impact of bandwidth saturation was not assessed.
While integrating a low-drift MasterClock was considered, it was excluded due to CPT7's inability to function as a time receiver and overall weight constraints.
\subsection{Cookbook: Time Synchronization IMUs}
\label{app:time_sync_imus}
Time synchronization for the STIM320 IMU is handled via custom kernel modules that use GPIO interrupts triggered on falling edges (\texttt{IRQF\_TRIGGER\_FALLING}). Each interrupt invokes a kernel-space ISR that captures precise timestamps using \texttt{ktime\_get\_real\_ts64()} and stores them in a ring buffer. These timestamps are accessed from userspace via a \textit{sysfs} interface, allowing safe retrieval and clearing by the front-end driver for integration with ROS. This setup requires kernel-level expertise in GPIOs, IRQ handling, timestamping, and \textit{sysfs} design. Fixed sensor latencies, such as low-pass filter delays, must be compensated manually.
In hindsight, synchronizing the IMU to an external trigger would offer cleaner integration than free-running operation. By contrast, the ADIS IMU leverages the Linux IIO subsystem, which natively supports timestamping. However, enabling this functionality on the Raspberry Pi Compute Module 4 requires non-trivial kernel modifications and device tree configuration.

\subsection{Cookbook: Time Synchronization Validation}
\label{app:time_sync_validation}
We developed a time synchronization verification tool that calculates the time offset between data streams of IMU measurements. 
It first transforms the angular velocity readings of the IMUs into a common reference frame by compensating for different orientations using their respective extrinsic calibration. In the second step, a single axis of the transformed measurements is selected, and the signal is linearly interpolated to a frequency of \SI{500}{\hertz}, the maximum rate collected with \textit{Boxi}. A correlation analysis between the two angular velocity signals provides an initial time offset estimate between the IMUs. 
We then refine this estimate using gradient-based optimization. This involves iteratively interpolating the original measurement signals based on the current time-offset estimate. We minimize the mean squared error (MSE) between the two interpolated IMU signals until convergence.\looseness-1

Initially, we conceived this tool as a proof of concept to verify that IMU sensor readings fall within the expected IMU rate intervals of \SI{10}{\milli\second} based on the HG4930 reference IMU operating at a \SI{100}{\hertz}.

The results of applying this tool across all recorded missions is provided in the main paper \cref{tab:alignment}.
We applied the tool to three \SI{30}{\second} snippets per axis for each mission to estimate the standard deviation. 
Since the missions were recorded across seven different days, we assume that any remaining biases originate from the internal IMU-specific triggering time. Surprisingly, we find that we can estimate the IMU offset with high accuracy, achieving an average sigma of \SI{0.5}{\milli\second} when averaging the estimated offset across runs, different axes, and different time intervals across deployments. The remaining time offsets most likely stem from the variations in the triggering of the IMU exposure time and internal IMU delays. 
This analysis highlights the importance of having robust verification tools to ensure high-quality sensor data and to understand biases within the data. The STIM320 IMU is excluded from the above analysis as during testing, it was observed that the userspace front-end might have an initial time offset while reading the \textit{sysfs} virtual file. This offset is constant for each sequence and is characterized by the period of the IMU \SI{2}{\milli\second}. This offset is compensated in hindsight with the tool discussed in this section.\looseness-1

\subsection{Cookbook: Hardware}
\label{app:advanced_hardware}

\subsubsection{Vibrations}
During operation, \textit{Boxi} experiences constant vibrations due to the radial fan and the motion of the legged robot. 
Using vibration-damping mounts or isolators between the base platform and \textit{Boxi} is prohibited by the fact that the kinematic chain must remain accurately observable.
Internal vibrations can also arise from rotating components, specifically the radial fan and internally spinning LiDARs.
These vibrations can interfere with the readings from the IMUs; therefore, securely mounting all components is essential. 
Similarly, screw-in or clip-in connectors can help prevent loose connections and prevent component failures. %

\begin{figure*}[t] 
    \centering 
    \includegraphics[width=1.0\linewidth]{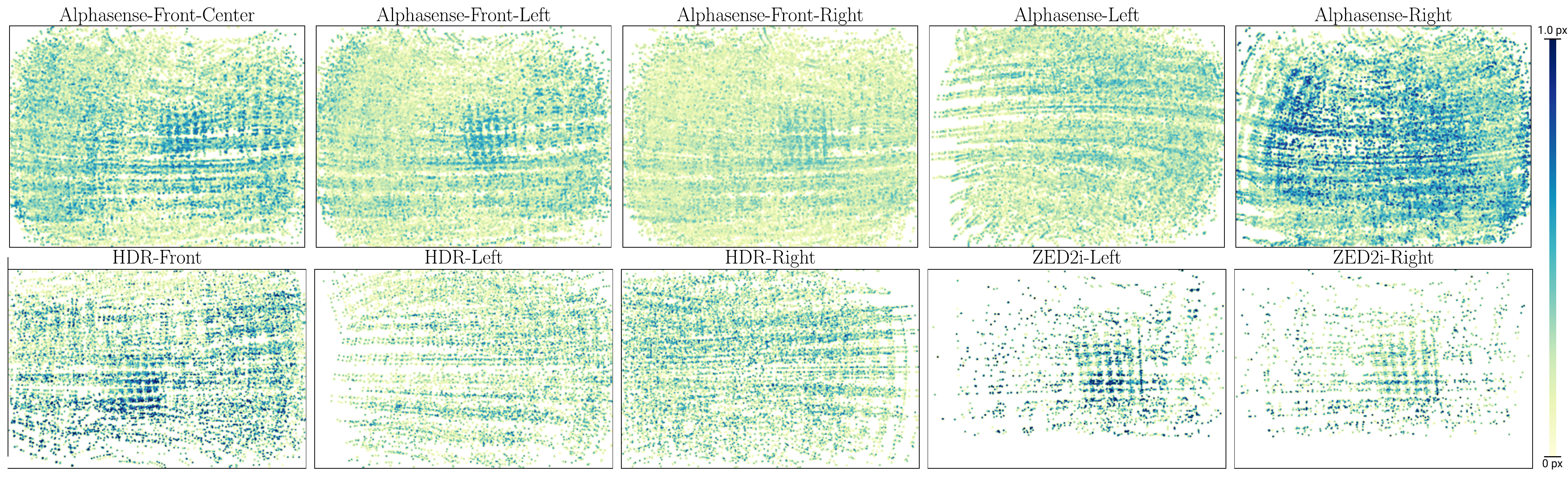} 
    \caption{Heatmap of reprojection errors across all cameras. The consistent sub-pixel reprojection array throughout the image plane verifies the selection of the respective distortion model. Units are in pixels.}
    \label{fig:reprojection_intrinsics}
\end{figure*}

\subsubsection{Weatherproofing}
To ensure weatherproofing, we incorporated an O-ring between the main body of \textit{Boxi} and the lid. A modular IP67-rated cable passthrough proved to be particularly useful for the flexible integration of the components on the lid.
The entire enclosure is air-tight; however, to prevent water ingress due to condensation caused by temperature fluctuations between the interior and exterior of the box, we added a pressure relief valve. In previous hardware designs, we experienced short circuits resulting from this condensation, which this design aims to mitigate.

\subsubsection{Protection}
We generally recommend performing a hazard analysis before the mechanical design of the payload. 
This analysis aims to identify risks early in the design process and, in turn, helps to design mitigation strategies. 
Generally, the later risks are identified, the more burdensome their mitigation becomes.
To protect \textit{Boxi} from impacts, we added a 3D-printed structure around the front and the side. The 3D structure specifically protects the exposed lenses from damage. 
The current dual-purpose roll-over cage protecting the LiDARs is insufficient to absorb the force generated by the robot falling. A more sophisticated protection concept could introduce a decoupled roll-over cage directly mounted to the robot or an intended breaking point between the robot and \textit{Boxi}.

\begin{magic}[Protection - Recipe]
    \item Hazard Analysis (\textit{Fall, Shock, Water, Temperature})
    \item Mitigation Plan (\textit{Cage, Lenses, Damping, IP-rating}) 
\end{magic}

\subsection{Calibration: Camera Intrinsic and Extrinsic}

\subsubsection{Kalibr Ceres Online Version}
\label{app:calib_int_method}
Initially, we experimented with the off-the-shelf tool Kalibr~\cite{furgale2013kalibr, rehder2016extending} but were quickly set back by three main challenges: \textit{i)} the batched nature of the calibration formulation, along with the information gain computation at each batch iteration made the optimization problem prohibitively slow for our use case, taking up to~\SI{6}{\hour}; \textit{ii)} given the wide coverage of our cameras, sub-batches of our data rarely contained observation data from all cameras, and the optimization problem typically diverged; \textit{iii)} the offline nature of the workflow required transfer of around \SI{100}{\giga\byte} of redundant uncompressed image data which was a bottleneck.

Guided by these challenges, we developed our own calibration workflow that provides live feedback on the validity of the calibration data collected, does not store redundant calibration data, and runs in real-time during data collection.

To make calibration feasible on-board, we delegated tasks across the running computers. The Jetson and NUC platforms that run the camera drivers would also perform calibration target detection. Only the target information is transferred to the operator PC, where the calibration optimization and live user visual feedback are run. Additionally, the calibration target detector nodes log the raw images where the calibration target was detected.\looseness-1

At initialization, the calibration routine estimates only the individual camera intrinsics. To achieve this, data is accumulated from all the cameras as the user moves the calibration target within the FoV of the camera bundle. Simultaneously, a separate visualization program shows a heatmap of the regions in each image that have already received calibration data. Once the program receives sufficient new observations (in this case, 300 corners), a non-linear least squares optimization, implemented using Ceres Solver~\cite{Agarwal_Ceres_Solver_2022}, is run to solve for the camera intrinsic parameters using a pre-determined initial guess of the parameters.\looseness-1
This phase temporarily halts the processing of new target detections, displaying on-screen feedback to inform the user that the program is currently busy running an optimization process.
To keep the interaction close to real-time, the optimization program is run for a maximum of three iterations.\looseness-1

\begin{figure*}[t]     
    \centering
    \includegraphics[width=1.0\linewidth]{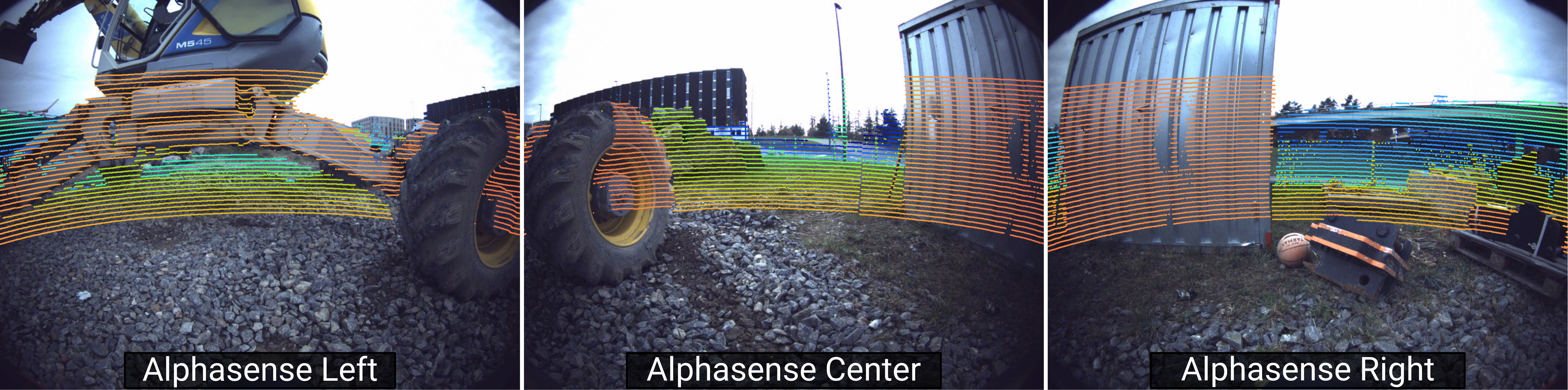}
    \caption{Overlay of the LiDAR point cloud while moving forward at \SI{1}{\meter\per\second}, showcasing the consistency of the sensor calibration and the motion undistortion of the point cloud.}
    \label{fig:lidar_reprojection}
\end{figure*}

After solving the optimization problem, we estimate the diagonal of the covariance of the intrinsic parameters and the reprojection errors for each camera. The user receives feedback on this information through a 2D plot of the reprojection errors and a bar plot of the uncertainty of the intrinsic parameters. Once the maximum standard deviation of the projection parameters is less than two pixels across all cameras, the calibration program proceeds and is ready for extrinsic calibration.\looseness-1

Unlike the data used for intrinsic calibration, the data for extrinsic calibration requires the calibration target to remain static relative to the sensors. To ensure this, we detect a static target if its previous and next positions show less than \SI{1}{\milli\meter} of movement. To synchronize between the cameras, we use a ring buffer to store the last 10 detected target positions, timestamps, and detections for each camera.

After the joint intrinsic/extrinsic optimization, we estimate the diagonal of the covariance of all intrinsic and extrinsic parameters, and report these to the user. The user is recommended to collect calibration data such that the estimated square root of the diagonal of the covariance matrix is under \SI{1}{\milli\meter}.

To reduce the amount of redundant calibration data collection, we first experimented with an information gain-based threshold but found that the computation of the information matrix was too time-consuming to use effectively on-board, even when using sparse matrix implementations~\cite{suitesparse}. Instead, we opted for an approximate but effective measure of information gain. We divided the image into $3\times3$ blocks, allocating to each block a maximum capacity of 10 corner observations. When a new observation is received (each observation contains up to 144 corner locations), the corners are mapped to their designated blocks, and if any of the mapped blocks still have the capacity for new observations, all corners of this new observation are added to the respective blocks. Note that this means that each block can exceed its set maximum capacity.

\subsubsection{Calibration Target}
\label{app:target}
To facilitate the convenient collection of calibration corner data close to the edges of the cameras with high distortion, we used a calibration target formed of a planar grid of AprilTag targets~\cite{aprilgrid2011} (also known as an AprilGrid), which enables partial detection of the calibration target. The  $6\times6$ of square patterns provide up to $144$ corner detections. With squares on the sides \SI{8.3}{\centi\meter}, the target is confidently detected in the range of the calibration data collected.\looseness-1 %

\subsubsection{Camera Calibration Reprojection Errors}
\label{app:cam_reproj_err}
The reprojection errors of the detected corners across all cameras are provided in \cref{fig:reprojection_intrinsics}. The projection errors are, on average, below \SI{0.5}{px} and the samples cover the full \gls{fov}.

\subsection{Calibration: Camera to IMU}
\label{app:camera_to_imu}
An operator performs the calibration motion manually, smoothly moving the cameras along all three axes of translation and rotation while maintaining the view of the calibration target in the cameras. This calibration is done in post-processing, as the calibration target detector is not fast enough to run in real-time without dropping some of the images streaming at \SI{17}{\hertz}.\looseness-1

Moreover, the STIM320 IMU has individual accelerometer axes, so we calibrate its extrinsic parameters relative to the camera accounting for the \textit{axis-scale} using the extension to Kalibr~\cite{kalibr_multi_axes2016}. All other IMUs are individually calibrated relative to the camera bundle, without refinements to the intra-camera extrinsic parameters. The IMUs are placed in very different locations across the sensor suite, with the HG4930 roughly \SI{30}{\centi\meter} away. From the individual calibrations, we can verify that the estimated relative transform between STIM320 and the HG4930 is consistent with the pinned-down CAD measurements. \looseness-1

\subsection{Calibration: LiDAR to Camera Extrinsic}
\label{app:lidar_to_camera}

For this calibration, we used a different calibration target formed of a checkerboard pattern with $8\times9$ squares of size~\SI{8}{\centi\meter} with a special surface finish. Crucially, the print finish of the calibration target admits an intensity contrast in the \SI{905}{\nano\meter} wavelength of the laser beams emitted by the LiDARs, as seen in~\cref{fig:lidar_contrast}. Each LiDAR is individually calibrated against the 5 CoreResearch cameras. Using the calibrated extrinsic parameters between the cameras, the detected calibration target poses from each camera are fused and jointly calibrated with respect to each LiDAR, without refining the intra-camera extrinsic parameters.\looseness-1

The calibration data is collected by placing the target statically in various poses around the sensors in the overlapping area between the~\gls{fov} of the LiDARs and the CoreResearch cameras. Our calibration program automatically detects these static segments by assessing the camera-detected calibration target motion. When the inter-frame motion is under \SI{1}{\milli\meter}, the segment is deemed static. Since the Hesai LiDAR has a sparse scanning pattern along the vertical axis, the calibration target is placed diagonally in certain poses as seen in~\cref{fig:lidar_contrast}, to better constrain the alignment of the vertical axis of the LiDAR to the camera. As the Livox LiDAR has a non-repeating pattern, we accumulate all its points in a given static segment to create a single dense point cloud to align to the camera detections. This data collection typically takes under \SI{10}{\min}.

\subsubsection{LiDAR to Camera Overlay}
\label{app:lidar_to_camera_overlay}
The established time synchronization, camera intrinsic, camera-to-LiDAR, and camera-to-IMU extrinsic calibration allow us to motion-compensate the LiDAR points and accurately project them onto all CoreResearch images. In \cref{fig:lidar_reprojection}, an example scene showcases the excellent data association that can be achieved across all directions during motion.

\section{Image Recording}
\label{app:image_recording}
In our evaluation of image recording, we considered different alternatives, each offering distinct advantages and limitations. Firstly, recording images in raw format was impractical due to the large storage requirements. Lossless PNG compression, while preserving image quality, proved too computationally expensive for real-time processing of large images. MJPEG, being more efficient, offered a higher compression rate (typically around 10:1, depending on quality settings), and benefited from native support in ROS.
Lastly, H.265 video compression can further decrease storage requirements up to a factor of 100 and leverage hardware acceleration on the Jetson, which significantly reduces the CPU workload. 

For all CoreResearch cameras, we opted for standard ROS1 JPEG compression due to the NUC's inability to support hardware video encoding, while still being capable of handling the CPU compression workload. A notable drawback of this method is that images must be transferred through the ROS1 TCP/IP layer before being stored, introducing significant overhead due to serialization and deserialization processes. While there are more sophisticated compression methods that can run on CPU, this approach was necessary as we lacked access to the camera drivers to maximize efficiency in storing serialized image data.
Similarly, for the HDR cameras, we also applied JPEG compression, but we used ROS2, which allowed us to leverage interprocess communication (IPC). This enables efficient storage of compressed images directly into MCAP files, highlighting one of the key advantages of ROS2 for our application.
Lastly, for the ZED2i camera, we used the StereoLabs API to record highly serialized and efficient SVO2 files, benefiting from hardware-accelerated H.265 video compression with an SSIM \SI{97.3}{\%} providing the best image quality and compression ratio while using the least resources.

\end{document}